\documentclass[11pt]{article}

\usepackage[preprint]{acl}
\usepackage{times}
\usepackage{latexsym}
\usepackage{adjustbox}
\usepackage{booktabs}
\usepackage[table]{xcolor}
\usepackage{array}
\usepackage{tabularx}
\usepackage{pifont}
\newcommand{\cmark}{\ding{51}}

\usepackage[multiple]{footmisc}
 \usepackage{longtable}

\newcolumntype{L}[1]{>{\raggedright\arraybackslash}p{#1}}

\usepackage[utf8]{inputenc}

\usepackage[T1]{fontenc}
\usepackage[english]{babel}

\usepackage{fontspec}
\usepackage{polyglossia}
\usepackage[most]{tcolorbox}
\newtcolorbox{promptbox}[1]{
  colback=gray!8, colframe=gray!45, boxrule=0.4pt, arc=1.5pt,
  left=3pt, right=3pt, top=3pt, bottom=3pt,
  fonttitle=\bfseries\small, title=#1,
  before upper={\small\ttfamily\setlength{\parindent}{0pt}}
}

\newfontfamily\timesfont{texgyretermes-regular.otf}[
  BoldFont       = texgyretermes-bold.otf,
  ItalicFont     = texgyretermes-italic.otf,
  BoldItalicFont = texgyretermes-bolditalic.otf
]

\usepackage{xurl}          
\usepackage{hyperref}
\usepackage{fontawesome5}  

\hypersetup{
  colorlinks=true,
  urlcolor=blue
}

\newcommand{\hfrepo}[1]{%
  \noindent
  \href{#1}{%
    \raisebox{-1.0\height}{%
      \includegraphics[height=1.45em]{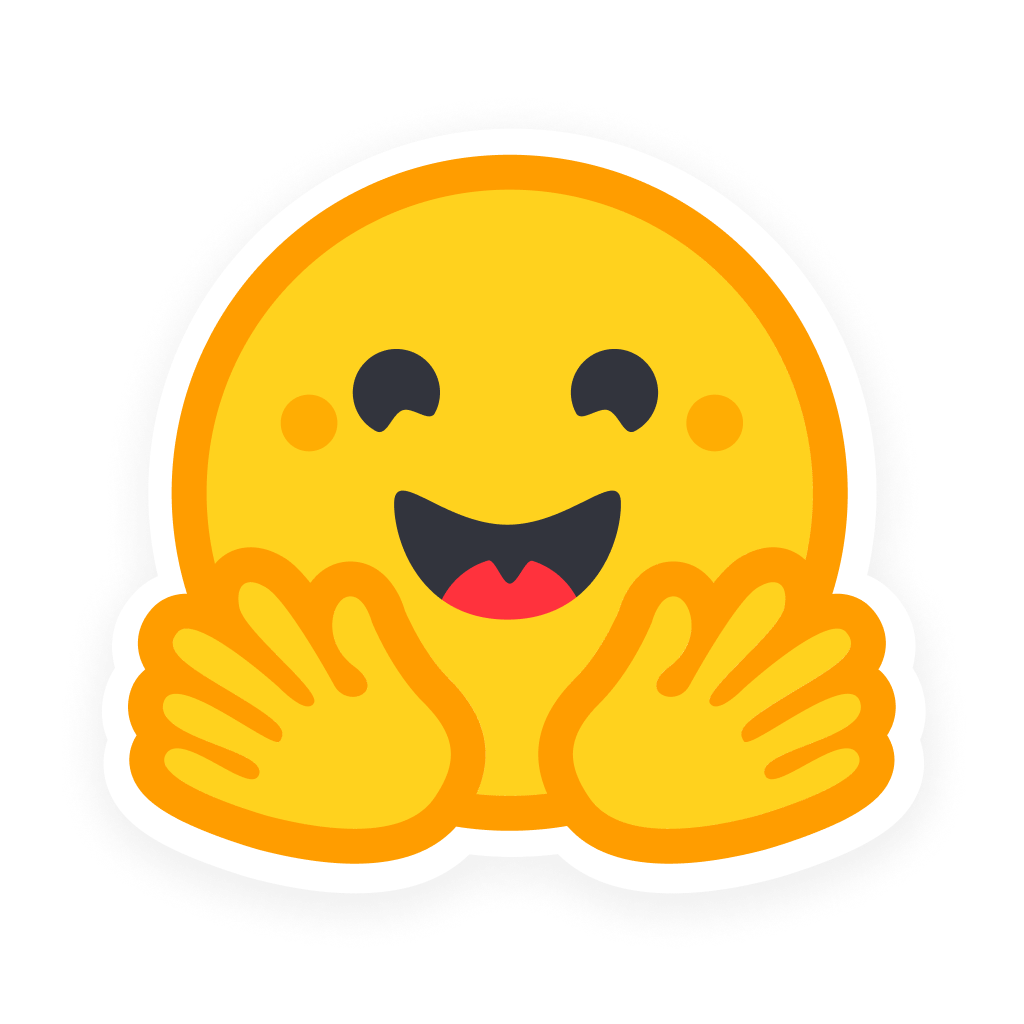}%
    }%
  }%
  \hspace{0.35em}%
  \begin{minipage}[t]{0.77\linewidth}
    \vspace{0pt}%
    \small\ttfamily\url{#1}%
  \end{minipage}%
  \par\medskip
}

\newfontfamily\greekfont[
  Script = Greek,
  BoldFont       = texgyretermes-bold.otf,
  ItalicFont     = texgyretermes-italic.otf,
  BoldItalicFont = texgyretermes-bolditalic.otf
]{texgyretermes-regular.otf}

\newfontfamily\cyrillicfont[
  Path = ./
]{NotoSerif-Regular.ttf}

\newfontfamily\ipafont[
  Path = ./
]{CharisSIL-Regular.ttf}

\newfontfamily\japanesefont[
  Path = ./
]{NotoSerifCJKjp-Regular.otf}

\newfontfamily\arabicfont[
  Path = ./
]{NotoNaskhArabic-Regular.ttf}

\newfontfamily\gurmukhifont[
  Path = ./
]{NotoSerifGurmukhi-Regular.ttf}

\newfontfamily\telugufont[
  Path = ./
]{NotoSerifTelugu-Regular.ttf}

\newfontfamily\kannadafont[
  Path = ./
]{NotoSerifKannada-Regular.ttf}

\newfontfamily\devanagarifont[
  Path = ./
]{NotoSerifDevanagari-Regular.ttf}

\newfontfamily\thaifont[
  Path = ./
]{NotoSerifThai-Regular.ttf}

\newfontfamily\khmerfont[
  Path = ./
]{NotoSerifKhmer-Regular.ttf}

\newfontfamily\bengalifont[
  Path = ./
]{NotoSerifBengali-Regular.ttf}

\newfontfamily\cuneiformfont[
  Path = ./
]{NotoSansCuneiform-Regular.ttf}

\newfontfamily\gujaratifont[
  Path = ./
]{NotoSerifGujarati-Regular.ttf}

\newfontfamily\geofont[
  Path = ./
]{NotoSerifGeorgian-Regular.ttf}

\newfontfamily\syriacfont[
  Path = ./
]{NotoSansSyriac-Regular.ttf}

\newfontfamily\hebrewfont[
  Path = ./
]{NotoSerifHebrew-Regular.ttf}

\usepackage{microtype}
\usepackage{circledsteps}
\usepackage{inconsolata}

\definecolor{HeaderBlue}{HTML}{1F4E79}
\definecolor{SubHeaderBlue}{HTML}{D9EAF7}
\definecolor{RowBlue}{HTML}{F4FAFF}
\definecolor{AggregateRow}{HTML}{FFF2CC}

\usepackage{graphicx}

\usepackage{xspace}

\newcommand{\psy}{\Psi}  
\newcommand{\cs}{\lambda}  
\newcommand{\gm}{\Phi} 

\newcommand{\systemname}{MUDIDI\xspace}
\newcommand{\stageone}{Stage~1\xspace}
\newcommand{\stagetwo}{Stage~2\xspace}

\newcommand{\vlm}{VLM\xspace}
\newcommand{\vlms}{VLMs\xspace}
\newcommand{\llm}{LLM\xspace}
\newcommand{\llms}{LLMs\xspace}

\title{ \raisebox{-0.25\height}{%
    \includegraphics[height=2.5em, 
      trim=140 120 180 120,
      clip]{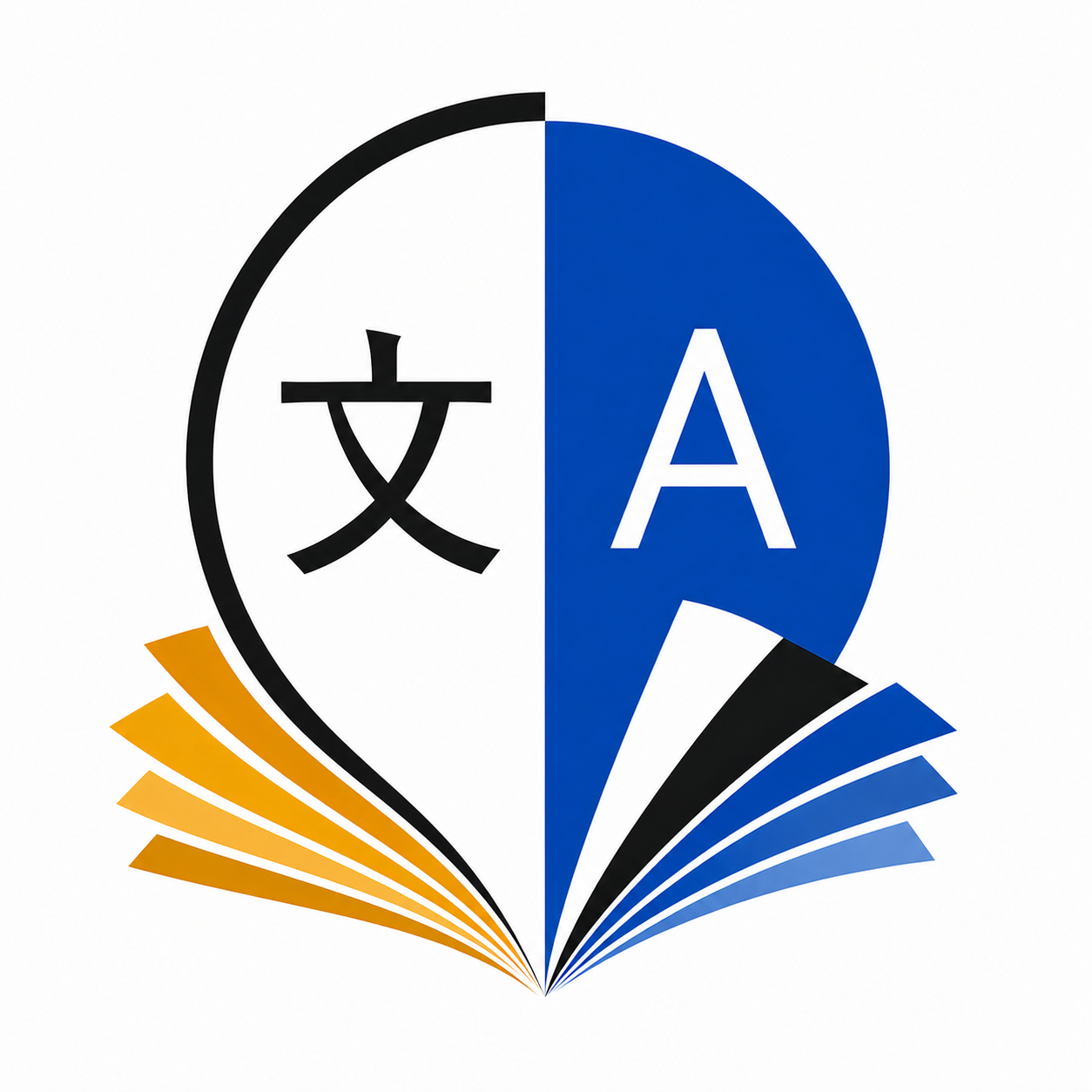}%
  }
  \hspace{0.4em}\textsc{MUDIDI}: A Two-Stage Framework for Multilingual Dictionary Digitization with Language Models}

\author{
  David Setiawan$^\cs$  ~~~
  Temuulen Khishigsuren$^\psy$~~~
  Milind Agarwal$^\gm$  \\
  \textbf{Pagnarith Pit$^\cs$~~
  Aso Mahmudi$^\cs$~~~
  Ekaterina Vylomova$^\cs$}\\
  $^\cs$School of Computing and Information Systems, The University of Melbourne \\
  $^\psy$Melbourne School of Psychological Sciences, The University of Melbourne ~~~ $^\gm$LILT\\
 \texttt{vylomovae@unimelb.edu.au}
}

\begin{document}

\maketitle

\begin{abstract}
Multilingual dictionaries are among the most valuable documentary resources for low-resource and endangered languages, yet many remain available only as scans. For many decades, their digitization and conversion into a machine-readable format was nearly impossible due to language-specific scripts and complex dictionary structure. 
Recent vision-language models offer a promising solution, but it is unclear how well they preserve characters, markup, and process lexicographic structure. We introduce \texttt{\systemname}, a two-stage framework for multilingual dictionary digitization, to fill this gap. \stageone evaluates the quality of character recognition and markup preservation and \stagetwo focuses on dictionary entry segmentation with subsequent mapping into a machine-readable lexicographic structure. 
We 
present a dataset that consists of human-annotated lexicographic entries collected from 30 public-domain dictionaries featuring diverse writing systems, language families, and formats. We benchmark OCR systems, general-purpose Large Language Models (\llms), and Vision Language Models (\vlms) on the dataset, demonstrating superior performance of \llms across most writing systems and languages in both stages. We further provide practical guidelines on improving the results for more challenging scenarios and demonstrate that
supplementing additional information, such as dictionary introduction, to the LLMs can improve the quality of the digitized dictionary.\noindent
\vspace{-1.5em}
\begin{center}
\hfrepo{https://huggingface.co/datasets/Davidsamuel101/MUDIDI/}

\faGithub\quad
\href{https://github.com/naarm-nlp/MUDIDI}{%
\texttt{https://github.com/naarm-nlp/}\\
\texttt{MUDIDI}}

\end{center}
\end{abstract}

\section{Introduction}
``How many words for \textsc{snow} does the Chukchi language have? How many of them share the same stem?'' Although this almost century-old question \cite{whorf1940science} has been widely debated, resolving 
it remains challenging because it requires relevant dictionaries to be available in a structured and machine-readable format \cite{khishigsuren2025computational}. Yet, the vast majority of the dictionaries published in the 19th and 20th centuries are still primarily inaccessible, even when scanned copies exist. Large archives of linguistic fieldwork materials such as PARADISEC \citep{paradisec}\footnote{\url{https://www.paradisec.org.au/}} and ELAR \citep{nathan2013access}\footnote{\url{https://www.elararchive.org/}} would certainly benefit from transforming their collections into machine-readable formats and unlocking the knowledge collected over decades of work. Properly digitized dictionaries would therefore greatly benefit not only linguists and cognitive scientists, but, most importantly, speaker communities.  Multilingual dictionaries are central to language documentation, education, translation, and community-led language revitalization \citep{mosel2004dictionary, garrett2018online}. Digitizing these dictionaries seems especially important because languages disappear at a rate of one every two weeks, with half of the world's languages spoken today predicted to be severely endangered or extinct by the end of the century \citep{un2019}.

\begin{figure*}[h!] 
    \centering
    \includegraphics[width=\textwidth,height=1\textheight,keepaspectratio]{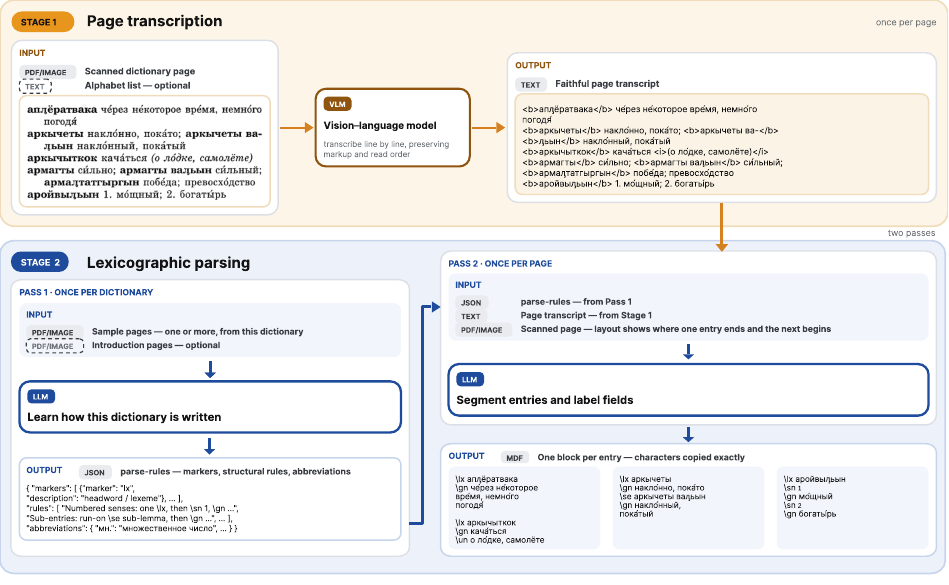}
    \caption{The two-stage dictionary digitization process: Stage 1 results in vanilla OCR, Stage 2 segments the dictionary entries and assigns their parts to MDF fields (for example, \textsc{\textbackslash lx} denotes the headword, \textsc{\textbackslash va} -- variant form, \textsc{\textbackslash hm} -- homonym number, \textsc{\textbackslash sn} -- sense number, \textsc{\textbackslash gn} -- gloss, \textsc{\textbackslash un} --  usage note, and \textsc{\textbackslash se} -- subentry).}
    \label{fig: system}
\end{figure*}

Until recently, the lack of reliable Optical Character Recognition (OCR) systems was one of the major obstacles in digitization, especially for authentic writing systems. Contemporary Vision Language Models (\vlms), general-purpose Large Language Models (\llms) and document-specific multimodal models have changed the landscape of document understanding, with strong results on visual-text and document parsing benchmarks such as OCRBench \citep{fu2026ocrbench}, OmniDocBench \citep{omnidocbench2025}, and Real5-OmniDocBench \citep{zhou2026real5}. The arrival of such technological capabilities is very timely, as they offer a promising solution to support linguistic communities across the world. However, to our knowledge, the models were not evaluated in terms of their ability to process diverse writing systems as well as interpret complex dictionary formats. Dictionary pages present particular challenges: they often contain multiple scripts, dense abbreviations, unusual punctuation, diacritics, multi-column layouts, cross-reference symbols, and entry-specific conventions. 

This paper fills this gap by introducing the first multilingual benchmark and systematically assessing the ability of contemporary language models and OCR systems to process diverse dictionary structures. In particular, we ask the following research questions:  \Circled{1} How well do the models handle diverse writing systems? \Circled{2} How accurately do the models segment and interpret dictionary entries?  
\Circled{3} To what extent can they incorporate external auxiliary information to improve the dictionary processing quality?
To address the questions, we decompose the OCR problem into two parts: (i) faithful page transcription and structure preservation (vanilla OCR), and (ii) lexicographic interpretation and transformation into SIL's Multidictionary Formatter \citep[MDF]{silmdf}. 

Our contributions are threefold:
\begin{enumerate}
    \item We propose a two-stage OCR quality evaluation framework: \stageone focuses on the quality of character recognition, markup preservation, and read order; \stagetwo measures entry segmentation and MDF field assignment;\noindent
    \item We release the first dictionary digitization dataset consisting of three human-annotated pages from each of 
    30 public-domain multilingual dictionaries, covering diverse writing systems, language families, regions, typographic conventions, page layouts, and digitization conditions;
    \item To our knowledge, this is the first paper that evaluates the ability of \llms and \vlms (i) to process such a diverse set of scripts, and (ii) to segment and interpret dictionary entries.
\end{enumerate}

\section{Related work}

\subsection{Document processing and OCR}
\vlms increasingly treat OCR as one capability within a broader document-understanding interface: Qwen2.5-VL's team reports strong document parsing and multilingual visual-text understanding \citep{bai2025qwen2}. Specialized document models such as PaddleOCR-VL-1.5, GLM-OCR, MinerU2.5-Pro, and dots.ocr explicitly target robust document parsing and OCR-style extraction \citep{cui2025paddleocrvl,duan2026glmocr,niu2025mineru,li2025dotsocr}, with PaddleOCR-VL-1.5 and dots.ocr focusing on multilingual page-level parsing. Yet, despite rapid progress in VLM-based document parsing, dictionary digitization remains underexplored.

\subsection{OCR for cultural heritage collections and low-resource languages}
Recent work has argued for revisiting OCR in cultural heritage collections using modern \vlms and large-context models, especially where earlier OCR engines performed poorly on historical scans or unusual layouts \citep{vanstrien2026reocr}. \citet{dasanaike2026vlms} reports successful use of Gemini for digitization of historical census images spanning 9 countries, 5 high-resource languages,  
and 200 years, emphasizing the importance of having access to high-quality scans. 
\citet{jurviste2025vision} used zero-shot \vlms to extract structured JSON from historical Estonian--German lexicographical material
and found, for example, 
that Claude 3.7 Sonnet could produce error-free structured JSON for 41\% of headword entries in one experiment.  Outside lexicography, recent low-resource language work has used digitized grammars and linguistic resources as context for LLM-based translation, but these studies typically rely on traditional OCR systems with post-correction rather than evaluating dictionary-specific extraction \citep{merx2025low,tanzer2024benchmark}. Overall, there is still little systematic work on using VLMs and LLMs to digitize multilingual dictionaries or other linguistic reference works into linguistically structured formats.



\section{Methodology}

\subsection{Task definition}
Given a dictionary page image $I$, the goal is to produce a structured representation $Y$ containing ordered dictionary entries. Following the MDF schema, each entry $e_i$ contains one or more fields from a predefined lexicographic inventory: $
    e_i = \{lx, ps, gn, xv, xn, sd, cf,  \ldots\}, $
where $lx$ is the headword, $ps$ is the part of speech, $gn$ is the translation/gloss in the national language, $xv$ is an example sentence, $xn$ is the national-language translation of the example sentence,  $sd$ is a semantic-domain label if available, and $cf$ is a cross-reference.

As Figure~\ref{fig: system} illustrates, we formalize dictionary digitization as two linked stages:

    \textbf{\stageone: page transcription.} Produce a \textit{faithful} transcription of the dictionary page, with Unicode text and markup preserved using <b> and <i> tags. For multi-column pages, lines follow column-major reading order: the first column top to bottom, and then the next column top to bottom.
    
    \textbf{\stagetwo: lexicographic parsing.} Convert the gold-reference transcript annotated by language experts, together with the original page image, into ordered dictionary entries with MDF-compatible fields. 
    
This decomposition allows transcription quality and lexicographic structure to be measured and improved independently, so errors can be attributed to faithful text recovery versus entry parsing. 

\subsection{Dictionary data}
Our sample primarily contains dictionaries published in the 19th and early 20th centuries that are in public domain and available through HathiTrust \citep{christenson2011hathitrust}.\footnote{It requires signing in to access the data. We also checked and confirmed that there was no data contamination.} We chose languages prioritizing the diversity of writing systems and dictionary formats, limiting the sample to the scripts included in Unicode. In addition, we mainly focused on those translating from a lower- into a higher-resource language as this direction opens opportunities for post-processing the dictionary data using technologies available for high-resource languages (e.g.\ automatic translation between them). 

The dataset features samples of the following 30 dictionaries, spanning 17 families and 20 scripts, with many of them representing endangered languages: Assyrian--English, Bengalese--English, Canala--English, Chepang--English, Chukchi--Russian, Circassian--English--Turkish, Efik--English, Evenki--Russian, Georgian--Russian, Gojri--English--Hindi, Greek--English, Gujarati--English, Iñupiatun Eskimo--English, Japanese--English, Kashmiri--English, Khmer--English, Malay--English, Na--English--Chinese--French, Nahuatl--French, Punjabi--English, Reel--English, Ritharngu--English, Sanskrit--English, Shilluk--English, Syriac--English, Telugu--English, Thai--Russian, Tiri--English, Vernacular Syriac--Kurdish--Turkish--English, Yiddish--English.
Appendix~\ref{app:dictionaries} (Table~\ref{tab:dictionaries-info}) provides more information about the samples, including the writing systems.

From each dictionary, we randomly extract three content pages. For \stagetwo, we additionally collect the most relevant dictionary introduction pages that provide alphabetic characters and explain abbreviations and entry structure. These pages are not part of the main page-level dataset, but are used in prompt-induction experiments. 


\subsection{Data annotation}
\label{sec:data-annotation}
In each stage, we first use a strong system to produce silver-standard annotations that are validated by native speakers and language experts. The final edited files are then included as gold-standard ones in the dataset. The detailed process for each stage is provided below.

 \textbf{Stage 1:} Silver-standard transcripts are produced by prompting a Gemini 3.5-Flash model with the dictionary PDF page and the alphabet list for the source language (prompts are provided in Appendix~\ref{sec:appendix-prompts}). Each dictionary is validated and corrected by a language expert for that language using Label Studio~\cite{labelstudio}; Appendix~\ref{sec:app-label-studio} provides a screenshot of the interface.  The \textbf{gold OCR text} used in model evaluation is derived from these column-structured annotations by fixing a canonical \emph{reading order}: for typical multi-column layouts, lines are ordered column-major (first column top-to-bottom, then the next). When a dictionary spreads a single entry across columns, e.g., a headword in the left column and gloss in the right, we instead order lines \emph{row-wise} (left to right within each row), because one row corresponds to one entry and matches how a reader follows the page. Header and footer lines keep their natural file order. This convention makes the reference sequence meaningful for both human review and automatic scoring; models are compared against the same ordered line list.

 \textbf{Stage 2:} Unlike Stage~1 transcription, Stage~2 requires parsing each page into lexicographic records: segmenting dictionary entries and subentries, assigning SIL's MDF field markers, and reconciling layout, typography, and multilingual gloss structure against both the page image and the Stage~1 transcript. This is a substantially more interpretive task than faithful character reproduction, 
 we therefore use Gemini~3.1-Pro for Stage~2
 to produce the strongest possible silver-standard MDF files for expert annotation. For each page, we produce an MDF file by prompting Gemini~3.1-Pro with gold reference text from Stage~1, the corresponding dictionary page image, and the dictionary's introduction pages when available. 
 
 Following the two-pass approach outlined in \S\ref{sec:stage2-annotation}, annotators correct both the first-pass ``\textit{field-discovery}'' JSON output and the MDF-formatted file produced by Stage~2. Field assignment in the MDF files is corrected in accordance with the SIL's MDF guidelines, and the validated annotations form the gold reference used for Stage~2 evaluation. 
 Because this task requires MDF expertise, which limits the pool of qualified annotators, we focused on 10 dictionaries. Two annotators independently annotated the same page from each dictionary.
 The following dictionaries were chosen to represent diverse formats and descriptive traditions: Chukchi--Russian, Circassian--English--Turkish, Efik--English, Evenki--Russian, Greek--English, I\~nupiatun Eskimo--English, Kashmiri--English, Na--English--Chinese--French, Nahuatl--French, and Tiri--English.
 Inter-annotator agreement between the two annotators was high across all 10 dictionaries, with a Cohen’s kappa of 0.99, indicating almost perfect agreement. The annotators then resolved all disagreements to produce the gold-standard Stage 2 annotations used to evaluate the model outputs.

 Appendix~\ref{app:dictionary_statistics} (Table~\ref{tab:dictionary-stats}) summarizes the resulting gold-standard corpus, including page layout, grapheme counts, and the number of bold and italic markup tags per dictionary, as well as MDF field counts for the 10 dictionaries included in Stage 2.

\subsection{Experimental design}

\subsubsection{Systems}
We compare specialized document VLMs, conventional OCR, and general-purpose LLMs. At Stage~1, we evaluate MinerU2.5 Pro, PaddleOCR-VL-1.5, and GLM-OCR, along with Mathpix as a commercial conventional-OCR baseline. Against these, we evaluate five general-purpose models: Gemini~3 Flash, Gemini~3.1 Pro, GPT-5.5, Claude Opus~4.7, and the open-source \vlm Qwen3-VL-235B-A22B-Instruct.\footnote{Gemini~3 Flash and Gemini~3.1 Pro are accessed via the Gemini API; GPT-5.5, Claude Opus~4.7, and Qwen3-VL are accessed via OpenRouter.} At Stage~2, the parsing experiments are run on Gemini~3.1 Pro, Claude Opus~4.7, GPT-5.5, and the same Qwen3-VL-235B-A22B-Instruct variant; extended reasoning is enabled for the first three models.


\subsubsection{Stage 1}
\label{sec:stage1-annotation}

Stage~1 is a \emph{faithful page transcription (vanilla OCR)} that preserves lightweight markup without segmenting dictionary entries or assigning lexicographic fields. We evaluate two groups of models, each producing the same output format but differing in how transcription is obtained.

\textbf{General LLM.} A prompted \llm receives the page image and, optionally, an alphabet list containing 
valid characters for the source language. The model transcribes the page \emph{line by line}, preserving markup with <b> and <i> tags. 

\textbf{Specialized VLM.} \label{par:specialized-vlm}
Unlike general LLMs, specialized document VLMs including MinerU, PaddleOCR-VL-1.5, and Mathpix operate as \emph{conventional OCR}. We provided a page image or PDF to these models. Because they cannot be prompted, we excluded them from prompt-based ablation studies, except for GLM-OCR which is the only specialized VLM that can be prompted.  A language agnostic post-processing script is applied to each model output so that it is best aligned with the \textbf{gold standard transcript}.

\textbf{Script-level analysis.} Each of the 30 dictionaries examined in this study contains a source language and a target language. The target is often a high-resource language, such as English, written in the Latin script, on which models are expected to perform relatively well. We therefore distinguish between source- and target-language scripts within each dictionary and implement \stageone evaluation separately for each script. To identify the script of each token, we first obtained labels using Gemini 3 Flash and then manually reviewed and corrected them. We then used the corrected labels to aggregate the evaluation metrics by script.

\textbf{Comparative analysis.} \label{par:stage1-comparative}
We investigate whether Stage~1 transcription improves when models receive a source-language alphabet list and auxiliary OCR text for post-correction, following \citet{greif2025multimodal}. We therefore run two ablation studies:  \textbf{1. Alphabet list:} With vs.\ without the source-language alphabet list, with OCR hints disabled throughout. This tests whether explicit character inventory helps models transcribe low-resource scripts more accurately; and \textbf{2. OCR-assisted post-correction:} With vs.\ without an OCR hint, holding Gemini~3.1 Pro with alphabet input fixed across all dictionaries. The hint is the raw output from Mathpix Convert, which achieves the lowest aggregate GCER and WER among the specialized OCR systems in our benchmark. Models are prompted that the OCR text may contain errors and should be used only to resolve ambiguous glyphs; the page image remains authoritative (see Appendix~\ref{sec:appendix-prompts}). This tests whether post-correction improves transcription beyond vision-only decoding.
We expect the alphabet list to help on unfamiliar scripts. We also expect OCR hints to help on dense pages and non-Latin scripts, but they may also anchor models to systematic OCR mistakes.


\subsubsection{Stage 2}
\label{sec:stage2-annotation}

Stage~2 is \emph{lexicographic parsing}: it segments the page transcript into ordered dictionary entries and assigns MDF field markers (headword, pos, definitions, examples, cross-references, and subentry/sense structure). To isolate parsing from OCR quality, we feed \emph{human-validated gold transcriptions} from Stage~1. The LLM is instructed to copy characters from this transcription exactly, using the page image (and optional introduction) only for entry boundaries and field assignment. We supply the page image at this stage because entry structure is often signalled visually rather than textually: indentation, hanging indents, column breaks, and the physical grouping of subentries under a headword carry information that a flat transcript does not preserve. We therefore expect layout context to help the model decide where one entry ends and the next begins, and which lines belong to a subentry rather than a new record. The only allowed changes are structural: removing inline markup, rejoining hyphenated line breaks, and splitting or merging spans across MDF fields. 

The parsing pipeline runs in two passes. \textbf{Pass~1} (once per dictionary) processes the dictionary introduction pages and one sample page to infer the dictionary's MDF markers and entry structure. Its output is a JSON file containing a list of MDF fields, abbreviations, and structural rules for dictionary entry boundaries; we refer to this output as the \textit{parse-rules}. \textbf{Pass~2} (once per page) receives the \textit{parse-rules}, a page snippet, the gold OCR text from Stage~1, and optionally the MDF guidelines and introduction pages (when available), and generates an MDF-formatted text file for the corresponding page.

\textbf{Comparative analysis.} We measure how much the models benefit from \textbf{1. ~Dictionary introduction}, which includes front-matter abbreviations, layout conventions, and part-of-speech keys, supplied to both passes when enabled, and \textbf{2. Official SIL's MDF guidelines}, attached at Pass~2 only to test whether explicit field documentation helps beyond the per-dictionary map inferred in Pass~1. The four combinations (both, introduction only, manual only, neither) are evaluated against the same fixed gold Stage~1 transcripts. 

\textbf{End-to-end analysis.} As mentioned above, we evaluated Stage~1 and Stage~2 independently to isolate the model’s capability to perform text recognition and parsing dictionary entries. However, it is important to evaluate whether errors from Stage~1 text recognition propagate and affect the model’s capability to parse the entries on Stage~2. We therefore also evaluate an end-to-end pipeline in which model-generated Stage~1 transcriptions, rather than gold-standard transcriptions, are provided as input to Stage~2.

\textbf{Other analysis.} We also run a gold parse-rules diagnostic for every dictionary whose best per-dictionary Stage~2 configuration has imperfect MDF Field F1, except Efik (Table~\ref{tab:stage2-mdf-by-dictionary-modelwise-best}). The above ablation studies use the parse-rules \emph{inferred} in Pass~1. To measure how much remaining error is attributable to Pass~1 quality rather than page-level parsing, we re-run each dictionary with the same model and introduction/manual setting but replace the inferred parse-rules with a human-validated gold version before Pass~2. This isolates the effect of parse-rules quality while holding all other inputs fixed (Appendix~\ref{sec:app-stage-2-analisys}, Table~\ref{tab:stage2-gold-cheat-sheet}).
Finally, we evaluate whether including markup from Stage~1 improves the ability of models in Stage~2 to parse OCR text into MDF-formatted entries. To this end, we remove bold and italic tags from the gold Stage~1 transcriptions, provide the resulting text as input to Stage~2, and evaluate the model outputs.

\subsection{Evaluation Metrics}

\subsubsection{\stageone}

Stage~1 evaluates faithful page transcription: whether a system recovers visible characters, preserves typographic emphasis (bold and italic), and outputs lines in an order consistent with the gold reference. In a dictionary, bold typically marks headwords, and italics signal examples or grammatical categories, so accurate transcription and markup are prerequisites for entry structuring in Stage~2. 

OCR-VLMs require running alignment before starting the evaluation part because they often merge or split lines relative to the reference. Following OmniDocBench~\cite{omnidocbench2025}, we use \emph{quick match} at line granularity: each flat line is one unit; we build a matrix of normalised grapheme edit distances (NED), greedily merge adjacent predicted lines while NED improves, then assign pairs by Hungarian matching; remaining units are linked by fuzzy subset search, and pairs whose NED exceeds a rejection threshold are discarded. Character and markup scores use only aligned units; unmatched gold or predicted lines count as errors. This is insensitive to harmless line splits/merges and to content-preserving reordering for character metrics, but order is scored separately below.

\textbf{Character recognition.} We report \textbf{GCER} (grapheme edits / gold graphemes), \textbf{WER}, and \textbf{TextEdit} (mean NED over aligned units), micro-averaged across pages. Distances use Unicode grapheme clusters. For the script-level analysis, where spans are pooled by language-script tag, TextEdit is computed as total grapheme edits divided by the larger of the total gold and predicted grapheme counts.

\textbf{Markup preservation.} On the same aligned units, words are matched in sequence (ignoring case and punctuation for alignment only; pairs kept only if character similarity $\geq 0.5$). We score whether bold and italic tags coincide with the reference on matched words. Summary tables report \textbf{Markup F1}, the F1 score over pooled bold and italic tag matches, indicating how reliably typographic emphasis is preserved.

\textbf{Line read order.} Following \citet{omnidocbench2025} we report \textbf{ReadOrderEdit} which asks whether the model predicted lines appear in the same sequence as the gold reference described above. Reordering lines or omitting them hurts the score even when the transcribed words are largely correct. Lower is better; zero indicates agreement with the annotator-defined order. Read-order scores are macro-averaged by page because each page represents a separate layout challenge.

\subsubsection{\stagetwo}

\stagetwo evaluates structured lexicon extraction: whether a system recovers the correct dictionary entries on a page and assigns MDF fields appropriately. With the transcript held fixed at the gold reference (\S\ref{sec:stage2-annotation}), the metrics below score lexicographic parsing alone, including entry segmentation, sub-entry nesting, and field-marker assignment. Predicted and gold outputs are blank-line-delimited MDF files. Both sides are normalised (Unicode NFC, typography stripped, whitespace collapsed) before matching.

\textbf{Dictionary entry detection.} Each blank-line-separated block is one dictionary entry. Predicted and gold entries are aligned by normalised field content (markup excluded) with a similarity threshold of 0.6. Matched pairs are true positives (TP), unmatched predictions are false positives (FP), and unmatched gold entries are false negatives (FN). These entry-level counts define \textbf{Entry F1}.

\textbf{Field assignment.} Within matched entries, field lines are aligned by content similarity (threshold 0.7). Correct markers, including equivalent gloss/definition markers (e.g. \verb|\ge| vs.\ \verb|\de|), are TP; mislabelled markers contribute one FP and one FN. Unmatched gold and predicted lines, including those in unmatched entries, contribute FN and FP, respectively. These line-level counts define \textbf{MDF Field F1}.

At their respective entry and field-line levels, both metrics are computed as
\[
F_1 = \frac{2\mathrm{TP}}{2\mathrm{TP}+\mathrm{FP}+\mathrm{FN}},
\]
where higher is better.

\textbf{Entry read order.} We report \textbf{ReadOrderEdit}: the normalised Levenshtein edit distance between the predicted and gold entry sequences (range $[0,1]$, lower is better), where unmatched entries count as insertions or deletions. The gold sequence follows the canonical entry order set during annotation (Section~\ref{sec:data-annotation}), which is column-major for typical multi-column layouts and row-wise for dictionaries whose entries span columns. A system that misreads layout is penalised even when individual records match.

\section{Results}
\subsection{Stage 1}
Table~\ref{tab:stage1-aggregate-alphabet} aggregates the results across all dictionaries and illustrates that general-purpose LLMs substantially outperform other types of models, with Gemini being the best performing one across all evaluation metrics except for ReadOrderEdit (Appendix~\ref{sec:app-stage1-full} provides dictionary-specific results). We also observed that the OCR quality of higher-resource parts (written in English, French, Russian) is nearly perfect (we disregarded some inconsistencies in absence/presence of stress marks in Russian) while less commonly used systems present higher error rate. As mentioned in \S\ref{sec:stage1-annotation}, we therefore produced token-level language-script annotations to evaluate OCR quality for each script individually.

\begin{table}[t]
\centering
\small
\setlength{\tabcolsep}{1.6pt}
\renewcommand{\arraystretch}{1.00}
\caption{Stage 1 OCR evaluation results aggregated across 30 dictionaries, comparing transcription with and without the source-language alphabet list in the prompt. Check marks indicate that the alphabet (A) was supplied. \textit{Best aggregate scores are bolded: lower is better for Edit, GCER, WER, and Order; higher is better for Markup F1.}}
\label{tab:stage1-aggregate-alphabet}
\begin{adjustbox}{max width=\textwidth}
\begin{tabular}{lcrrrrr}
\toprule
\textbf{Model} & \textbf{A} & \textbf{Edit} & \textbf{GCER} & \textbf{WER} & \textbf{Mrk. F1} & \textbf{Order} \\
\midrule
\multicolumn{7}{l}{\cellcolor{gray!10}\textit{OCR systems}} \\
GLM-OCR &  & 0.38 & 0.42 & 0.53 & 0.02 & 0.43 \\
GLM-OCR & \cmark & 0.55 & 0.86 & 0.97 & 0.00 & 0.49 \\
Mathpix &  & 0.32 & 0.26 & 0.44 & 0.00 & 0.43 \\
MinerU2.5-Pro &  & 0.29 & 0.56 & 0.67 & 0.00 & 0.35 \\
PaddleOCR-VL-1.5 &  & 0.36 & 0.42 & 0.52 & 0.00 & 0.42 \\
\midrule
\multicolumn{7}{l}{\cellcolor{gray!10}\textit{Vision Language Models}} \\
Qwen3-VL-235B &  & 0.11 & 0.12 & 0.21 & 0.01 & 0.13 \\
Qwen3-VL-235B & \cmark & 0.11 & 0.14 & 0.23 & 0.01 & 0.15 \\
\midrule
\multicolumn{7}{l}{\cellcolor{gray!10}\textit{General-purpose LLMs}} \\
Claude Opus 4.7 &  & 0.06 & 0.07 & 0.14 & 0.71 & 0.11 \\
Claude Opus 4.7 & \cmark & 0.07 & 0.07 & 0.15 & 0.72 & 0.12 \\
GPT-5.5 &  & 0.06 & 0.06 & 0.15 & 0.60 & \textbf{0.02} \\
GPT-5.5 & \cmark & 0.06 & 0.05 & 0.15 & 0.63 & 0.02 \\
Gemini 3 Flash &  & 0.05 & 0.06 & 0.12 & 0.80 & 0.04 \\
Gemini 3 Flash & \cmark & 0.04 & \textbf{0.04} & \textbf{0.12} & \textbf{0.80} & 0.05 \\
Gemini 3.1 Pro &  & 0.04 & 0.05 & 0.12 & 0.77 & 0.03 \\
Gemini 3.1 Pro & \cmark & \textbf{0.04} & 0.05 & 0.12 & 0.80 & 0.04 \\
\bottomrule
\end{tabular}
\end{adjustbox}
\end{table}

\textbf{Script-level transcription quality evaluation.} Appendix~\ref{sec:app-stage1-script} reports transcription performance by language-script pair. 
Overall, general-purpose LLMs perform best across scripts. Gemini achieves the best results on all three metrics for 16 of the 20 scripts, while Claude performs best on all three metrics for Arabic, Latin, and Thai.

Overall, writing systems that are less commonly used or have limited presence in the digital world achieve highest WER and CER scores, with Cuneiform (Assyrian) being the most challenging as the majority of digital resources  primarily rely on romanized transliteration. Special characters in all extended scripts are more likely to be incorrectly predicted (and confused with a similar but more commonly used counterpart). Diacritics present another challenge, especially in less commonly used character combinations. 
Some Arabic-based scripts such as the ones used in Syriac, Kashmiri and Circassian are also challenging. The Circassian case can probably be explained by the change to cyrillic-based writing in the early 20th century, i.e.\ the script is no longer used by speakers of that language. A similar situation appears to be in Thai: the dictionary provides word forms in both Thai script and cyrillic with excessive use of tone markers, with the latter one no longer being in use. Conventional OCR systems primarily perform well on languages with latin-based scripts such as Chepang, Efik, Eskimo, Ritharngu, and others, with Mathpix and MinerU2.5-Pro being particularly strong, but they still lag behind VLMs and LLMs.

\textbf{Alphabet-aware transcription.}
Table~\ref{tab:stage1-aggregate-alphabet} reports the alphabet ablation from \S\ref{par:stage1-comparative} (OCR hints disabled), aggregated over 30 dictionaries.
For general-purpose LLMs, effects are small, model-specific, and vary by metric. Alphabet input improves TextEdit and GCER for both Gemini models and GPT, and improves Markup F1 for Gemini~3.1 Pro, GPT, and Claude, but worsens aggregate reading order for every model. Its effect on WER is also mixed: Gemini~3 Flash and GPT improve slightly, while Gemini~3.1 Pro and Claude get worse scores.
Gemini~3.1 Pro with alphabet input yields the lowest aggregate TextEdit, while Gemini~3 Flash with alphabet input yields the lowest GCER and WER and the highest Markup F1.
Among conventional OCR systems, only GLM-OCR receives the alphabet list and its performance with the alphabet list degrades sharply across all metrics. 
For Qwen3-VL-235B, TextEdit remains similar, while GCER, WER, and ReadOrderEdit worsen with alphabet input. The complete results per dictionary can be found in Appendix~\ref {sec:app-stage1-full}.

\textbf{OCR-assisted prompting} has a mixed but modest effect for Gemini~3.1 Pro with alphabet input. Across the 30 dictionaries, the Mathpix hint slightly improves TextEdit (0.043 to 0.041), GCER (0.045 to 0.043), WER (0.122 to 0.112), and ReadOrderEdit (0.039 to 0.036), while Markup F1 decreases from 0.800 to 0.784. The full per-dictionary breakdown is reported in Table~\ref{tab:stage1-ocr-hint-per-language} in Appendix~\ref{sec:app-ocr-asst}.

\textbf{Markup and layout preservation.}
OCR systems preserve little to no markup and often fail in terms of the reading order. This result contrasts highly with the behavior of LLMs -- they excel in preserving the structure and achieve high F1 in markup recovery. Still, some dictionaries, especially those with Arabic-based orthographies (Circassian, Malay, and Kashmiri), as well as Syriac dictionaries, remain 
challenging.




\subsection{Stage 2}

\begin{table}[t]
\centering
\small
\setlength{\tabcolsep}{2pt}
\caption{Stage 2 MDF evaluation results aggregated by model. Check marks indicate that the corresponding condition was enabled.}
\label{tab:stage2-mdf-aggregate}
\begin{tabular}{lccccc}
\toprule
\textbf{Model} & \textbf{Intro} & \textbf{MDF} & \textbf{Ent. F1} & \textbf{MDF F1} & \textbf{ROE} \\
\midrule
\multicolumn{6}{l}{\cellcolor{gray!10}\textit{Vision Language Models}} \\
Qwen3-VL-235B &  &  & 0.90 & 0.58 & 0.20 \\
Qwen3-VL-235B & \cmark &  & 0.92 & 0.61 & 0.13 \\
Qwen3-VL-235B &  & \cmark & 0.88 & 0.59 & 0.24 \\
Qwen3-VL-235B & \cmark & \cmark & 0.92 & 0.65 & 0.13 \\
\midrule
\multicolumn{6}{l}{\cellcolor{gray!10}\textit{General-purpose LLMs}} \\
Claude Opus 4.7 &  &  & 0.99 & 0.80 & 0.02 \\
Claude Opus 4.7 & \cmark &  & \textbf{1.00} & 0.79 & \textbf{0.00} \\
Claude Opus 4.7 &  & \cmark & \textbf{1.00} & 0.84 & \textbf{0.00} \\
Claude Opus 4.7 & \cmark & \cmark & 0.99 & 0.78 & 0.01 \\
GPT-5.5 &  &  & \textbf{1.00} & 0.78 & \textbf{0.00} \\
GPT-5.5 & \cmark &  & 0.99 & 0.81 & 0.01 \\
GPT-5.5 &  & \cmark & \textbf{1.00} & 0.81 & \textbf{0.00} \\
GPT-5.5 & \cmark & \cmark & \textbf{1.00} & 0.80 & 0.01 \\
Gemini 3.1 Pro &  &  & 0.98 & 0.78 & 0.03 \\
Gemini 3.1 Pro & \cmark &  & 0.99 & 0.83 & 0.02 \\
Gemini 3.1 Pro &  & \cmark & 0.99 & 0.83 & 0.02 \\
Gemini 3.1 Pro & \cmark & \cmark & 0.99 & \textbf{0.88} & 0.01 \\
\bottomrule
\end{tabular}
\end{table}

\label{subsec:stage2-results}
Table~\ref{tab:stage2-mdf-aggregate} summarizes the Stage~2 results from \S\ref{sec:stage2-annotation} on gold Stage~1 transcripts. All general-purpose LLMs achieve high Entry F1 ($\geq$0.99 in most conditions); differences appear mainly in MDF F1 and read-order error.

\textbf{Dictionary introduction.} Including the dictionary introduction improves MDF field assignment by 4--5 F1 points for Gemini and by 4 points for GPT without the manual. Qwen improves on all three metrics, whereas Claude remains near-perfect on Entry F1 but loses MDF F1.

\textbf{MDF reference manual.} The SIL MDF guidelines improve MDF F1 for Gemini by 4--5 points and for Claude by 4 points without the introduction; other effects are mixed. Gemini with both inputs yields the best overall MDF F1. Qwen gains 1--4 points but remains below the general-purpose LLMs on all metrics.

\textbf{End-to-end pipeline.} Table~\ref{tab:stage2-mdf-aggregate-e2e} (Appendix~\ref{sec:app-stage-2-analisys}) reports aggregate results across the 10 Stage~2 dictionaries when model-generated Stage~1 transcriptions are supplied to Stage~2. On average across the general-purpose LLMs and conditions, Entry F1 and MDF F1 are lower and read-order error is higher than in Table~\ref{tab:stage2-mdf-aggregate}. For Qwen, however, the condition without either input achieves the highest Entry F1 and lowest read-order error, and exceeds its parsing-only counterpart on all three metrics.

\textbf{Gold Pass~1 parse-rules.} Human-validated gold parse-rules improve MDF field assignment F1 by 4 points on average across the five evaluated dictionaries, although effects vary by dictionary. Table~\ref{tab:stage2-gold-cheat-sheet} reports the results using each dictionary's best Stage~2 model and introduction/manual setting; Efik is not included (Appendix~\ref{sec:app-stage-2-analisys}).

\textbf{Removal of markups.} Table~\ref{tab:stage2-no-typography} reports aggregate Stage~2 results across 10 dictionaries, comparing performance with and without bold and italic markup in the gold Stage~2 input. For this analysis, we evaluated only Claude Opus 4.7, Gemini 3.1 Pro, and GPT-5.5, with both the introduction and the MDF reference manual included. When markup is removed, Entry F1 and read-order error degrade for Gemini and GPT, but improve for Claude. For MDF field assignment, removing markup improves performance for Claude and GPT, but worsens performance for Gemini.


\section{Practical Recommendations}
\label{sec:recommendations}

Based on our Stage~1 and Stage~2 experiments and on the human annotation workflow, we summarize the following guidance for practitioners digitizing multilingual dictionaries.

\textbf{1. Model choice.}
Among the general-purpose LLMs we evaluated, \textbf{Gemini~3 Flash} and \textbf{Gemini~3.1 Pro} achieve similarly strong character-level transcription quality. Across the tested configurations, Pro yields the lowest aggregate TextEdit, while Flash yields the lowest GCER and WER and the highest markup preservation (bold and italic). Given this trade-off and the lower cost of Flash, we recommend \textbf{Gemini~3 Flash} as the default for Stage~1 transcription unless a specific language or page type clearly favors Pro.
  
\textbf{2. Alphabet list.}
 Supplying a source-language alphabet list can improve transcription on some languages and models, but the effect is modest, model-specific, and varies by metric. A practical workflow is to run Stage~1 \emph{without} an alphabet list first, compare it with an alphabet-conditioned run on a small representative sample, and retain the alphabet list only when it improves transcription on that sample.

\textbf{3. Stochastic and systematic errors.}
During gold OCR text annotation, we observed that many remaining Stage~1 errors are \emph{systematic} (for example, repeated glyph confusions or column-order mistakes on a particular language) rather than random noise, and that some errors change or disappear across repeated runs with the same model.
Concrete examples of both failure modes are shown in Appendix~\ref{sec:app-error-analysis}: a mixed-direction reading-order error in Yiddish--English and a repeated IPA glyph substitution in Chepang--English. When a stable pattern emerges, practitioners should add a dictionary-specific prompt note or lightweight post-processing rule, then validate the intervention on representative pages before full re-transcription. This ability to inject custom rules is a practical advantage of LLM/VLM-based transcription compared to conventional OCR.

\textbf{4. MDF guidelines.} SIL's MDF guidelines can improve field assignment, with Gemini benefiting most consistently; effects for other models depend on whether the dictionary introduction is also supplied.
 
\textbf{5. Stage~2 parse rules.}
  We recommend doing a brief human review and reannotation (when necessary) of each dictionary's \textit{parse rules} and field-assignment conventions, informed by the initial Pass~2 output. A second run with the \emph{modified} \textit{parse rules} can reduce recurring errors, although gains are dictionary-dependent. This step is lower effort than MDF-level re-annotation.

\textbf{6. End-to-end pipeline.} Our analysis suggests that the best-performing models achieve relatively high accuracy for both entry detection and MDF field assignment. This result indicates that practitioners can use the end-to-end pipeline to digitize dictionaries, since annotating gold transcripts for Stage~1 across hundreds of pages is often infeasible in practice.

\section{Conclusion}
We introduced \texttt{\systemname}, a two-stage framework and the first highly multilingual dictionary digitization benchmark, spanning 30 public-domain dictionaries. Separating faithful transcription from MDF parsing enables fine-grained analysis of character, markup, reading-order, entry-segmentation, and field-assignment errors. Gemini~3 Flash and Gemini~3.1 Pro outperform specialized OCR engines and document-focused VLMs across most writing systems, although cuneiform, Syriac, and Arabic-based orthographies remain challenging. Dictionary introductions and SIL MDF guidelines can improve field assignment, although effects vary by model and configuration. We hope \texttt{\systemname} accelerates the conversion of legacy dictionaries into structured resources for speaker communities, archivists, and linguists.


\section{Limitations}

Limitations inspire future research directions. We outline major limitations observed in the current approach.

First, although we aim to cover as diverse a set of languages and scripts as possible, many remain unrepresented in this study. Future work should include a more geographically and genealogically balanced sample of languages and broaden script coverage to include scripts such as Ethiopic (Amharic and Tigrinya), Hangul (Korean), traditional Mongolian, Tamil, Malayalam, and Kannada.

Second, we evaluate only three pages per dictionary in Stage~1 and one page per dictionary in Stage~2. Because bilingual dictionaries typically contain many more pages, it remains unclear how model performance varies across an entire dictionary. Future studies could examine fewer dictionaries while evaluating all their pages. Such studies could help identify recurring errors, particularly in dictionary-entry parsing, and inform the development of automated correction tools.

Third, we did not evaluate whether excluding the dictionary page image from Stage~2 extraction degrades model performance. More generally, it would also be interesting to evaluate a single-stage approach in which the dictionary page image is provided directly to language models to extract structured information. 

Finally, we provide the alphabet lists in text format. Whether providing them as images would improve model performance in the alphabet ablation study remains an open question. Future work could also provide images of language-specific word pairs alongside their corresponding text to examine whether this improves accuracy.


\section{Ethical Considerations}

\paragraph{Human Annotators}
 Gold-standard Stage~1 annotations were collected from one language expert per source language (\S\ref{sec:stage1-annotation}). For Stage~2, two annotators independently annotated one page from each of the ten dictionaries and resolved disagreements (\S\ref{sec:stage2-annotation}). Annotators were recruited through the authors’ personal networks and compensated at approximately US\$30/hour, above the national minimum wage in the relevant jurisdictions and above prevailing local rates for similar annotation work in the annotators’ countries of residence. Informed consent was obtained prior to annotation. The protocol was not subject to formal ethics review: annotation consisted of correcting model-generated drafts.
 \paragraph{Risks}
Dictionary digitization can support language maintenance, education, and research, but it also raises questions of rights, consent, cultural sensitivity, and community control. Some dictionaries contain culturally restricted knowledge or were produced under colonial conditions. 
If the work is done on Indigenous languages, it should follow the FAIR principles balanced by the CARE principles for Indigenous Data Governance and respect Indigenous Cultural and Intellectual Property, involving prior consent on the data usage, attribution of communities as the ICIP owners, and benefit sharing. For community and archival materials, model outputs should be treated as provisional drafts rather than final products, and integrated into community-led data review and management workflows in which speakers play an active role in validating, correcting, and curating their language data. We also emphasize that the paper provides a framework and recommendations to improve the quality of dictionary digitization but the practical implementation, incorporation and use of such a system need to respect the copyright law.

\paragraph{AI assistant disclosure} LLM-based coding assistants were used for code authoring, testing and analysis, and LLMs were used for proofreading and adjusting the phrasing in the manuscript. All claims, methodological designs, and analysis decisions are the authors'. 

\section*{Acknowledgments}
The authors express gratitude to Charles Kemp, Nick Thieberger, and Trevor Cohn for their valuable feedback and helpful discussions. We also gratefully acknowledge the work of our language experts: Doreen Osmelak, Chris Guest, Lakeshia Erlino Kuswoyo, Anudeex Shetty, and  Usha Natalla.
This work was supported by the ARC Discovery Early Career Researcher Award (Grant No. DE260100695). We appreciate the computational resources provided for this research by The University of Melbourne’s Research Computing Services and the Petascale Campus Initiative. 
\bibliography{custom}

\begin{thebibliography}{26}
\providecommand{\natexlab}[1]{#1}

\bibitem[{Bai et~al.(2025)Bai, Chen, Liu, Wang, Ge, Song, Dang, Wang, Wang,
  Tang et~al.}]{bai2025qwen2}
Shuai Bai, Keqin Chen, Xuejing Liu, Jialin Wang, Wenbin Ge, Sibo Song, Kai
  Dang, Peng Wang, Shijie Wang, Jun Tang, and 1 others. 2025.
\newblock {Qwen2. 5-VL} {T}echnical {R}eport.
\newblock \emph{arXiv preprint arXiv:2502.13923}.

\bibitem[{Christenson(2011)}]{christenson2011hathitrust}
Heather Christenson. 2011.
\newblock {HathiTrust}.
\newblock \emph{Library Resources \& Technical Services}, 55(2):93--102.

\bibitem[{Cui et~al.(2025)Cui, Sun, Liang, Gao, Zhang, Liu, Wang, Zhou, Liu,
  Lin et~al.}]{cui2025paddleocrvl}
Cheng Cui, Ting Sun, Suyin Liang, Tingquan Gao, Zelun Zhang, Jiaxuan Liu,
  Xueqing Wang, Changda Zhou, Hongen Liu, Manhui Lin, and 1 others. 2025.
\newblock {PaddleOCR-VL}: Boosting multilingual document parsing via a 0.9b
  ultra-compact vision-language model.
\newblock \emph{arXiv preprint arXiv:2510.14528}.

\bibitem[{Dasanaike(2026)}]{dasanaike2026vlms}
Noah Dasanaike. 2026.
\newblock \href
  {https://www.dropbox.com/scl/fi/kjstgkkofqjs45jugxpcc/dasanaike_vlms.pdf?rlkey=ewkv46l5ghil61u3l66441k31&st=q5zd7410&dl=0}
  {Zero-shot digitization of historical documents with vision language models}.

\bibitem[{Duan et~al.(2026)Duan, Xue, Wang, Su, Liu, Yang, Gan, Wang, Wang, Yan
  et~al.}]{duan2026glmocr}
Shuaiqi Duan, Yadong Xue, Weihan Wang, Zhe Su, Huan Liu, Sheng Yang, Guobing
  Gan, Guo Wang, Zihan Wang, Shengdong Yan, and 1 others. 2026.
\newblock {GLM-OCR} technical report.
\newblock \emph{arXiv preprint arXiv:2603.10910}.

\bibitem[{Fu et~al.(2025)Fu, Kuang, Song, Huang, Yang, Li, Zhu, Luo, Wang, Lu,
  Li, Tang, Shan, Lin, Liu, Wu, Feng, Liu, Huang, Tang, Chen, Jin, Liu, and
  Bai}]{fu2026ocrbench}
Ling Fu, Zhebin Kuang, Jiajun Song, Mingxin Huang, Biao Yang, Yuzhe Li, Linghao
  Zhu, Qidi Luo, Xinyu Wang, Hao Lu, Zhang Li, Guozhi Tang, Bin Shan, Chunhui
  Lin, Qi~Liu, Binghong Wu, Hao Feng, Hao Liu, Can Huang, and 5 others. 2025.
\newblock \href
  {https://proceedings.neurips.cc/paper_files/paper/2025/file/8c2e6bb15be1894b8fb4e0f9bcad1739-Paper-Datasets_and_Benchmarks_Track.pdf}
  {{OCRBench v2}: An improved benchmark for evaluating large multimodal models
  on visual text localization and reasoning}.
\newblock In \emph{Advances in Neural Information Processing Systems},
  volume~38. Curran Associates, Inc.

\bibitem[{Garrett(2018)}]{garrett2018online}
Andrew Garrett. 2018.
\newblock Online dictionaries for language revitalization.
\newblock In \emph{The Routledge handbook of language revitalization}, pages
  197--206. Routledge.

\bibitem[{Greif et~al.(2025)Greif, Griesshaber, and
  Greif}]{greif2025multimodal}
Gavin Greif, Niclas Griesshaber, and Robin Greif. 2025.
\newblock Multimodal {LLMs} for {OCR}, {OCR} post-correction, and named entity
  recognition in historical documents.
\newblock \emph{arXiv preprint arXiv:2504.00414}.

\bibitem[{Harris et~al.(2015)Harris, Thieberger, and Barwick}]{paradisec}
Amanda Harris, Nick Thieberger, and Linda Barwick. 2015.
\newblock \href {https://www.paradisec.org.au/} {Research, records and
  responsibility: Ten years of {PARADISEC}.}
\newblock Sydney University Press.

\bibitem[{Joshi et~al.(2020)Joshi, Santy, Budhiraja, Bali, and
  Choudhury}]{joshi-etal-2020-state}
Pratik Joshi, Sebastin Santy, Amar Budhiraja, Kalika Bali, and Monojit
  Choudhury. 2020.
\newblock \href {https://doi.org/10.18653/v1/2020.acl-main.560} {The state and
  fate of linguistic diversity and inclusion in the {NLP} world}.
\newblock In \emph{Proceedings of the 58th Annual Meeting of the Association
  for Computational Linguistics}, pages 6282--6293, Online. Association for
  Computational Linguistics.

\bibitem[{J{\"u}rviste and Jakobson(2025)}]{jurviste2025vision}
Madis J{\"u}rviste and Joonatan Jakobson. 2025.
\newblock Vision-enabled {LLMs} in historical lexicography: Digitising and
  enriching {Estonian-German} dictionaries from the 17th and 18th centuries.
\newblock \emph{arXiv preprint arXiv:2510.07931}.

\bibitem[{Khishigsuren et~al.(2025)Khishigsuren, Regier, Vylomova, and
  Kemp}]{khishigsuren2025computational}
Temuulen Khishigsuren, Terry Regier, Ekaterina Vylomova, and Charles Kemp.
  2025.
\newblock A computational analysis of lexical elaboration across languages.
\newblock \emph{Proceedings of the National Academy of Sciences},
  122(15):e2417304122.

\bibitem[{Lewis and Simons(2010)}]{lewis2010assessing}
M.~Paul Lewis and Gary~F. Simons. 2010.
\newblock Assessing endangerment: expanding {Fishman’s GIDS}.
\newblock \emph{Revue roumaine de linguistique}, 55(2):103--120.

\bibitem[{Li et~al.(2025)Li, Yang, Liu, Wang, and Zhang}]{li2025dotsocr}
Yumeng Li, Guang Yang, Hao Liu, Bowen Wang, and Colin Zhang. 2025.
\newblock dots. ocr: {M}ultilingual document layout parsing in a single
  vision-language model.
\newblock \emph{arXiv preprint arXiv:2512.02498}.

\bibitem[{Merx et~al.(2025)Merx, Correia, Suominen, and Vylomova}]{merx2025low}
Raphael Merx, Ad{\'e}rito Jos{\'e}~Guterres Correia, Hanna Suominen, and
  Ekaterina Vylomova. 2025.
\newblock \href {https://aclanthology.org/2025.loresmt-1.7/} {Low-resource
  machine translation: {W}hat for? {W}ho for? {A}n observational study on a
  dedicated {Tetun} language translation service}.
\newblock In \emph{Proceedings of the Eighth Workshop on Technologies for
  Machine Translation of Low-Resource Languages (LoResMT 2025)}, pages 54--65.

\bibitem[{Mosel(2004)}]{mosel2004dictionary}
Ulrike Mosel. 2004.
\newblock Dictionary making in endangered speech communities.
\newblock \emph{Language documentation and description}, 2.

\bibitem[{Nathan(2013)}]{nathan2013access}
David Nathan. 2013.
\newblock Access and accessibility at {ELAR}, a social networking archive for
  endangered languages documentation.
\newblock \emph{Oral literature in the digital age: archiving orality and
  connecting with communities}, 2:21.

\bibitem[{Niu et~al.(2026)Niu, Liu, Gu, Wang, Ouyang, Zhao, Chu, He, Wu, Zhang,
  Jin, Liang, Zhang, Zhang, Qu, Ren, Sun, Tang, Niu, Zheng, Ma, Miao, Dong,
  Qian, Zhang, Wang, Chen, Zhao, Wei, Li, Wang, Xu, Cao, Chen, Wu, Gu, Lu, Lin,
  Shenguanlin, Zhou, Zhang, Zang, Dong, Wang, Zhang, Bai, Chu, Li, Wu, Wu, Li,
  Wang, Tu, Xu, Chen, Zhou, Lin, Zhang, and He}]{niu2025mineru}
Junbo Niu, Zheng Liu, Zhuangcheng Gu, Bin Wang, Linke Ouyang, Zhiyuan Zhao, Tao
  Chu, Tianyao He, Fan Wu, Qintong Zhang, Zhenjiang Jin, Guang Liang, Rui
  Zhang, Wenzheng Zhang, Yuan Qu, Zhifei Ren, Yuefeng Sun, Zirui Tang, Boyu
  Niu, and 40 others. 2026.
\newblock \href {https://doi.org/10.18653/v1/2026.acl-industry.3}
  {{M}iner{U}2.5: A decoupled vision-language model for efficient
  high-resolution document parsing}.
\newblock In \emph{Proceedings of the 64th Annual Meeting of the {A}ssociation
  for {C}omputational {L}inguistics (Volume 6: Industry Track)}, pages 13--42,
  San Diego, California, USA. Association for Computational Linguistics.

\bibitem[{Ouyang et~al.(2025)Ouyang, Qu, Zhou, Zhu, Zhang, Lin, Wang, Zhao,
  Jiang, Zhao, Shi, Wu, Chu, Liu, Li, Xu, Zhang, Shi, Tu, and
  He}]{omnidocbench2025}
Linke Ouyang, Yuan Qu, Hongbin Zhou, Jiawei Zhu, Rui Zhang, Qunshu Lin, Bin
  Wang, Zhiyuan Zhao, Man Jiang, Xiaomeng Zhao, Jin Shi, Fan Wu, Pei Chu,
  Minghao Liu, Zhenxiang Li, Chao Xu, Bo~Zhang, Botian Shi, Zhongying Tu, and
  Conghui He. 2025.
\newblock {OmniDocBench}: Benchmarking diverse {PDF} document parsing with
  comprehensive annotations.
\newblock In \emph{Proceedings of the IEEE/CVF Conference on Computer Vision
  and Pattern Recognition (CVPR)}, pages 24838--24848.

\bibitem[{{SIL Language Technology}(2000)}]{silmdf}
{SIL Language Technology}. 2000.
\newblock \href {https://software.sil.org/shoebox/mdf/} {Multi-dictionary
  formatter ({MDF})}.

\bibitem[{Tanzer et~al.(2024)Tanzer, Suzgun, Visser, Jurafsky, and
  Melas-Kyriazi}]{tanzer2024benchmark}
Garrett Tanzer, Mirac Suzgun, Eline Visser, Dan Jurafsky, and Luke
  Melas-Kyriazi. 2024.
\newblock A benchmark for learning to translate a new language from one grammar
  book.
\newblock In \emph{International Conference on Learning Representations},
  volume 2024, pages 18955--18985.

\bibitem[{Tkachenko et~al.(2020)Tkachenko, Malyuk, Holmanyuk, and
  Liubimov}]{labelstudio}
Maxim Tkachenko, Mikhail Malyuk, Andrey Holmanyuk, and Nikolai Liubimov. 2020.
\newblock \href {https://github.com/HumanSignal/label-studio} {{Label Studio}:
  Data labeling software}.
\newblock Open source software available from
  https://github.com/HumanSignal/label-studio.

\bibitem[{{United Nations}(2019)}]{un2019}
{United Nations}. 2019.
\newblock \href
  {https://www.un.org/development/desa/indigenouspeoples/indigenous-languages.html}
  {Rights of indigenous peoples. report of the third committee}.

\bibitem[{van Strien(2026)}]{vanstrien2026reocr}
Daniel van Strien. 2026.
\newblock \href {https://danielvanstrien.xyz/posts/2026/re-ocr-collections/}
  {Re-{OCRing} collections}.
\newblock Blog post.

\bibitem[{Whorf(1940)}]{whorf1940science}
Benjamin~Lee Whorf. 1940.
\newblock Science and linguistics.
\newblock \emph{Technology Review}, 42(6):229--231, 247--248.

\bibitem[{Zhou et~al.(2026)Zhou, Gao, Wang, Gao, Cui, Tang, and
  Liu}]{zhou2026real5}
Changda Zhou, Ziyue Gao, Xueqing Wang, Tingquan Gao, Cheng Cui, Jing Tang, and
  Yi~Liu. 2026.
\newblock {Real5-OmniDocBench}: A full-scale physical reconstruction benchmark
  for robust document parsing in the wild.
\newblock \emph{arXiv preprint arXiv:2603.04205}.

\end{thebibliography}

\clearpage
\onecolumn
\appendix

\section{List of Dictionaries}
\label{app:dictionaries}

\begin{table*}[!ht]
\centering
\small
\adjustbox{max width=\textwidth}{%
\begin{tabular}{L{2.5cm} c c L{2.6cm} L{1.7cm} L{2.6cm} L{2.4cm} L{1.8cm} L{3cm}}
\toprule
\textbf{Source} & \textbf{J20} & \textbf{EGIDS} & \textbf{Language family} & \textbf{Area} & \textbf{Script} & \textbf{Characters} & \textbf{Target} & \textbf{Citation}\\
\midrule
Assyrian & 0 & 10 & Afro-Asiatic & Eurasia & Cuneiform, Latin-EH & {\cuneiformfont \char"12362\char"12154\char"12072} & English & Williams \& Northgate, 1868
\\
Bengali & 3 & 1 & Indo-European & Eurasia & Bengali & {\bengalifont ত্ম ত জ্ঞ} & English  &  Mendies, 1828\\
Canala (Xârâcùù) & 0 & 6a & Austronesian & Papunesia & Latin-EH & {\ipafont mʷ ã ɨ} & English  & Grace, 1975\\
Chepang & 0 & 6b & Sino-Tibetan & Eurasia & Latin-EM & {\ipafont ŋ a ʔ} & English &  Caughley, 2000 \\
Chukchi & 0 & 6b & Chukotko-Kamchatkan & Eurasia & Cyrillic-EM & {\cyrillicfont қ ӈ ԓ} & Russian  & Inenlikei, 1982\\
Circassian (Adyghe) & 1 & 5 & Northwest Caucasian & Eurasia & Arabic-EH, Latin-EM & {\arabicfont ش و نِه} & English, Turkish  & Loewe, 1854 \\
Efik & 0 & 3 & Atlantic-Congo & Africa & Latin-EL & {\ipafont ö ñ ë} & English  & Hugh, 1886\\
Evenki & 0 & 6b & Tungusic & Eurasia & Cyrillic-EL & {\cyrillicfont ӯ э̄ н} & Russian & Vasilevish, 1958 \\
Georgian & 3 & 1 & Kartvelian & Eurasia & Georgian & {\geofont ე ს თ} & Russian  & Kankava, 2001 (3rd ed)\\
Gojri & 0 & 5 & Indo-European & Eurasia & Devanagari, Latin & {\devanagarifont सा र णू} & English, Hindi & Anjum \& Sadiq, 2021 \\
Greek & 3 & 1 & Indo-European & Eurasia & Greek & {\greekfont ξ λ ψ} & English  & Kyriakidēs, 1892\\
Gujarati & 1 & 2 & Indo-European & Eurasia & Gujarati & {\gujaratifont ગ બ ત્તી} & English  & Edalji, 1863\\
Iñupiatun Eskimo & 1 & 8a & Eskimo-Aleut & North America & Latin-EH & u t m & English & Seiler, 2012 \\
Japanese & 5 & 1 & Japonic & Eurasia & Katakana, Kanji, Latin-EL & {\japanesefont フ ナ ド} & English  &  Hepburn, 1886\\
Kashmiri & 1 & 4 & Indo-European & Eurasia & Kashmiri (Arabic-EH), Latin & {\arabicfont کہِ لٹہِ کٹَ} & English  & Chaltra, 1922 \\
Khmer (Cambodian) & 1 & 1 & Austroasiatic & Eurasia & Khmer & {\khmerfont ផ្ អើ ល} & English  & ICC, 2012 \\
Malay & 3 & 3 & Austronesian & Eurasia & Arabic-EM, Latin-EL & {\arabicfont يا غ قر} & English  & Howison, 1801\\
Na (Mosuo) & 0 & 6b & Sino-Tibetan & Eurasia & IPA, Latin-EH & {\ipafont ˩ ɕ ˧} & English, Chinese, French  & Michaud \& Galliot, 2018 \\
Nahuatl & 1 & 6a/b & Uto-Aztecan & North America & Latin-EL & {\ipafont Ç O T} & French &  Siméon, 1885 \\
Punjabi & 2 & 2 & Indo-European & Eurasia & Gurmukhi, Latin-EM & {\gurmukhifont ਕੁ ਚਾ ਰੀ} & English  & Janvier, 1854\\
Reel & 0 & 6a & Nilotic & Africa & Latin-EM, IPA & {\ipafont ɛ̈ ŋ ä} & English  & Cien et al., 2015\\
Ritharngu & 0 & 8b & Pama-Nyungan & Australia & Latin-EH & {\ipafont ṛ č ḍ} & English  & Heath, 1980\\
Sanskrit & 2 & 9 & Indo-European & Eurasia & Devanagari & {\devanagarifont क झ त} & English  & Yates, 1846\\
Shilluk & 0 & 5 & Nilotic & Africa & Latin-EH & {\ipafont ä r ø} & English  & Ayoker \& Kur, 2016 \\
Syriac & 0 & 9 & Afro-Asiatic & Eurasia & Syriac & {\syriacfont ܡ ܪܵ ܐ} & English  & Yohannan, 1900 \\
Telugu & 1 & 2 & Dravidian & Eurasia & Telugu & {\telugufont అ వ ష్టం} & English & Sankaranarayana,  1900\\
Thai & 3 & 1 & Tai-Kadai & Eurasia & Thai, Cyrillic-EH & {\thaifont วั ฒ นะ} & Russian & Morev, 1964 \\
Tiri (Grand Couli) & 0 & 7 & Austronesian & Papunesia & Latin-EH & {\ipafont ɔ̃ bʷ ŋ} & English  & Grace, 1976\\
Vernacular Syriac & 0 & 6b & Afro-Asiatic & Eurasia & Syriac, Latin-EM & {\syriacfont ܬ ܫܸ ܡܲ} & Kurdish, Turkish, English &  Maclean, 1901\\
Yiddish & 1 & 9 & Indo-European & Europasia & Hebrew & {\hebrewfont ע ן פ} & English  & Harkavy, 1901\\
\bottomrule
\end{tabular}}
\caption{Languages and scripts included in the dataset and evaluation; listed alphabetically by source language. The J20 column follows the resource taxonomy of \citet{joshi-etal-2020-state}, ranging from 0 for low-resource languages to 5 for high-resource languages. The EGIDS column follows the Expanded Graded Intergenerational Disruption Scale~\cite{lewis2010assessing}, ranging from 0 for international languages to 10 for extinct languages. In addition, we provide source language scripts. Many of these scripts are based on Latin/Arabic/Cyrillic scripts and include special characters or diacritics; we arranged them into three categories: Lightly Extended (EL) with up to 3 special characters/diacritics; Moderately Extended (EM) with 3 to 6 special characters/diacritics; and Highly Extended (EH) with 6 or more special characters/diacritics. Note that this is a very rough reflection of the authenticity of the corresponding writing system as it neither incorporates the frequency of their usage nor does it distinguish rarely used special characters from tonal markers that could be applied to a range of vowels.}
\label{tab:dictionaries-info}
\end{table*}

\clearpage
\onecolumn

\section{Dictionary Statistics}
\label{app:dictionary_statistics}
\begin{table*}[h!]
\centering
\small
\begin{adjustbox}{max width=\textwidth}
\begin{tabular}{lrrrrrr}
\toprule
\textbf{Dictionary} & \textbf{Rows} & \textbf{Columns} & \textbf{Graphemes} & \textbf{Bold Tags} & \textbf{Italic Tags} & \textbf{MDF Fields} \\
\midrule
Assyrian-English & 99 & 1 & 3,483 & 60 & 47 & -- \\
Bengalese-English & 294 & 2 & 7,611 & 241 & 248 & -- \\
Canala-English & 254 & 2 & 4,098 & 0 & 0 & -- \\
Chepang-English & 254 & 2 & 5,990 & 196 & 141 & -- \\
Chukchi-Russian & 100 & 1 & 2,280 & 94 & 20 & 80 \\
Circassian-English-Turkish & 219 & 3 & 2,117 & 50 & 55 & 110 \\
Efik-English & 296 & 2 & 5,426 & 96 & 227 & 145 \\
Evenki-Russian & 291 & 2 & 4,138 & 242 & 274 & 214 \\
Georgian-Russian & 129 & 2 & 4,302 & 183 & 38 & -- \\
Gojri-English-Hindi & 88 & 1 & 3,503 & 79 & 0 & -- \\
Greek-English & 330 & 2 & 7,637 & 198 & 89 & 246 \\
Gujarati-English & 210 & 2 & 3,847 & 125 & 106 & -- \\
Iñupiatun Eskimo-English & 324 & 2 & 9,283 & 321 & 280 & 273 \\
Japanese-English & 260 & 2 & 9,608 & 243 & 111 & -- \\
Kashmiri-English & 189 & 2 & 3,168 & 110 & 37 & 172 \\
Khmer-English & 244 & 2 & 3,846 & 119 & 0 & -- \\
Malay-English & 214 & 2 & 4,442 & 135 & 0 & -- \\
Na-English-Chinese-French & 255 & 2 & 6,369 & 39 & 0 & 116 \\
Nahuatl-French & 300 & 2 & 10,194 & 102 & 252 & 172 \\
Punjabi-English & 422 & 3 & 7,634 & 253 & 191 & -- \\
Reel-English & 136 & 2 & 2,463 & 198 & 38 & -- \\
Ritharngu-English & 248 & 2 & 4,773 & 203 & 179 & -- \\
Sanskrit-English & 228 & 2 & 3,861 & 293 & 184 & -- \\
Shilluk-English & 150 & 1 & 6,792 & 153 & 98 & -- \\
Syriac-English & 167 & 2 & 3,389 & 159 & 72 & -- \\
Telugu-English & 206 & 2 & 3,954 & 126 & 111 & -- \\
Thai-Russian & 192 & 2 & 4,651 & 127 & 91 & -- \\
Tiri-English & 244 & 2 & 3,873 & 147 & 0 & 102 \\
Vernacular Syriac-Kurdish-Turkish-English & 263 & 2 & 6,436 & 314 & 210 & -- \\
Yiddish-English & 141 & 2 & 5,686 & 201 & 174 & -- \\
\bottomrule
\end{tabular}
\end{adjustbox}
\caption{Per-dictionary corpus statistics. Rows and columns indicate the number of rows and columns included per each dictionary. Graphemes indicate the number of graphemes per gold transcription. Bold and italic tags are the number of markup tags from gold transcription. MDF fields indicate total number of fields included per dictionary, shown only for 10 dictionaries included in Stage 2 evaluation.}
\label{tab:dictionary-stats}
\end{table*}

\clearpage
\onecolumn

\section{Label Studio Setup}
\label{sec:app-label-studio}
\begin{figure*}[h!] 
    \centering
    \includegraphics[width=\textwidth,height=0.8\textheight,keepaspectratio]{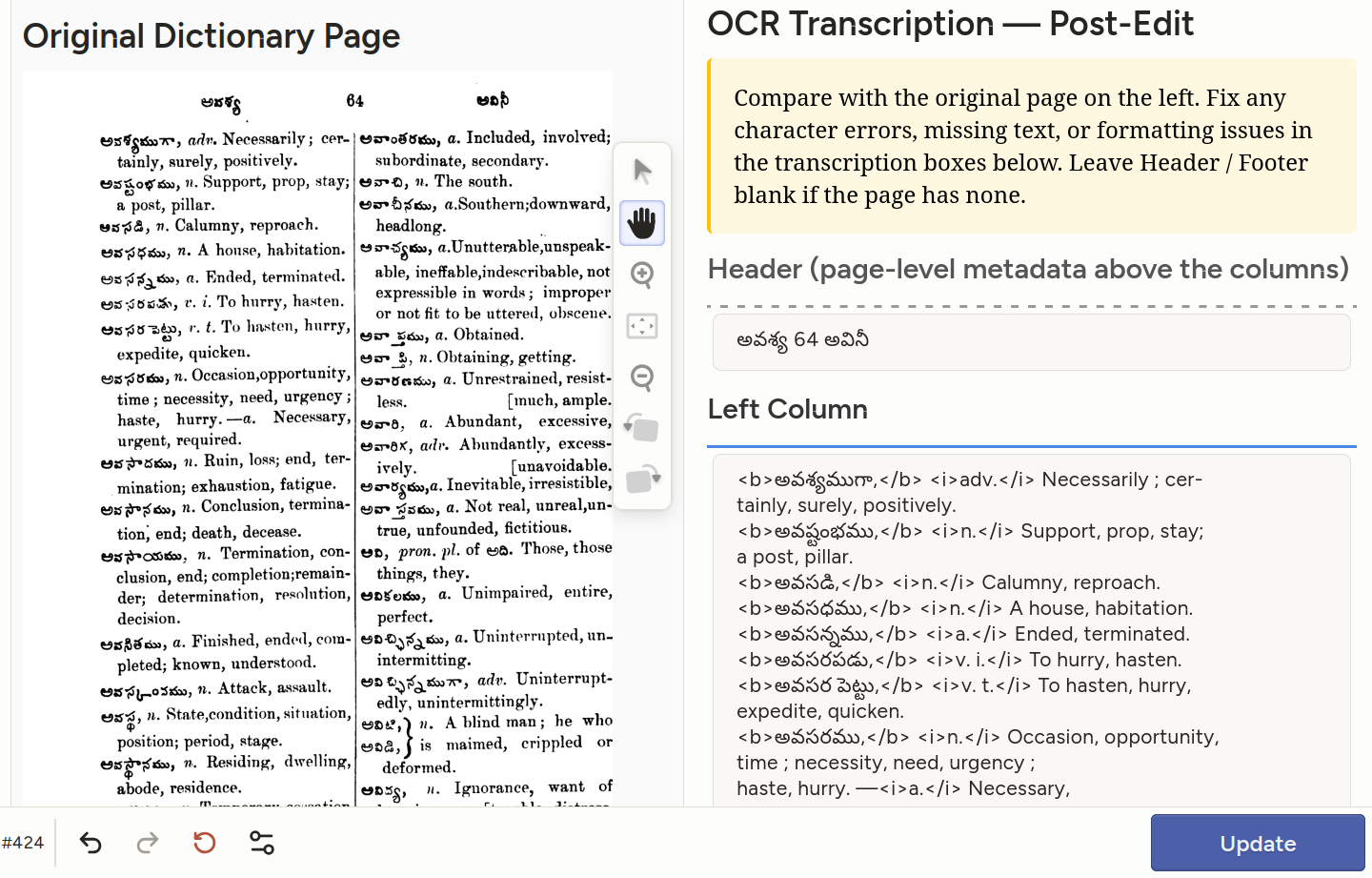}
    \caption{Label Studio Setup. The dictionary page scan is on the left. The language expert uses editor on the right to correct silver-standard annotations produced by a system. Once ready, they press ``Update''. To assist the annotators, we group the text into header, body, and footer; for multi-column pages, we also separate the text by column. }
    \label{fig: label-studio}
\end{figure*}

\clearpage
\onecolumn
\section{Stage 1 Results for Language-Script Pairs}
\label{sec:app-stage1-script}
\begin{table*}[!hp]
\centering
\scriptsize
\setlength{\tabcolsep}{2.2pt}
\renewcommand{\arraystretch}{1.08}
\begin{adjustbox}{max width=\textwidth}
\begin{tabular}{lc@{\quad}rrr@{\qquad}lc@{\quad}rrr}
\toprule
\textbf{Model} & \textbf{Alph.} & \textbf{Edit} & \textbf{GCER} & \textbf{WER} & \textbf{Model} & \textbf{Alph.} & \textbf{Edit} & \textbf{GCER} & \textbf{WER} \\
\midrule

\multicolumn{5}{>{\columncolor{green!12}}l}{\textbf{Assyrian-Cuneiform} (1 dictionaries, 147 words, 352 graphemes)} & \multicolumn{5}{>{\columncolor{green!12}}l}{\textbf{Assyrian-Latin} (1 dictionaries, 140 words, 663 graphemes)} \\
\multicolumn{5}{>{\columncolor{gray!12}}l}{\emph{OCR systems}} & \multicolumn{5}{>{\columncolor{gray!12}}l}{\emph{OCR systems}} \\
GLM-OCR &  & 1.00 & 1.00 & 1.00 & GLM-OCR &  & 0.07 & 0.07 & 0.22 \\
GLM-OCR & \cmark & 1.00 & 1.00 & 1.00 & GLM-OCR & \cmark & 0.08 & 0.08 & 0.24 \\
Mathpix &  & 1.00 & 1.00 & 1.00 & Mathpix &  & 0.31 & 0.31 & 0.47 \\
MinerU2.5-Pro &  & 1.00 & 1.00 & 1.00 & MinerU2.5-Pro &  & 0.65 & 1.51 & 0.35 \\
PaddleOCR-VL-1.5 &  & 1.00 & 1.00 & 1.00 & PaddleOCR-VL-1.5 &  & 0.78 & 0.78 & 0.85 \\
\bottomrule
\end{tabular}
\end{adjustbox}
\caption{Stage 1 language-script-specific evaluation results. \textit{Best scores are bolded (lower is better for all three metrics)}.}
\label{tab:stage1-by-langscript-1}
\end{table*}

\begin{table*}[!hp]
\centering
\scriptsize
\setlength{\tabcolsep}{2.2pt}
\renewcommand{\arraystretch}{1.08}
\begin{adjustbox}{max width=\textwidth}
\begin{tabular}{lc@{\quad}rrr@{\qquad}lc@{\quad}rrr}
\toprule
\textbf{Model} & \textbf{Alph.} & \textbf{Edit} & \textbf{GCER} & \textbf{WER} & \textbf{Model} & \textbf{Alph.} & \textbf{Edit} & \textbf{GCER} & \textbf{WER} \\
\midrule
\multicolumn{5}{>{\columncolor{green!12}}l}{\textbf{Chukchi-Cyrillic} (1 dictionaries, 108 words, 750 graphemes)} & \multicolumn{5}{>{\columncolor{green!12}}l}{\textbf{Circassian-Arabic} (1 dictionaries, 99 words, 442 graphemes)} \\
\multicolumn{5}{>{\columncolor{gray!12}}l}{\emph{OCR systems}} & \multicolumn{5}{>{\columncolor{gray!12}}l}{\emph{OCR systems}} \\
GLM-OCR &  & 0.26 & 0.27 & 0.75 & GLM-OCR &  & 1.00 & 1.00 & 1.00 \\
GLM-OCR & \cmark & 0.26 & 0.27 & 0.81 & GLM-OCR & \cmark & 1.00 & 1.00 & 1.00 \\
Mathpix &  & 0.11 & 0.11 & 0.60 & Mathpix &  & 0.67 & 0.67 & 0.94 \\
MinerU2.5-Pro &  & 0.10 & 0.10 & 0.57 & MinerU2.5-Pro &  & 0.81 & 0.81 & 0.96 \\
PaddleOCR-VL-1.5 &  & 0.17 & 0.17 & 0.70 & PaddleOCR-VL-1.5 &  & 0.88 & 0.88 & 1.00 \\
\bottomrule
\end{tabular}
\end{adjustbox}
\caption{Stage 1 language-script-specific evaluation results (continued). \textit{Best scores are bolded (lower is better for all three metrics)}.}
\label{tab:stage1-by-langscript-2}
\end{table*}

\begin{table*}[!hp]
\centering
\scriptsize
\setlength{\tabcolsep}{2.2pt}
\renewcommand{\arraystretch}{1.08}
\begin{adjustbox}{max width=\textwidth}
\begin{tabular}{lc@{\quad}rrr@{\qquad}lc@{\quad}rrr}
\toprule
\textbf{Model} & \textbf{Alph.} & \textbf{Edit} & \textbf{GCER} & \textbf{WER} & \textbf{Model} & \textbf{Alph.} & \textbf{Edit} & \textbf{GCER} & \textbf{WER} \\
\midrule
\multicolumn{5}{>{\columncolor{green!12}}l}{\textbf{French-Latin} (5 dictionaries, 1,911 words, 8,246 graphemes)} & \multicolumn{5}{>{\columncolor{green!12}}l}{\textbf{Georgian-Georgian} (1 dictionaries, 189 words, 1,383 graphemes)} \\
\multicolumn{5}{>{\columncolor{gray!12}}l}{\emph{OCR systems}} & \multicolumn{5}{>{\columncolor{gray!12}}l}{\emph{OCR systems}} \\
GLM-OCR &  & 0.03 & 0.03 & 0.06 & GLM-OCR &  & 0.93 & 0.93 & 1.00 \\
GLM-OCR & \cmark & 0.69 & 2.05 & 0.15 & GLM-OCR & \cmark & 0.93 & 5.62 & 1.00 \\
Mathpix &  & 0.03 & 0.03 & 0.06 & Mathpix &  & 0.13 & 0.13 & 0.50 \\
MinerU2.5-Pro &  & 0.00 & 0.00 & 0.01 & MinerU2.5-Pro &  & 1.01 & 1.01 & 1.00 \\
PaddleOCR-VL-1.5 &  & 0.02 & 0.02 & 0.03 & PaddleOCR-VL-1.5 &  & 1.00 & 1.00 & 1.00 \\
\bottomrule
\end{tabular}
\end{adjustbox}
\caption{Stage 1 language-script-specific evaluation results (continued). \textit{Best scores are bolded (lower is better for all three metrics)}.}
\label{tab:stage1-by-langscript-3}
\end{table*}

\begin{table*}[!hp]
\centering
\scriptsize
\setlength{\tabcolsep}{2.2pt}
\renewcommand{\arraystretch}{1.08}
\begin{adjustbox}{max width=\textwidth}
\begin{tabular}{lc@{\quad}rrr@{\qquad}lc@{\quad}rrr}
\toprule
\textbf{Model} & \textbf{Alph.} & \textbf{Edit} & \textbf{GCER} & \textbf{WER} & \textbf{Model} & \textbf{Alph.} & \textbf{Edit} & \textbf{GCER} & \textbf{WER} \\
\midrule
\multicolumn{5}{>{\columncolor{green!12}}l}{\textbf{Japanese-Kanji} (1 dictionaries, 101 words, 200 graphemes)} & \multicolumn{5}{>{\columncolor{green!12}}l}{\textbf{Japanese-Katakana} (1 dictionaries, 148 words, 575 graphemes)} \\
\multicolumn{5}{>{\columncolor{gray!12}}l}{\emph{OCR systems}} & \multicolumn{5}{>{\columncolor{gray!12}}l}{\emph{OCR systems}} \\
GLM-OCR &  & 0.38 & 0.38 & 0.60 & GLM-OCR &  & 0.15 & 0.15 & 0.43 \\
GLM-OCR & \cmark & 0.39 & 0.39 & 0.59 & GLM-OCR & \cmark & 0.16 & 0.16 & 0.47 \\
Mathpix &  & 0.48 & 0.48 & 0.71 & Mathpix &  & 0.37 & 0.37 & 0.69 \\
MinerU2.5-Pro &  & 0.35 & 0.35 & 0.61 & MinerU2.5-Pro &  & 0.07 & 0.07 & 0.22 \\
PaddleOCR-VL-1.5 &  & 0.35 & 0.35 & 0.58 & PaddleOCR-VL-1.5 &  & 0.20 & 0.20 & 0.48 \\
\bottomrule
\end{tabular}
\end{adjustbox}
\caption{Stage 1 language-script-specific evaluation results (continued). \textit{Best scores are bolded (lower is better for all three metrics)}.}
\label{tab:stage1-by-langscript-4}
\end{table*}

\begin{table*}[!hp]
\centering
\scriptsize
\setlength{\tabcolsep}{2.2pt}
\renewcommand{\arraystretch}{1.08}
\begin{adjustbox}{max width=\textwidth}
\begin{tabular}{lc@{\quad}rrr@{\qquad}lc@{\quad}rrr}
\toprule
\textbf{Model} & \textbf{Alph.} & \textbf{Edit} & \textbf{GCER} & \textbf{WER} & \textbf{Model} & \textbf{Alph.} & \textbf{Edit} & \textbf{GCER} & \textbf{WER} \\
\midrule
\multicolumn{5}{>{\columncolor{green!12}}l}{\textbf{Na-IPA} (1 dictionaries, 185 words, 991 graphemes)} & \multicolumn{5}{>{\columncolor{green!12}}l}{\textbf{Na-Latin} (1 dictionaries, 31 words, 124 graphemes)} \\
\multicolumn{5}{>{\columncolor{gray!12}}l}{\emph{OCR systems}} & \multicolumn{5}{>{\columncolor{gray!12}}l}{\emph{OCR systems}} \\
GLM-OCR &  & 0.73 & 0.73 & 0.88 & GLM-OCR &  & 0.01 & 0.01 & 0.03 \\
GLM-OCR & \cmark & 1.00 & 4.00 & 0.89 & GLM-OCR & \cmark & 0.01 & 0.01 & 0.03 \\
Mathpix &  & 0.84 & 0.84 & 1.00 & Mathpix &  & 0.26 & 0.26 & 0.26 \\
MinerU2.5-Pro &  & 0.73 & 0.73 & 0.88 & MinerU2.5-Pro &  & 0.16 & 0.19 & 0.03 \\
PaddleOCR-VL-1.5 &  & 0.74 & 0.74 & 0.89 & PaddleOCR-VL-1.5 &  & 0.14 & 0.14 & 0.19 \\
\bottomrule
\end{tabular}
\end{adjustbox}
\caption{Stage 1 language-script-specific evaluation results (continued). \textit{Best scores are bolded (lower is better for all three metrics)}.}
\label{tab:stage1-by-langscript-5}
\end{table*}

\begin{table*}[!hp]
\centering
\scriptsize
\setlength{\tabcolsep}{2.2pt}
\renewcommand{\arraystretch}{1.08}
\begin{adjustbox}{max width=\textwidth}
\begin{tabular}{lc@{\quad}rrr@{\qquad}lc@{\quad}rrr}
\toprule
\textbf{Model} & \textbf{Alph.} & \textbf{Edit} & \textbf{GCER} & \textbf{WER} & \textbf{Model} & \textbf{Alph.} & \textbf{Edit} & \textbf{GCER} & \textbf{WER} \\
\midrule
\multicolumn{5}{>{\columncolor{green!12}}l}{\textbf{Sanskrit-Devanagari} (1 dictionaries, 320 words, 1,034 graphemes)} & \multicolumn{5}{>{\columncolor{green!12}}l}{\textbf{Shilluk-Latin} (1 dictionaries, 157 words, 835 graphemes)} \\
\multicolumn{5}{>{\columncolor{gray!12}}l}{\emph{OCR systems}} & \multicolumn{5}{>{\columncolor{gray!12}}l}{\emph{OCR systems}} \\
GLM-OCR &  & 0.90 & 0.90 & 0.98 & GLM-OCR &  & 0.08 & 0.08 & 0.31 \\
GLM-OCR & \cmark & 0.98 & 0.98 & 1.00 & GLM-OCR & \cmark & 0.07 & 0.07 & 0.29 \\
Mathpix &  & 0.44 & 0.44 & 0.68 & Mathpix &  & 0.04 & 0.04 & 0.14 \\
MinerU2.5-Pro &  & 0.55 & 0.55 & 0.83 & MinerU2.5-Pro &  & 0.04 & 0.04 & 0.19 \\
PaddleOCR-VL-1.5 &  & 0.92 & 0.92 & 0.97 & PaddleOCR-VL-1.5 &  & 0.22 & 0.23 & 0.55 \\
\bottomrule
\end{tabular}
\end{adjustbox}
\caption{Stage 1 language-script-specific evaluation results (continued). \textit{Best scores are bolded (lower is better for all three metrics)}.}
\label{tab:stage1-by-langscript-6}
\end{table*}

\begin{table*}[!hp]
\centering
\scriptsize
\setlength{\tabcolsep}{2.2pt}
\renewcommand{\arraystretch}{1.08}
\begin{adjustbox}{max width=\textwidth}
\begin{tabular}{lc@{\quad}rrr@{\qquad}lc@{\quad}rrr}
\toprule
\textbf{Model} & \textbf{Alph.} & \textbf{Edit} & \textbf{GCER} & \textbf{WER} & \textbf{Model} & \textbf{Alph.} & \textbf{Edit} & \textbf{GCER} & \textbf{WER} \\
\midrule
\multicolumn{5}{>{\columncolor{green!12}}l}{\textbf{Vernacular Syriac-Latin} (1 dictionaries, 159 words, 913 graphemes)} & \multicolumn{5}{>{\columncolor{green!12}}l}{\textbf{Vernacular Syriac-Syriac} (1 dictionaries, 317 words, 1,310 graphemes)} \\
\multicolumn{5}{>{\columncolor{gray!12}}l}{\emph{OCR systems}} & \multicolumn{5}{>{\columncolor{gray!12}}l}{\emph{OCR systems}} \\
GLM-OCR &  & 0.24 & 0.24 & 0.54 & GLM-OCR &  & 1.00 & 1.00 & 1.00 \\
GLM-OCR & \cmark & 0.48 & 0.50 & 0.66 & GLM-OCR & \cmark & 1.19 & 2.41 & 1.00 \\
Mathpix &  & 0.47 & 0.47 & 0.62 & Mathpix &  & 1.00 & 1.00 & 1.00 \\
MinerU2.5-Pro &  & 0.88 & 5.09 & 0.62 & MinerU2.5-Pro &  & 1.00 & 1.00 & 1.00 \\
PaddleOCR-VL-1.5 &  & 0.57 & 0.57 & 0.86 & PaddleOCR-VL-1.5 &  & 1.00 & 1.00 & 1.00 \\
\bottomrule
\end{tabular}
\end{adjustbox}
\caption{Stage 1 language-script-specific evaluation results (continued). \textit{Best scores are bolded (lower is better for all three metrics)}.}
\label{tab:stage1-by-langscript-7}
\end{table*}

\clearpage
\section{Error Analysis}
\label{sec:app-error-analysis}

This appendix illustrates two systematic Stage~1 failure modes in which a model produces a coherent but incorrect transcription rule across many entries. The examples are drawn from the canonical gold OCR and Stage~1 output trees and are intended to show why dictionary-specific instructions can be more effective than treating every discrepancy as independent noise.

\subsection{Mixed-direction reading order in Yiddish--English}
\label{subsec:error-yiddish-order}

The first page of the Yiddish--English dictionary is printed in two columns. Within each entry, the English gloss is on the left and the Hebrew-script headword is on the right. The gold transcription preserves that visual order and reads the complete left column before the right column. Gemini~3.1 Pro without the alphabet list instead emits the Hebrew headword before the English gloss for many entries, effectively reconstructing a conventional headword--gloss order rather than transcribing the page.

\begin{table}[!ht]
\centering
\small
\begin{tabularx}{\linewidth}{@{}>{\bfseries\raggedright\arraybackslash}p{3.0cm}X@{}}
\toprule
Source & Transcription of one entry \\
\midrule
Gold (visual order) &
\texttt{1) to finish praying; 2) to va.} {\hebrewfont אָבבעטען} \\
Gemini~3.1 Pro (no alphabet) &
{\hebrewfont אָבּבּעטען} \texttt{va. 1) to finish praying; 2) to} \\
\bottomrule
\end{tabularx}
\caption{Yiddish--English reading-order error. The model moves the Hebrew headword to the beginning of the entry even though it appears after the English gloss in the image.}
\label{tab:error-yiddish-order}
\end{table}

This is a mixed left-to-right/right-to-left ordering problem rather than a failure to recognise the script alone. The Hebrew token is often legible, but the model's dictionary-schema prior and the absence of explicit token coordinates encourage a semantically natural headword-first serialization. A prompt-level intervention should state that the task is visual transcription: preserve the left-to-right order within every printed line, keep right-to-left tokens in their original visual position, and do not reorder a line into headword--gloss fields.

\subsection{IPA glottal-stop substitution in Chepang--English}
\label{subsec:error-chepang-glottal}

Chepang is marked as IPA+Latin-based in the dictionary configuration, and its alphabet list includes the IPA character {\ipafont ʔ}. Gemini~3 Flash repeatedly replaces this character with the ASCII question mark \texttt{?}. The substitution occurs in word-initial, medial, and final positions, including multiple occurrences in the same word.

\begin{table}[!ht]
\centering
\small
\begin{tabularx}{\linewidth}{@{}>{\bfseries\raggedright\arraybackslash}p{3.0cm}X@{}}
\toprule
Source & Repeated forms \\
\midrule
Gold &
{\ipafont -coʔ.caŋ} \quad {\ipafont ʔuʔ.dhən.ʔi} \\
Gemini~3 Flash (no alphabet) &
{\ipafont -co?.caŋ} \quad {\ipafont ?u?.dhən.?i} \\
\bottomrule
\end{tabularx}
\caption{Chepang--English glyph confusion. The IPA glottal stop {\ipafont ʔ} is systematically rendered as the visually familiar question mark {\ipafont ?}.}
\label{tab:error-chepang-glottal}
\end{table}

Across the three evaluated Chepang pages, the gold contains 272 occurrences of {\ipafont ʔ}; the Gemini~3 Flash no-alphabet output contains none and uses \texttt{?} throughout the corresponding material. The alphabet list alone does not prevent the error, so the useful intervention is an explicit confusable-character instruction: ``The Chepang character {\ipafont ʔ} (Unicode U+0294, LATIN LETTER GLOTTAL STOP) is a valid letter and is distinct from \texttt{?} (U+003F). Preserve every occurrence exactly; never replace it with punctuation.'' A one-page rerun can verify the intervention by checking preservation of U+0294 before expanding it to the full dictionary.

\clearpage
\onecolumn
\section{Stage 1 Results for Each Dictionary}
\label{sec:app-stage1-full}



\begin{table*}[!hp]
\centering
\scriptsize
\setlength{\tabcolsep}{2.2pt}
\renewcommand{\arraystretch}{1.08}
\begin{adjustbox}{max width=\textwidth}
\begin{tabular}{lcrrrrr@{\qquad}lcrrrrr}
\toprule
\textbf{Model} & \textbf{Alph.} & \textbf{Edit} & \textbf{GCER} & \textbf{WER} & \textbf{Mrk. F1} & \textbf{RO} & \textbf{Model} & \textbf{Alph.} & \textbf{Edit} & \textbf{GCER} & \textbf{WER} & \textbf{Mrk. F1} & \textbf{RO} \\
\midrule

\multicolumn{7}{>{\columncolor{green!12}}l}{\textbf{Assyrian-English}} & \multicolumn{7}{>{\columncolor{green!12}}l}{\textbf{Bengalese-English}} \\
\multicolumn{7}{>{\columncolor{gray!12}}l}{\emph{OCR systems}} & \multicolumn{7}{>{\columncolor{gray!12}}l}{\emph{OCR systems}} \\
GLM-OCR &  & 0.56 & 0.23 & 0.36 & 0.00 & 0.33 & GLM-OCR &  & 0.20 & 0.19 & 0.25 & 0.00 & 0.21 \\
GLM-OCR & \cmark & 0.57 & 0.22 & 0.35 & 0.00 & 0.33 & GLM-OCR & \cmark & 0.52 & 0.45 & 0.48 & 0.00 & 0.52 \\
Mathpix &  & 0.70 & 0.36 & 0.46 & 0.00 & 0.45 & Mathpix &  & 0.07 & 0.07 & 0.16 & 0.00 & 0.20 \\
MinerU2.5-Pro &  & 0.56 & 1.82 & 2.42 & 0.00 & 0.42 & MinerU2.5-Pro &  & 0.11 & 0.11 & 0.20 & 0.00 & 0.10 \\
PaddleOCR-VL-1.5 &  & 0.81 & 0.57 & 0.66 & 0.00 & 0.55 & PaddleOCR-VL-1.5 &  & 0.19 & 0.19 & 0.30 & 0.00 & 0.23 \\
\multicolumn{7}{>{\columncolor{gray!12}}l}{\emph{Vision Language Models}} & \multicolumn{7}{>{\columncolor{gray!12}}l}{\emph{Vision Language Models}} \\
Qwen3-VL-235B &  & 0.30 & 0.19 & 0.33 & 0.00 & 0.21 & Qwen3-VL-235B &  & 0.04 & 0.04 & 0.10 & 0.00 & \textbf{0.00} \\
Qwen3-VL-235B & \cmark & 0.29 & 1.11 & 0.98 & 0.00 & 0.19 & Qwen3-VL-235B & \cmark & 0.04 & 0.05 & 0.10 & 0.00 & \textbf{0.00} \\
\multicolumn{7}{>{\columncolor{gray!12}}l}{\emph{General-purpose LLMs}} & \multicolumn{7}{>{\columncolor{gray!12}}l}{\emph{General-purpose LLMs}} \\
Claude Opus 4.7 &  & 0.30 & 0.18 & 0.24 & 0.71 & 0.21 & Claude Opus 4.7 &  & \textbf{0.01} & 0.01 & 0.02 & 0.66 & \textbf{0.00} \\
Claude Opus 4.7 & \cmark & 0.25 & \textbf{0.12} & \textbf{0.23} & 0.71 & 0.19 & Claude Opus 4.7 & \cmark & 0.01 & 0.01 & 0.02 & 0.49 & \textbf{0.00} \\
GPT-5.5 &  & 0.28 & 0.15 & 0.27 & \textbf{0.71} & 0.16 & GPT-5.5 &  & 0.02 & 0.01 & 0.04 & 0.00 & \textbf{0.00} \\
GPT-5.5 & \cmark & 0.29 & 0.18 & 0.28 & 0.71 & 0.12 & GPT-5.5 & \cmark & 0.01 & 0.01 & 0.03 & 0.00 & \textbf{0.00} \\
Gemini 3 Flash &  & 0.26 & 0.18 & 0.45 & 0.52 & 0.11 & Gemini 3 Flash &  & 0.01 & 0.01 & 0.04 & 0.96 & \textbf{0.00} \\
Gemini 3 Flash & \cmark & 0.25 & 0.15 & 0.33 & 0.58 & 0.08 & Gemini 3 Flash & \cmark & 0.01 & 0.01 & 0.04 & \textbf{0.97} & \textbf{0.00} \\
Gemini 3.1 Pro &  & 0.26 & 0.14 & 0.29 & 0.63 & 0.15 & Gemini 3.1 Pro &  & 0.02 & 0.02 & 0.04 & 0.76 & \textbf{0.00} \\
Gemini 3.1 Pro & \cmark & \textbf{0.22} & 0.14 & 0.32 & 0.56 & \textbf{0.06} & Gemini 3.1 Pro & \cmark & 0.01 & \textbf{0.01} & \textbf{0.02} & 0.80 & \textbf{0.00} \\
\addlinespace[0.35em]
\midrule
\addlinespace[0.35em]

\multicolumn{7}{>{\columncolor{green!12}}l}{\textbf{Canala-English}} & \multicolumn{7}{>{\columncolor{green!12}}l}{\textbf{Chepang-English}} \\
\multicolumn{7}{>{\columncolor{gray!12}}l}{\emph{OCR systems}} & \multicolumn{7}{>{\columncolor{gray!12}}l}{\emph{OCR systems}} \\
GLM-OCR &  & 0.13 & 0.11 & 0.22 & \textbf{0.00} & 0.42 & GLM-OCR &  & 0.11 & 0.11 & 0.24 & 0.00 & 0.39 \\
GLM-OCR & \cmark & 0.25 & 0.18 & 0.29 & \textbf{0.00} & 0.43 & GLM-OCR & \cmark & 0.11 & 0.11 & 0.24 & 0.00 & 0.46 \\
Mathpix &  & 0.09 & 0.08 & 0.20 & \textbf{0.00} & 0.39 & Mathpix &  & 0.10 & 0.10 & 0.26 & 0.00 & 0.30 \\
MinerU2.5-Pro &  & 0.12 & 0.10 & 0.26 & \textbf{0.00} & 0.24 & MinerU2.5-Pro &  & 0.10 & 0.11 & 0.25 & 0.00 & 0.34 \\
PaddleOCR-VL-1.5 &  & 0.13 & 0.10 & 0.24 & \textbf{0.00} & 0.24 & PaddleOCR-VL-1.5 &  & 0.13 & 0.12 & 0.28 & 0.00 & 0.47 \\
\multicolumn{7}{>{\columncolor{gray!12}}l}{\emph{Vision Language Models}} & \multicolumn{7}{>{\columncolor{gray!12}}l}{\emph{Vision Language Models}} \\
Qwen3-VL-235B &  & 0.04 & 0.03 & 0.12 & \textbf{0.00} & 0.01 & Qwen3-VL-235B &  & 0.04 & 0.05 & 0.18 & 0.00 & 0.01 \\
Qwen3-VL-235B & \cmark & 0.02 & 0.02 & 0.07 & \textbf{0.00} & \textbf{0.00} & Qwen3-VL-235B & \cmark & 0.04 & 0.05 & 0.18 & 0.00 & 0.01 \\
\multicolumn{7}{>{\columncolor{gray!12}}l}{\emph{General-purpose LLMs}} & \multicolumn{7}{>{\columncolor{gray!12}}l}{\emph{General-purpose LLMs}} \\
Claude Opus 4.7 &  & \textbf{0.01} & \textbf{0.01} & \textbf{0.02} & \textbf{0.00} & \textbf{0.00} & Claude Opus 4.7 &  & 0.01 & 0.01 & 0.04 & \textbf{0.98} & \textbf{0.00} \\
Claude Opus 4.7 & \cmark & 0.01 & 0.01 & 0.03 & \textbf{0.00} & \textbf{0.00} & Claude Opus 4.7 & \cmark & 0.01 & 0.01 & 0.04 & \textbf{0.98} & \textbf{0.00} \\
GPT-5.5 &  & 0.03 & 0.02 & 0.10 & \textbf{0.00} & \textbf{0.00} & GPT-5.5 &  & \textbf{0.01} & \textbf{0.01} & \textbf{0.03} & 0.84 & \textbf{0.00} \\
GPT-5.5 & \cmark & 0.02 & 0.02 & 0.09 & \textbf{0.00} & \textbf{0.00} & GPT-5.5 & \cmark & 0.01 & 0.01 & 0.03 & 0.70 & \textbf{0.00} \\
Gemini 3 Flash &  & 0.02 & 0.02 & 0.07 & \textbf{0.00} & \textbf{0.00} & Gemini 3 Flash &  & 0.05 & 0.05 & 0.19 & 0.90 & \textbf{0.00} \\
Gemini 3 Flash & \cmark & 0.03 & 0.03 & 0.12 & \textbf{0.00} & \textbf{0.00} & Gemini 3 Flash & \cmark & 0.05 & 0.05 & 0.17 & 0.90 & \textbf{0.00} \\
Gemini 3.1 Pro &  & 0.01 & 0.01 & 0.05 & \textbf{0.00} & \textbf{0.00} & Gemini 3.1 Pro &  & 0.04 & 0.04 & 0.15 & 0.92 & \textbf{0.00} \\
Gemini 3.1 Pro & \cmark & 0.02 & 0.01 & 0.07 & \textbf{0.00} & \textbf{0.00} & Gemini 3.1 Pro & \cmark & 0.02 & 0.02 & 0.08 & 0.96 & \textbf{0.00} \\
\addlinespace[0.35em]
\midrule
\addlinespace[0.35em]

\multicolumn{7}{>{\columncolor{green!12}}l}{\textbf{Chukchi-Russian}} & \multicolumn{7}{>{\columncolor{green!12}}l}{\textbf{Circassian-English-Turkish}} \\
\multicolumn{7}{>{\columncolor{gray!12}}l}{\emph{OCR systems}} & \multicolumn{7}{>{\columncolor{gray!12}}l}{\emph{OCR systems}} \\
GLM-OCR &  & 0.17 & 0.17 & 0.67 & 0.33 & \textbf{0.00} & GLM-OCR &  & 0.64 & 0.86 & 0.96 & 0.00 & 0.25 \\
GLM-OCR & \cmark & 0.18 & 0.17 & 0.63 & 0.00 & 0.01 & GLM-OCR & \cmark & 0.72 & 0.71 & 0.81 & 0.00 & 0.35 \\
Mathpix &  & 0.04 & 0.05 & 0.27 & 0.00 & \textbf{0.00} & Mathpix &  & 0.40 & 0.38 & 0.72 & 0.00 & 0.03 \\
MinerU2.5-Pro &  & 0.10 & 0.10 & 0.63 & 0.00 & \textbf{0.00} & MinerU2.5-Pro &  & 0.92 & 1.14 & 1.18 & 0.00 & 0.22 \\
PaddleOCR-VL-1.5 &  & 0.15 & 0.14 & 0.73 & 0.00 & \textbf{0.00} & PaddleOCR-VL-1.5 &  & 0.99 & 1.57 & 1.50 & 0.00 & 0.39 \\
\multicolumn{7}{>{\columncolor{gray!12}}l}{\emph{Vision Language Models}} & \multicolumn{7}{>{\columncolor{gray!12}}l}{\emph{Vision Language Models}} \\
Qwen3-VL-235B &  & 0.11 & 0.10 & 0.44 & 0.00 & \textbf{0.00} & Qwen3-VL-235B &  & 0.46 & 0.48 & 0.69 & 0.00 & \textbf{0.00} \\
Qwen3-VL-235B & \cmark & 0.10 & 0.10 & 0.45 & 0.00 & \textbf{0.00} & Qwen3-VL-235B & \cmark & 0.44 & 0.44 & 0.66 & 0.00 & \textbf{0.00} \\
\multicolumn{7}{>{\columncolor{gray!12}}l}{\emph{General-purpose LLMs}} & \multicolumn{7}{>{\columncolor{gray!12}}l}{\emph{General-purpose LLMs}} \\
Claude Opus 4.7 &  & 0.04 & 0.04 & 0.20 & 0.97 & \textbf{0.00} & Claude Opus 4.7 &  & 0.19 & 0.23 & 0.48 & 0.46 & \textbf{0.00} \\
Claude Opus 4.7 & \cmark & 0.05 & 0.05 & 0.34 & 0.97 & \textbf{0.00} & Claude Opus 4.7 & \cmark & \textbf{0.14} & \textbf{0.18} & \textbf{0.36} & 0.60 & \textbf{0.00} \\
GPT-5.5 &  & 0.08 & 0.08 & 0.41 & 0.97 & \textbf{0.00} & GPT-5.5 &  & 0.33 & 0.38 & 0.65 & 0.55 & \textbf{0.00} \\
GPT-5.5 & \cmark & 0.08 & 0.08 & 0.44 & 0.97 & \textbf{0.00} & GPT-5.5 & \cmark & 0.50 & 0.49 & 0.74 & 0.32 & 0.40 \\
Gemini 3 Flash &  & 0.05 & 0.05 & 0.31 & 0.99 & \textbf{0.00} & Gemini 3 Flash &  & 0.27 & 0.33 & 0.59 & \textbf{0.63} & \textbf{0.00} \\
Gemini 3 Flash & \cmark & \textbf{0.01} & \textbf{0.01} & \textbf{0.07} & \textbf{1.00} & \textbf{0.00} & Gemini 3 Flash & \cmark & 0.28 & 0.32 & 0.61 & 0.48 & \textbf{0.00} \\
Gemini 3.1 Pro &  & 0.03 & 0.04 & 0.19 & 0.80 & \textbf{0.00} & Gemini 3.1 Pro &  & 0.38 & 0.45 & 0.71 & 0.40 & 0.15 \\
Gemini 3.1 Pro & \cmark & 0.05 & 0.05 & 0.30 & 0.98 & \textbf{0.00} & Gemini 3.1 Pro & \cmark & 0.28 & 0.32 & 0.67 & 0.48 & \textbf{0.00} \\
\bottomrule
\end{tabular}
\end{adjustbox}
\caption{Stage 1 dictionary-specific evaluation results grouped by dictionary.}
\end{table*}

\clearpage

\begin{table*}[p]
\centering
\scriptsize
\setlength{\tabcolsep}{2.2pt}
\renewcommand{\arraystretch}{1.08}
\begin{adjustbox}{max width=\textwidth}
\begin{tabular}{lcrrrrr@{\qquad}lcrrrrr}
\toprule
\textbf{Model} & \textbf{Alph.} & \textbf{Edit} & \textbf{GCER} & \textbf{WER} & \textbf{Mrk. F1} & \textbf{RO} & \textbf{Model} & \textbf{Alph.} & \textbf{Edit} & \textbf{GCER} & \textbf{WER} & \textbf{Mrk. F1} & \textbf{RO} \\
\midrule

\multicolumn{7}{>{\columncolor{green!12}}l}{\textbf{Efik-English}} & \multicolumn{7}{>{\columncolor{green!12}}l}{\textbf{Evenki-Russian}} \\
\multicolumn{7}{>{\columncolor{gray!12}}l}{\emph{OCR systems}} & \multicolumn{7}{>{\columncolor{gray!12}}l}{\emph{OCR systems}} \\
GLM-OCR &  & 0.13 & 0.09 & 0.17 & 0.00 & 0.51 & GLM-OCR &  & 0.59 & 2.32 & 2.39 & 0.00 & 0.58 \\
GLM-OCR & \cmark & 0.16 & 0.10 & 0.17 & 0.00 & 0.51 & GLM-OCR & \cmark & 0.78 & 3.16 & 3.28 & 0.00 & 0.50 \\
Mathpix &  & 0.10 & 0.06 & 0.13 & 0.00 & 0.51 & Mathpix &  & 0.23 & 0.30 & 0.57 & 0.00 & 0.69 \\
MinerU2.5-Pro &  & 0.10 & 0.08 & 0.17 & 0.00 & 0.46 & MinerU2.5-Pro &  & 0.18 & 0.23 & 0.55 & 0.00 & 0.51 \\
PaddleOCR-VL-1.5 &  & 0.11 & 0.08 & 0.20 & 0.00 & 0.43 & PaddleOCR-VL-1.5 &  & 0.19 & 0.20 & 0.53 & 0.00 & 0.58 \\
\multicolumn{7}{>{\columncolor{gray!12}}l}{\emph{Vision Language Models}} & \multicolumn{7}{>{\columncolor{gray!12}}l}{\emph{Vision Language Models}} \\
Qwen3-VL-235B &  & 0.03 & 0.05 & 0.09 & 0.00 & 0.17 & Qwen3-VL-235B &  & 0.12 & 0.18 & 0.48 & 0.00 & 0.27 \\
Qwen3-VL-235B & \cmark & 0.06 & 0.07 & 0.12 & 0.00 & 0.36 & Qwen3-VL-235B & \cmark & 0.12 & 0.19 & 0.48 & 0.00 & 0.42 \\
\multicolumn{7}{>{\columncolor{gray!12}}l}{\emph{General-purpose LLMs}} & \multicolumn{7}{>{\columncolor{gray!12}}l}{\emph{General-purpose LLMs}} \\
Claude Opus 4.7 &  & 0.02 & 0.03 & 0.06 & 0.96 & 0.17 & Claude Opus 4.7 &  & 0.06 & 0.07 & 0.23 & 0.74 & 0.17 \\
Claude Opus 4.7 & \cmark & 0.02 & 0.04 & 0.06 & 0.96 & 0.17 & Claude Opus 4.7 & \cmark & 0.06 & 0.07 & 0.29 & 0.90 & 0.19 \\
GPT-5.5 &  & 0.02 & 0.02 & 0.09 & 0.78 & \textbf{0.00} & GPT-5.5 &  & 0.06 & 0.06 & 0.30 & 0.80 & \textbf{0.00} \\
GPT-5.5 & \cmark & 0.02 & 0.02 & 0.08 & 0.89 & \textbf{0.00} & GPT-5.5 & \cmark & 0.06 & 0.06 & 0.29 & 0.76 & \textbf{0.00} \\
Gemini 3 Flash &  & 0.01 & 0.01 & 0.03 & 0.94 & \textbf{0.00} & Gemini 3 Flash &  & 0.02 & 0.02 & 0.12 & 0.94 & \textbf{0.00} \\
Gemini 3 Flash & \cmark & \textbf{0.01} & \textbf{0.01} & \textbf{0.03} & 0.85 & \textbf{0.00} & Gemini 3 Flash & \cmark & 0.03 & 0.04 & 0.16 & 0.92 & 0.18 \\
Gemini 3.1 Pro &  & 0.02 & 0.02 & 0.06 & 0.96 & \textbf{0.00} & Gemini 3.1 Pro &  & \textbf{0.02} & \textbf{0.01} & \textbf{0.09} & 0.94 & \textbf{0.00} \\
Gemini 3.1 Pro & \cmark & 0.01 & 0.02 & 0.05 & \textbf{0.97} & \textbf{0.00} & Gemini 3.1 Pro & \cmark & 0.02 & 0.02 & 0.12 & \textbf{0.96} & \textbf{0.00} \\
\addlinespace[0.35em]
\midrule
\addlinespace[0.35em]

\multicolumn{7}{>{\columncolor{green!12}}l}{\textbf{Georgian-Russian}} & \multicolumn{7}{>{\columncolor{green!12}}l}{\textbf{Gojri-English-Hindi}} \\
\multicolumn{7}{>{\columncolor{gray!12}}l}{\emph{OCR systems}} & \multicolumn{7}{>{\columncolor{gray!12}}l}{\emph{OCR systems}} \\
GLM-OCR &  & 0.65 & 0.72 & 0.98 & 0.00 & 0.65 & GLM-OCR &  & 0.16 & 0.13 & 0.63 & 0.00 & 0.05 \\
GLM-OCR & \cmark & 0.88 & 1.29 & 1.59 & 0.00 & 0.51 & GLM-OCR & \cmark & 0.24 & 0.22 & 0.67 & 0.00 & 0.12 \\
Mathpix &  & 0.37 & 0.40 & 0.84 & 0.00 & 0.64 & Mathpix &  & 0.26 & 0.00 & 0.03 & 0.00 & 0.03 \\
MinerU2.5-Pro &  & 0.52 & 3.55 & 4.90 & 0.00 & 0.63 & MinerU2.5-Pro &  & 0.01 & 0.01 & 0.14 & 0.00 & \textbf{0.00} \\
PaddleOCR-VL-1.5 &  & 0.59 & 0.64 & 0.92 & 0.00 & 0.57 & PaddleOCR-VL-1.5 &  & 0.10 & 0.07 & 0.35 & 0.00 & 0.05 \\
\multicolumn{7}{>{\columncolor{gray!12}}l}{\emph{Vision Language Models}} & \multicolumn{7}{>{\columncolor{gray!12}}l}{\emph{Vision Language Models}} \\
Qwen3-VL-235B &  & 0.42 & 0.42 & 0.75 & 0.00 & 0.12 & Qwen3-VL-235B &  & 0.01 & 0.01 & 0.13 & 0.00 & \textbf{0.00} \\
Qwen3-VL-235B & \cmark & 0.33 & 0.34 & 0.71 & 0.00 & 0.11 & Qwen3-VL-235B & \cmark & 0.01 & 0.01 & 0.13 & 0.00 & \textbf{0.00} \\
\multicolumn{7}{>{\columncolor{gray!12}}l}{\emph{General-purpose LLMs}} & \multicolumn{7}{>{\columncolor{gray!12}}l}{\emph{General-purpose LLMs}} \\
Claude Opus 4.7 &  & 0.09 & 0.14 & 0.34 & 0.53 & 0.49 & Claude Opus 4.7 &  & \textbf{0.00} & \textbf{0.00} & \textbf{0.01} & 0.00 & \textbf{0.00} \\
Claude Opus 4.7 & \cmark & 0.05 & 0.05 & 0.25 & 0.24 & 0.22 & Claude Opus 4.7 & \cmark & 0.00 & 0.00 & 0.02 & 0.00 & \textbf{0.00} \\
GPT-5.5 &  & 0.09 & 0.09 & 0.28 & 0.00 & 0.07 & GPT-5.5 &  & 0.01 & 0.01 & 0.11 & 0.00 & \textbf{0.00} \\
GPT-5.5 & \cmark & 0.11 & 0.10 & 0.37 & 0.10 & 0.09 & GPT-5.5 & \cmark & 0.01 & 0.01 & 0.12 & 0.00 & \textbf{0.00} \\
Gemini 3 Flash &  & \textbf{0.01} & \textbf{0.01} & \textbf{0.06} & 0.91 & \textbf{0.00} & Gemini 3 Flash &  & 0.01 & 0.01 & 0.05 & \textbf{0.79} & \textbf{0.00} \\
Gemini 3 Flash & \cmark & 0.03 & 0.03 & 0.18 & \textbf{0.91} & \textbf{0.00} & Gemini 3 Flash & \cmark & 0.00 & 0.00 & 0.04 & \textbf{0.79} & \textbf{0.00} \\
Gemini 3.1 Pro &  & 0.04 & 0.03 & 0.23 & 0.91 & 0.04 & Gemini 3.1 Pro &  & 0.00 & 0.00 & 0.03 & 0.00 & \textbf{0.00} \\
Gemini 3.1 Pro & \cmark & 0.04 & 0.03 & 0.25 & 0.52 & 0.06 & Gemini 3.1 Pro & \cmark & 0.00 & 0.00 & 0.03 & 0.00 & \textbf{0.00} \\
\addlinespace[0.35em]
\midrule
\addlinespace[0.35em]

\multicolumn{7}{>{\columncolor{green!12}}l}{\textbf{Greek-English}} & \multicolumn{7}{>{\columncolor{green!12}}l}{\textbf{Gujarati-English}} \\
\multicolumn{7}{>{\columncolor{gray!12}}l}{\emph{OCR systems}} & \multicolumn{7}{>{\columncolor{gray!12}}l}{\emph{OCR systems}} \\
GLM-OCR &  & 0.14 & 0.13 & 0.33 & 0.00 & 0.64 & GLM-OCR &  & 0.39 & 0.33 & 0.47 & 0.00 & 0.41 \\
GLM-OCR & \cmark & 0.14 & 0.13 & 0.34 & 0.00 & 0.64 & GLM-OCR & \cmark & 0.92 & 0.89 & 0.94 & 0.00 & 0.49 \\
Mathpix &  & 0.13 & 0.14 & 0.32 & 0.00 & 0.62 & Mathpix &  & 0.10 & 0.08 & 0.24 & 0.00 & 0.26 \\
MinerU2.5-Pro &  & 0.10 & 0.11 & 0.25 & 0.00 & 0.57 & MinerU2.5-Pro &  & 0.29 & 0.24 & 0.40 & 0.00 & 0.27 \\
PaddleOCR-VL-1.5 &  & 0.13 & 0.13 & 0.29 & 0.00 & 0.59 & PaddleOCR-VL-1.5 &  & 0.36 & 0.32 & 0.52 & 0.00 & 0.42 \\
\multicolumn{7}{>{\columncolor{gray!12}}l}{\emph{Vision Language Models}} & \multicolumn{7}{>{\columncolor{gray!12}}l}{\emph{Vision Language Models}} \\
Qwen3-VL-235B &  & 0.02 & 0.03 & 0.09 & 0.00 & \textbf{0.00} & Qwen3-VL-235B &  & 0.08 & 0.13 & 0.24 & 0.00 & 0.11 \\
Qwen3-VL-235B & \cmark & 0.02 & 0.03 & 0.09 & 0.00 & \textbf{0.00} & Qwen3-VL-235B & \cmark & 0.09 & 0.13 & 0.25 & 0.00 & 0.25 \\
\multicolumn{7}{>{\columncolor{gray!12}}l}{\emph{General-purpose LLMs}} & \multicolumn{7}{>{\columncolor{gray!12}}l}{\emph{General-purpose LLMs}} \\
Claude Opus 4.7 &  & 0.02 & 0.03 & 0.07 & 0.31 & \textbf{0.00} & Claude Opus 4.7 &  & 0.02 & 0.04 & 0.14 & 0.62 & \textbf{0.00} \\
Claude Opus 4.7 & \cmark & 0.05 & 0.07 & 0.14 & 0.00 & 0.42 & Claude Opus 4.7 & \cmark & 0.02 & 0.03 & 0.14 & 0.54 & \textbf{0.00} \\
GPT-5.5 &  & 0.02 & 0.03 & 0.08 & 0.00 & \textbf{0.00} & GPT-5.5 &  & 0.03 & 0.03 & 0.16 & 0.25 & \textbf{0.00} \\
GPT-5.5 & \cmark & 0.01 & 0.01 & 0.04 & 0.30 & \textbf{0.00} & GPT-5.5 & \cmark & 0.03 & 0.03 & 0.16 & 0.00 & \textbf{0.00} \\
Gemini 3 Flash &  & 0.00 & 0.00 & 0.02 & 0.79 & \textbf{0.00} & Gemini 3 Flash &  & \textbf{0.02} & \textbf{0.02} & 0.12 & 0.77 & \textbf{0.00} \\
Gemini 3 Flash & \cmark & \textbf{0.00} & \textbf{0.00} & \textbf{0.02} & \textbf{0.87} & \textbf{0.00} & Gemini 3 Flash & \cmark & 0.02 & 0.03 & 0.18 & 0.83 & \textbf{0.00} \\
Gemini 3.1 Pro &  & 0.01 & 0.03 & 0.05 & 0.53 & \textbf{0.00} & Gemini 3.1 Pro &  & 0.02 & 0.03 & \textbf{0.11} & 0.86 & \textbf{0.00} \\
Gemini 3.1 Pro & \cmark & 0.01 & 0.02 & 0.06 & 0.61 & \textbf{0.00} & Gemini 3.1 Pro & \cmark & 0.02 & 0.04 & 0.15 & \textbf{0.87} & \textbf{0.00} \\
\bottomrule
\end{tabular}
\end{adjustbox}
\caption{Stage 1 dictionary-specific evaluation results grouped by dictionary (continued).}
\end{table*}

\clearpage

\begin{table*}[p]
\centering
\scriptsize
\setlength{\tabcolsep}{2.2pt}
\renewcommand{\arraystretch}{1.08}
\begin{adjustbox}{max width=\textwidth}
\begin{tabular}{lcrrrrr@{\qquad}lcrrrrr}
\toprule
\textbf{Model} & \textbf{Alph.} & \textbf{Edit} & \textbf{GCER} & \textbf{WER} & \textbf{Mrk. F1} & \textbf{RO} & \textbf{Model} & \textbf{Alph.} & \textbf{Edit} & \textbf{GCER} & \textbf{WER} & \textbf{Mrk. F1} & \textbf{RO} \\
\midrule

\multicolumn{7}{>{\columncolor{green!12}}l}{\textbf{Iñupiatun Eskimo-English}} & \multicolumn{7}{>{\columncolor{green!12}}l}{\textbf{Japanese-English}} \\
\multicolumn{7}{>{\columncolor{gray!12}}l}{\emph{OCR systems}} & \multicolumn{7}{>{\columncolor{gray!12}}l}{\emph{OCR systems}} \\
GLM-OCR &  & 0.29 & 0.24 & 0.27 & 0.00 & 0.65 & GLM-OCR &  & 0.13 & 0.08 & 0.20 & 0.00 & 0.55 \\
GLM-OCR & \cmark & 0.11 & 0.12 & 0.15 & 0.00 & 0.63 & GLM-OCR & \cmark & 0.13 & 0.09 & 0.21 & 0.00 & 0.57 \\
Mathpix &  & 0.10 & 0.12 & 0.14 & 0.00 & 0.67 & Mathpix &  & 0.36 & 0.27 & 0.70 & 0.00 & 0.47 \\
MinerU2.5-Pro &  & 0.08 & 0.08 & 0.12 & 0.00 & 0.43 & MinerU2.5-Pro &  & 0.07 & 0.07 & 0.17 & 0.00 & 0.60 \\
PaddleOCR-VL-1.5 &  & 0.09 & 0.08 & 0.13 & 0.00 & 0.58 & PaddleOCR-VL-1.5 &  & 0.16 & 0.11 & 0.23 & 0.00 & 0.46 \\
\multicolumn{7}{>{\columncolor{gray!12}}l}{\emph{Vision Language Models}} & \multicolumn{7}{>{\columncolor{gray!12}}l}{\emph{Vision Language Models}} \\
Qwen3-VL-235B &  & 0.01 & 0.01 & 0.03 & 0.00 & \textbf{0.00} & Qwen3-VL-235B &  & 0.01 & 0.02 & 0.06 & 0.00 & 0.01 \\
Qwen3-VL-235B & \cmark & 0.01 & 0.01 & 0.03 & 0.01 & \textbf{0.00} & Qwen3-VL-235B & \cmark & 0.02 & 0.04 & 0.06 & 0.00 & 0.12 \\
\multicolumn{7}{>{\columncolor{gray!12}}l}{\emph{General-purpose LLMs}} & \multicolumn{7}{>{\columncolor{gray!12}}l}{\emph{General-purpose LLMs}} \\
Claude Opus 4.7 &  & \textbf{0.00} & \textbf{0.00} & \textbf{0.01} & 0.99 & \textbf{0.00} & Claude Opus 4.7 &  & 0.09 & 0.07 & 0.10 & 0.78 & 0.58 \\
Claude Opus 4.7 & \cmark & 0.03 & 0.05 & 0.04 & 0.95 & 0.32 & Claude Opus 4.7 & \cmark & 0.03 & 0.03 & 0.06 & 0.78 & 0.18 \\
GPT-5.5 &  & 0.01 & 0.01 & 0.04 & \textbf{1.00} & \textbf{0.00} & GPT-5.5 &  & 0.03 & 0.02 & 0.06 & 0.80 & 0.02 \\
GPT-5.5 & \cmark & 0.01 & 0.01 & 0.04 & \textbf{1.00} & \textbf{0.00} & GPT-5.5 & \cmark & 0.02 & 0.02 & 0.06 & 0.80 & 0.02 \\
Gemini 3 Flash &  & 0.00 & 0.00 & 0.02 & 0.99 & \textbf{0.00} & Gemini 3 Flash &  & 0.02 & 0.01 & \textbf{0.03} & 0.86 & 0.02 \\
Gemini 3 Flash & \cmark & 0.01 & 0.00 & 0.02 & 0.90 & \textbf{0.00} & Gemini 3 Flash & \cmark & 0.02 & 0.02 & 0.04 & \textbf{0.91} & 0.01 \\
Gemini 3.1 Pro &  & 0.00 & 0.00 & 0.01 & 0.81 & \textbf{0.00} & Gemini 3.1 Pro &  & \textbf{0.01} & \textbf{0.01} & 0.04 & 0.80 & 0.01 \\
Gemini 3.1 Pro & \cmark & 0.00 & 0.00 & 0.01 & 1.00 & \textbf{0.00} & Gemini 3.1 Pro & \cmark & 0.02 & 0.01 & 0.04 & 0.80 & \textbf{0.00} \\
\addlinespace[0.35em]
\midrule
\addlinespace[0.35em]

\multicolumn{7}{>{\columncolor{green!12}}l}{\textbf{Kashmiri-English}} & \multicolumn{7}{>{\columncolor{green!12}}l}{\textbf{Khmer-English}} \\
\multicolumn{7}{>{\columncolor{gray!12}}l}{\emph{OCR systems}} & \multicolumn{7}{>{\columncolor{gray!12}}l}{\emph{OCR systems}} \\
GLM-OCR &  & 0.44 & 0.50 & 0.59 & 0.00 & 0.57 & GLM-OCR &  & 0.43 & 0.38 & 0.42 & 0.00 & 0.49 \\
GLM-OCR & \cmark & 0.60 & 0.67 & 0.75 & 0.00 & 0.68 & GLM-OCR & \cmark & 0.44 & 0.40 & 0.43 & 0.00 & 0.49 \\
Mathpix &  & 0.53 & 0.59 & 0.78 & 0.00 & 0.75 & Mathpix &  & 0.75 & 0.70 & 0.81 & 0.00 & 0.82 \\
MinerU2.5-Pro &  & 0.95 & 1.04 & 1.10 & 0.00 & 0.05 & MinerU2.5-Pro &  & 0.41 & 1.69 & 2.17 & 0.00 & 0.47 \\
PaddleOCR-VL-1.5 &  & 0.74 & 0.86 & 0.96 & 0.00 & 0.26 & PaddleOCR-VL-1.5 &  & 0.50 & 0.53 & 0.55 & 0.00 & 0.29 \\
\multicolumn{7}{>{\columncolor{gray!12}}l}{\emph{Vision Language Models}} & \multicolumn{7}{>{\columncolor{gray!12}}l}{\emph{Vision Language Models}} \\
Qwen3-VL-235B &  & 0.19 & 0.34 & 0.51 & 0.00 & 0.70 & Qwen3-VL-235B &  & 0.21 & 0.20 & 0.24 & 0.00 & 0.28 \\
Qwen3-VL-235B & \cmark & 0.22 & 0.31 & 0.46 & 0.00 & 0.70 & Qwen3-VL-235B & \cmark & 0.30 & 0.31 & 0.34 & 0.00 & 0.42 \\
\multicolumn{7}{>{\columncolor{gray!12}}l}{\emph{General-purpose LLMs}} & \multicolumn{7}{>{\columncolor{gray!12}}l}{\emph{General-purpose LLMs}} \\
Claude Opus 4.7 &  & 0.13 & 0.16 & 0.29 & 0.35 & 0.43 & Claude Opus 4.7 &  & 0.07 & 0.07 & 0.13 & 0.00 & 0.21 \\
Claude Opus 4.7 & \cmark & 0.05 & 0.07 & 0.24 & 0.35 & 0.01 & Claude Opus 4.7 & \cmark & 0.09 & 0.09 & 0.15 & 0.00 & 0.54 \\
GPT-5.5 &  & 0.11 & 0.14 & 0.30 & 0.16 & \textbf{0.00} & GPT-5.5 &  & 0.05 & 0.05 & 0.09 & 0.00 & 0.02 \\
GPT-5.5 & \cmark & 0.05 & 0.06 & 0.27 & 0.16 & \textbf{0.00} & GPT-5.5 & \cmark & 0.05 & 0.04 & 0.09 & 0.00 & \textbf{0.01} \\
Gemini 3 Flash &  & 0.11 & 0.15 & 0.31 & \textbf{0.84} & 0.34 & Gemini 3 Flash &  & 0.06 & 0.07 & 0.11 & 0.59 & 0.03 \\
Gemini 3 Flash & \cmark & 0.05 & 0.06 & 0.24 & 0.43 & \textbf{0.00} & Gemini 3 Flash & \cmark & 0.07 & 0.06 & 0.16 & \textbf{0.61} & 0.17 \\
Gemini 3.1 Pro &  & \textbf{0.05} & \textbf{0.05} & \textbf{0.22} & 0.17 & \textbf{0.00} & Gemini 3.1 Pro &  & \textbf{0.02} & \textbf{0.02} & \textbf{0.05} & 0.00 & 0.17 \\
Gemini 3.1 Pro & \cmark & 0.14 & 0.26 & 0.45 & 0.20 & 0.88 & Gemini 3.1 Pro & \cmark & 0.06 & 0.10 & 0.16 & 0.00 & 0.02 \\
\addlinespace[0.35em]
\midrule
\addlinespace[0.35em]

\multicolumn{7}{>{\columncolor{green!12}}l}{\textbf{Malay-English}} & \multicolumn{7}{>{\columncolor{green!12}}l}{\textbf{Na-English-Chinese-French}} \\
\multicolumn{7}{>{\columncolor{gray!12}}l}{\emph{OCR systems}} & \multicolumn{7}{>{\columncolor{gray!12}}l}{\emph{OCR systems}} \\
GLM-OCR &  & 0.43 & 0.41 & 0.48 & 0.00 & 0.39 & GLM-OCR &  & 0.57 & 0.30 & 0.42 & 0.00 & 0.33 \\
GLM-OCR & \cmark & 0.65 & 4.11 & 4.50 & 0.00 & 0.59 & GLM-OCR & \cmark & 0.60 & 2.04 & 2.40 & 0.00 & 0.37 \\
Mathpix &  & 0.39 & 0.37 & 0.54 & 0.00 & 0.39 & Mathpix &  & 0.68 & 0.51 & 0.79 & 0.00 & 0.34 \\
MinerU2.5-Pro &  & 0.20 & 0.20 & 0.41 & 0.00 & 0.22 & MinerU2.5-Pro &  & 0.50 & 0.27 & 0.45 & 0.00 & 0.37 \\
PaddleOCR-VL-1.5 &  & 0.38 & 0.38 & 0.57 & 0.00 & 0.35 & PaddleOCR-VL-1.5 &  & 0.40 & 0.28 & 0.37 & 0.00 & 0.35 \\
\multicolumn{7}{>{\columncolor{gray!12}}l}{\emph{Vision Language Models}} & \multicolumn{7}{>{\columncolor{gray!12}}l}{\emph{Vision Language Models}} \\
Qwen3-VL-235B &  & 0.14 & 0.16 & 0.32 & 0.00 & 0.62 & Qwen3-VL-235B &  & 0.15 & 0.14 & 0.25 & 0.00 & 0.22 \\
Qwen3-VL-235B & \cmark & 0.16 & 0.18 & 0.36 & 0.00 & 0.62 & Qwen3-VL-235B & \cmark & 0.14 & 0.12 & 0.22 & 0.00 & 0.17 \\
\multicolumn{7}{>{\columncolor{gray!12}}l}{\emph{General-purpose LLMs}} & \multicolumn{7}{>{\columncolor{gray!12}}l}{\emph{General-purpose LLMs}} \\
Claude Opus 4.7 &  & \textbf{0.03} & 0.05 & \textbf{0.11} & 0.00 & \textbf{0.00} & Claude Opus 4.7 &  & 0.06 & 0.06 & 0.17 & \textbf{0.74} & 0.15 \\
Claude Opus 4.7 & \cmark & 0.03 & 0.05 & 0.14 & 0.00 & \textbf{0.00} & Claude Opus 4.7 & \cmark & 0.09 & 0.08 & 0.21 & 0.67 & 0.18 \\
GPT-5.5 &  & 0.06 & 0.06 & 0.23 & 0.00 & \textbf{0.00} & GPT-5.5 &  & 0.07 & 0.06 & 0.14 & 0.00 & 0.01 \\
GPT-5.5 & \cmark & 0.06 & 0.06 & 0.22 & 0.00 & \textbf{0.00} & GPT-5.5 & \cmark & 0.06 & 0.05 & 0.12 & 0.00 & \textbf{0.00} \\
Gemini 3 Flash &  & 0.13 & 0.24 & 0.19 & \textbf{0.76} & \textbf{0.00} & Gemini 3 Flash &  & 0.01 & 0.01 & 0.05 & 0.22 & 0.03 \\
Gemini 3 Flash & \cmark & 0.05 & 0.06 & 0.17 & 0.64 & 0.31 & Gemini 3 Flash & \cmark & 0.01 & \textbf{0.01} & \textbf{0.03} & 0.16 & \textbf{0.00} \\
Gemini 3.1 Pro &  & 0.04 & 0.05 & 0.19 & 0.00 & \textbf{0.00} & Gemini 3.1 Pro &  & \textbf{0.01} & 0.01 & 0.03 & 0.28 & \textbf{0.00} \\
Gemini 3.1 Pro & \cmark & 0.03 & \textbf{0.04} & 0.17 & 0.00 & \textbf{0.00} & Gemini 3.1 Pro & \cmark & 0.01 & 0.01 & 0.05 & 0.07 & \textbf{0.00} \\
\bottomrule
\end{tabular}
\end{adjustbox}
\caption{Stage 1 dictionary-specific evaluation results grouped by dictionary (continued).}
\end{table*}

\clearpage

\begin{table*}[p]
\centering
\scriptsize
\setlength{\tabcolsep}{2.2pt}
\renewcommand{\arraystretch}{1.08}
\begin{adjustbox}{max width=\textwidth}
\begin{tabular}{lcrrrrr@{\qquad}lcrrrrr}
\toprule
\textbf{Model} & \textbf{Alph.} & \textbf{Edit} & \textbf{GCER} & \textbf{WER} & \textbf{Mrk. F1} & \textbf{RO} & \textbf{Model} & \textbf{Alph.} & \textbf{Edit} & \textbf{GCER} & \textbf{WER} & \textbf{Mrk. F1} & \textbf{RO} \\
\midrule

\multicolumn{7}{>{\columncolor{green!12}}l}{\textbf{Nahuatl-French}} & \multicolumn{7}{>{\columncolor{green!12}}l}{\textbf{Punjabi-English}} \\
\multicolumn{7}{>{\columncolor{gray!12}}l}{\emph{OCR systems}} & \multicolumn{7}{>{\columncolor{gray!12}}l}{\emph{OCR systems}} \\
GLM-OCR &  & 0.30 & 0.22 & 0.35 & 0.00 & 0.62 & GLM-OCR &  & 0.50 & 0.42 & 0.58 & 0.00 & 0.69 \\
GLM-OCR & \cmark & 0.44 & 1.66 & 1.78 & 0.00 & 0.60 & GLM-OCR & \cmark & 0.90 & 1.45 & 1.69 & 0.00 & 0.56 \\
Mathpix &  & 0.24 & 0.19 & 0.28 & 0.00 & 0.63 & Mathpix &  & 0.55 & 0.49 & 0.67 & 0.00 & 0.66 \\
MinerU2.5-Pro &  & 0.20 & 0.17 & 0.26 & 0.00 & 0.60 & MinerU2.5-Pro &  & 0.33 & 0.30 & 0.48 & 0.00 & 0.58 \\
PaddleOCR-VL-1.5 &  & 0.17 & 0.14 & 0.25 & 0.00 & 0.60 & PaddleOCR-VL-1.5 &  & 0.39 & 0.34 & 0.49 & 0.00 & 0.57 \\
\multicolumn{7}{>{\columncolor{gray!12}}l}{\emph{Vision Language Models}} & \multicolumn{7}{>{\columncolor{gray!12}}l}{\emph{Vision Language Models}} \\
Qwen3-VL-235B &  & 0.03 & 0.03 & 0.12 & 0.00 & 0.19 & Qwen3-VL-235B &  & 0.11 & 0.11 & 0.22 & 0.00 & 0.15 \\
Qwen3-VL-235B & \cmark & 0.02 & 0.02 & 0.10 & 0.00 & \textbf{0.00} & Qwen3-VL-235B & \cmark & 0.08 & 0.07 & 0.18 & 0.00 & 0.14 \\
\multicolumn{7}{>{\columncolor{gray!12}}l}{\emph{General-purpose LLMs}} & \multicolumn{7}{>{\columncolor{gray!12}}l}{\emph{General-purpose LLMs}} \\
Claude Opus 4.7 &  & 0.01 & 0.01 & 0.03 & 0.97 & \textbf{0.00} & Claude Opus 4.7 &  & 0.05 & 0.05 & 0.13 & 0.65 & 0.14 \\
Claude Opus 4.7 & \cmark & 0.00 & 0.01 & 0.02 & \textbf{0.98} & \textbf{0.00} & Claude Opus 4.7 & \cmark & 0.07 & 0.08 & 0.17 & 0.84 & 0.22 \\
GPT-5.5 &  & 0.07 & 0.11 & 0.12 & 0.97 & \textbf{0.00} & GPT-5.5 &  & 0.04 & 0.04 & 0.13 & 0.00 & 0.14 \\
GPT-5.5 & \cmark & 0.04 & 0.06 & 0.10 & 0.93 & \textbf{0.00} & GPT-5.5 & \cmark & 0.04 & 0.03 & 0.14 & 0.49 & \textbf{0.00} \\
Gemini 3 Flash &  & 0.02 & 0.02 & 0.03 & 0.97 & \textbf{0.00} & Gemini 3 Flash &  & 0.04 & \textbf{0.03} & \textbf{0.12} & \textbf{0.97} & 0.14 \\
Gemini 3 Flash & \cmark & \textbf{0.00} & \textbf{0.00} & \textbf{0.02} & 0.96 & \textbf{0.00} & Gemini 3 Flash & \cmark & 0.04 & 0.03 & 0.13 & 0.79 & 0.14 \\
Gemini 3.1 Pro &  & 0.00 & 0.00 & 0.02 & 0.96 & \textbf{0.00} & Gemini 3.1 Pro &  & 0.04 & 0.04 & 0.14 & 0.82 & 0.08 \\
Gemini 3.1 Pro & \cmark & 0.00 & 0.00 & 0.02 & 0.96 & \textbf{0.00} & Gemini 3.1 Pro & \cmark & \textbf{0.04} & 0.04 & 0.13 & 0.88 & 0.08 \\
\addlinespace[0.35em]
\midrule
\addlinespace[0.35em]

\multicolumn{7}{>{\columncolor{green!12}}l}{\textbf{Reel-English}} & \multicolumn{7}{>{\columncolor{green!12}}l}{\textbf{Ritharngu-English}} \\
\multicolumn{7}{>{\columncolor{gray!12}}l}{\emph{OCR systems}} & \multicolumn{7}{>{\columncolor{gray!12}}l}{\emph{OCR systems}} \\
GLM-OCR &  & 0.27 & 0.23 & 0.34 & 0.10 & 0.51 & GLM-OCR &  & 0.11 & 0.07 & 0.18 & 0.00 & 0.51 \\
GLM-OCR & \cmark & 0.39 & 0.53 & 0.56 & 0.00 & 0.46 & GLM-OCR & \cmark & 0.10 & 0.07 & 0.19 & 0.00 & 0.45 \\
Mathpix &  & 0.30 & 0.27 & 0.49 & 0.00 & 0.50 & Mathpix &  & 0.09 & 0.06 & 0.18 & 0.00 & 0.44 \\
MinerU2.5-Pro &  & 0.21 & 0.19 & 0.40 & 0.02 & 0.27 & MinerU2.5-Pro &  & 0.08 & 0.06 & 0.18 & 0.00 & 0.32 \\
PaddleOCR-VL-1.5 &  & 0.21 & 0.19 & 0.40 & 0.00 & 0.36 & PaddleOCR-VL-1.5 &  & 0.29 & 0.19 & 0.29 & 0.00 & 0.58 \\
\multicolumn{7}{>{\columncolor{gray!12}}l}{\emph{Vision Language Models}} & \multicolumn{7}{>{\columncolor{gray!12}}l}{\emph{Vision Language Models}} \\
Qwen3-VL-235B &  & 0.09 & 0.09 & 0.23 & 0.00 & 0.07 & Qwen3-VL-235B &  & 0.02 & 0.02 & 0.09 & 0.00 & 0.20 \\
Qwen3-VL-235B & \cmark & 0.09 & 0.09 & 0.24 & 0.00 & 0.06 & Qwen3-VL-235B & \cmark & 0.04 & 0.04 & 0.11 & 0.00 & 0.38 \\
\multicolumn{7}{>{\columncolor{gray!12}}l}{\emph{General-purpose LLMs}} & \multicolumn{7}{>{\columncolor{gray!12}}l}{\emph{General-purpose LLMs}} \\
Claude Opus 4.7 &  & 0.05 & 0.05 & 0.15 & 0.95 & 0.06 & Claude Opus 4.7 &  & \textbf{0.01} & \textbf{0.01} & \textbf{0.05} & 0.54 & \textbf{0.00} \\
Claude Opus 4.7 & \cmark & 0.06 & 0.06 & 0.17 & 0.95 & 0.15 & Claude Opus 4.7 & \cmark & 0.01 & 0.01 & 0.07 & 0.62 & \textbf{0.00} \\
GPT-5.5 &  & 0.04 & 0.04 & 0.14 & 0.98 & \textbf{0.00} & GPT-5.5 &  & 0.01 & 0.01 & 0.07 & 0.70 & \textbf{0.00} \\
GPT-5.5 & \cmark & 0.06 & 0.06 & 0.18 & 0.96 & \textbf{0.00} & GPT-5.5 & \cmark & 0.02 & 0.02 & 0.10 & 0.71 & \textbf{0.00} \\
Gemini 3 Flash &  & \textbf{0.00} & \textbf{0.00} & \textbf{0.00} & \textbf{0.99} & \textbf{0.00} & Gemini 3 Flash &  & 0.02 & 0.02 & 0.08 & 0.80 & \textbf{0.00} \\
Gemini 3 Flash & \cmark & \textbf{0.00} & \textbf{0.00} & \textbf{0.00} & 0.99 & \textbf{0.00} & Gemini 3 Flash & \cmark & 0.03 & 0.03 & 0.12 & \textbf{0.96} & \textbf{0.00} \\
Gemini 3.1 Pro &  & 0.00 & 0.00 & 0.00 & 0.93 & \textbf{0.00} & Gemini 3.1 Pro &  & 0.02 & 0.02 & 0.09 & 0.86 & \textbf{0.00} \\
Gemini 3.1 Pro & \cmark & \textbf{0.00} & \textbf{0.00} & \textbf{0.00} & 0.98 & \textbf{0.00} & Gemini 3.1 Pro & \cmark & 0.02 & 0.02 & 0.11 & 0.75 & \textbf{0.00} \\
\addlinespace[0.35em]
\midrule
\addlinespace[0.35em]

\multicolumn{7}{>{\columncolor{green!12}}l}{\textbf{Sanskrit-English}} & \multicolumn{7}{>{\columncolor{green!12}}l}{\textbf{Shilluk-English}} \\
\multicolumn{7}{>{\columncolor{gray!12}}l}{\emph{OCR systems}} & \multicolumn{7}{>{\columncolor{gray!12}}l}{\emph{OCR systems}} \\
GLM-OCR &  & 0.40 & 0.39 & 0.43 & 0.00 & 0.47 & GLM-OCR &  & 0.14 & 0.10 & 0.11 & 0.00 & 0.22 \\
GLM-OCR & \cmark & 0.85 & 0.79 & 0.78 & 0.00 & 0.68 & GLM-OCR & \cmark & 0.14 & 0.10 & 0.10 & 0.00 & 0.22 \\
Mathpix &  & 0.20 & 0.17 & 0.36 & 0.00 & 0.30 & Mathpix &  & 0.14 & 0.10 & 0.19 & 0.00 & 0.22 \\
MinerU2.5-Pro &  & 0.20 & 0.19 & 0.30 & 0.00 & 0.27 & MinerU2.5-Pro &  & 0.05 & 0.06 & 0.09 & 0.01 & 0.16 \\
PaddleOCR-VL-1.5 &  & 0.30 & 1.00 & 0.40 & 0.00 & 0.25 & PaddleOCR-VL-1.5 &  & 0.18 & 0.14 & 0.18 & 0.00 & 0.25 \\
\multicolumn{7}{>{\columncolor{gray!12}}l}{\emph{Vision Language Models}} & \multicolumn{7}{>{\columncolor{gray!12}}l}{\emph{Vision Language Models}} \\
Qwen3-VL-235B &  & 0.13 & 0.16 & 0.21 & 0.00 & \textbf{0.00} & Qwen3-VL-235B &  & 0.01 & 0.01 & 0.02 & 0.00 & \textbf{0.00} \\
Qwen3-VL-235B & \cmark & 0.12 & 0.14 & 0.22 & 0.00 & 0.05 & Qwen3-VL-235B & \cmark & 0.01 & 0.01 & 0.02 & 0.00 & \textbf{0.00} \\
\multicolumn{7}{>{\columncolor{gray!12}}l}{\emph{General-purpose LLMs}} & \multicolumn{7}{>{\columncolor{gray!12}}l}{\emph{General-purpose LLMs}} \\
Claude Opus 4.7 &  & 0.05 & 0.13 & 0.18 & 0.55 & \textbf{0.00} & Claude Opus 4.7 &  & 0.04 & 0.06 & 0.06 & 0.83 & 0.16 \\
Claude Opus 4.7 & \cmark & 0.03 & 0.03 & 0.11 & 0.55 & \textbf{0.00} & Claude Opus 4.7 & \cmark & 0.03 & 0.05 & 0.06 & 0.83 & 0.16 \\
GPT-5.5 &  & 0.12 & 0.12 & 0.16 & 0.04 & \textbf{0.00} & GPT-5.5 &  & 0.01 & 0.01 & 0.02 & 0.87 & \textbf{0.00} \\
GPT-5.5 & \cmark & 0.07 & 0.09 & 0.15 & 0.07 & \textbf{0.00} & GPT-5.5 & \cmark & 0.01 & 0.01 & 0.03 & 0.84 & \textbf{0.00} \\
Gemini 3 Flash &  & 0.04 & 0.05 & \textbf{0.10} & 0.73 & \textbf{0.00} & Gemini 3 Flash &  & 0.01 & 0.00 & 0.01 & 0.97 & \textbf{0.00} \\
Gemini 3 Flash & \cmark & 0.03 & 0.05 & 0.11 & 0.72 & \textbf{0.00} & Gemini 3 Flash & \cmark & 0.00 & \textbf{0.00} & 0.00 & 0.98 & 0.01 \\
Gemini 3.1 Pro &  & 0.03 & 0.04 & 0.11 & 0.63 & \textbf{0.00} & Gemini 3.1 Pro &  & \textbf{0.00} & \textbf{0.00} & \textbf{0.00} & \textbf{0.98} & \textbf{0.00} \\
Gemini 3.1 Pro & \cmark & \textbf{0.03} & \textbf{0.03} & 0.10 & \textbf{0.79} & \textbf{0.00} & Gemini 3.1 Pro & \cmark & 0.01 & 0.00 & 0.01 & 0.95 & \textbf{0.00} \\
\bottomrule
\end{tabular}
\end{adjustbox}
\caption{Stage 1 dictionary-specific evaluation results grouped by dictionary (continued).}
\end{table*}

\clearpage

\begin{table*}[p]
\centering
\scriptsize
\setlength{\tabcolsep}{2.2pt}
\renewcommand{\arraystretch}{1.08}
\begin{adjustbox}{max width=\textwidth}
\begin{tabular}{lcrrrrr@{\qquad}lcrrrrr}
\toprule
\textbf{Model} & \textbf{Alph.} & \textbf{Edit} & \textbf{GCER} & \textbf{WER} & \textbf{Mrk. F1} & \textbf{RO} & \textbf{Model} & \textbf{Alph.} & \textbf{Edit} & \textbf{GCER} & \textbf{WER} & \textbf{Mrk. F1} & \textbf{RO} \\
\midrule

\multicolumn{7}{>{\columncolor{green!12}}l}{\textbf{Syriac-English}} & \multicolumn{7}{>{\columncolor{green!12}}l}{\textbf{Telugu-English}} \\
\multicolumn{7}{>{\columncolor{gray!12}}l}{\emph{OCR systems}} & \multicolumn{7}{>{\columncolor{gray!12}}l}{\emph{OCR systems}} \\
GLM-OCR &  & 0.67 & 0.67 & 0.75 & 0.00 & 0.37 & GLM-OCR &  & 0.53 & 0.51 & 0.57 & 0.00 & 0.26 \\
GLM-OCR & \cmark & 0.76 & 0.77 & 0.81 & 0.00 & 0.57 & GLM-OCR & \cmark & 0.57 & 0.56 & 0.63 & 0.00 & 0.29 \\
Mathpix &  & 0.63 & 0.63 & 0.70 & 0.00 & 0.45 & Mathpix &  & 0.23 & 0.24 & 0.38 & 0.00 & 0.45 \\
MinerU2.5-Pro &  & 0.52 & 0.51 & 0.56 & 0.00 & 0.31 & MinerU2.5-Pro &  & 0.35 & 0.95 & 2.26 & 0.00 & 0.22 \\
PaddleOCR-VL-1.5 &  & 0.61 & 0.62 & 0.67 & 0.00 & 0.38 & PaddleOCR-VL-1.5 &  & 0.23 & 0.19 & 0.35 & 0.00 & 0.26 \\
\multicolumn{7}{>{\columncolor{gray!12}}l}{\emph{Vision Language Models}} & \multicolumn{7}{>{\columncolor{gray!12}}l}{\emph{Vision Language Models}} \\
Qwen3-VL-235B &  & 0.23 & 0.24 & 0.26 & 0.00 & 0.01 & Qwen3-VL-235B &  & 0.09 & 0.09 & 0.18 & 0.00 & 0.02 \\
Qwen3-VL-235B & \cmark & 0.23 & 0.25 & 0.26 & 0.00 & \textbf{0.00} & Qwen3-VL-235B & \cmark & 0.11 & 0.11 & 0.21 & 0.00 & 0.08 \\
\multicolumn{7}{>{\columncolor{gray!12}}l}{\emph{General-purpose LLMs}} & \multicolumn{7}{>{\columncolor{gray!12}}l}{\emph{General-purpose LLMs}} \\
Claude Opus 4.7 &  & 0.16 & 0.18 & 0.25 & 0.49 & 0.07 & Claude Opus 4.7 &  & 0.04 & 0.04 & 0.15 & 0.66 & \textbf{0.01} \\
Claude Opus 4.7 & \cmark & 0.15 & 0.16 & 0.25 & 0.55 & 0.02 & Claude Opus 4.7 & \cmark & 0.04 & 0.04 & 0.15 & 0.78 & \textbf{0.01} \\
GPT-5.5 &  & 0.17 & 0.17 & 0.23 & 0.30 & 0.02 & GPT-5.5 &  & 0.05 & 0.05 & 0.15 & 0.48 & 0.02 \\
GPT-5.5 & \cmark & 0.15 & 0.15 & 0.23 & 0.58 & 0.02 & GPT-5.5 & \cmark & 0.05 & 0.05 & 0.18 & 0.51 & 0.01 \\
Gemini 3 Flash &  & 0.19 & 0.20 & 0.26 & 0.46 & 0.02 & Gemini 3 Flash &  & \textbf{0.03} & \textbf{0.02} & 0.08 & 0.88 & 0.01 \\
Gemini 3 Flash & \cmark & 0.17 & 0.19 & 0.28 & 0.79 & 0.31 & Gemini 3 Flash & \cmark & 0.05 & 0.05 & 0.12 & 0.89 & 0.06 \\
Gemini 3.1 Pro &  & 0.16 & 0.16 & \textbf{0.23} & 0.66 & 0.04 & Gemini 3.1 Pro &  & 0.03 & 0.03 & \textbf{0.06} & \textbf{0.98} & 0.02 \\
Gemini 3.1 Pro & \cmark & \textbf{0.13} & \textbf{0.13} & \textbf{0.23} & \textbf{0.80} & 0.02 & Gemini 3.1 Pro & \cmark & 0.03 & 0.03 & 0.09 & 0.96 & 0.02 \\
\addlinespace[0.35em]
\midrule
\addlinespace[0.35em]

\multicolumn{7}{>{\columncolor{green!12}}l}{\textbf{Thai-Russian}} & \multicolumn{7}{>{\columncolor{green!12}}l}{\textbf{Tiri-English}} \\
\multicolumn{7}{>{\columncolor{gray!12}}l}{\emph{OCR systems}} & \multicolumn{7}{>{\columncolor{gray!12}}l}{\emph{OCR systems}} \\
GLM-OCR &  & 0.74 & 0.79 & 0.96 & 0.03 & 0.19 & GLM-OCR &  & 0.31 & 0.25 & 0.36 & 0.00 & 0.68 \\
GLM-OCR & \cmark & 0.97 & 0.96 & 1.00 & 0.00 & 0.82 & GLM-OCR & \cmark & 0.25 & 0.22 & 0.32 & 0.00 & 0.44 \\
Mathpix &  & 0.26 & 0.24 & 0.75 & 0.00 & 0.15 & Mathpix &  & 0.17 & 0.18 & 0.29 & 0.00 & 0.66 \\
MinerU2.5-Pro &  & 0.57 & 2.25 & 3.75 & 0.00 & 0.21 & MinerU2.5-Pro &  & 0.17 & 0.15 & 0.29 & 0.00 & 0.59 \\
PaddleOCR-VL-1.5 &  & 0.78 & 3.35 & 3.31 & 0.00 & 0.64 & PaddleOCR-VL-1.5 &  & 0.23 & 0.22 & 0.33 & 0.00 & 0.54 \\
\multicolumn{7}{>{\columncolor{gray!12}}l}{\emph{Vision Language Models}} & \multicolumn{7}{>{\columncolor{gray!12}}l}{\emph{Vision Language Models}} \\
Qwen3-VL-235B &  & 0.15 & 0.17 & 0.40 & 0.00 & \textbf{0.00} & Qwen3-VL-235B &  & 0.03 & 0.02 & 0.09 & 0.00 & \textbf{0.00} \\
Qwen3-VL-235B & \cmark & 0.16 & 0.18 & 0.47 & 0.00 & 0.11 & Qwen3-VL-235B & \cmark & 0.02 & 0.02 & 0.08 & 0.00 & \textbf{0.00} \\
\multicolumn{7}{>{\columncolor{gray!12}}l}{\emph{General-purpose LLMs}} & \multicolumn{7}{>{\columncolor{gray!12}}l}{\emph{General-purpose LLMs}} \\
Claude Opus 4.7 &  & 0.09 & 0.10 & 0.40 & 0.67 & \textbf{0.00} & Claude Opus 4.7 &  & 0.02 & 0.01 & 0.05 & 0.00 & \textbf{0.00} \\
Claude Opus 4.7 & \cmark & 0.10 & 0.12 & 0.43 & 0.65 & 0.11 & Claude Opus 4.7 & \cmark & \textbf{0.01} & \textbf{0.01} & \textbf{0.03} & 0.00 & \textbf{0.00} \\
GPT-5.5 &  & 0.09 & 0.11 & 0.31 & 0.19 & \textbf{0.00} & GPT-5.5 &  & 0.02 & 0.02 & 0.09 & 0.00 & \textbf{0.00} \\
GPT-5.5 & \cmark & 0.07 & 0.09 & 0.29 & 0.19 & 0.01 & GPT-5.5 & \cmark & 0.03 & 0.02 & 0.07 & 0.00 & 0.00 \\
Gemini 3 Flash &  & 0.07 & 0.08 & 0.17 & 0.75 & \textbf{0.00} & Gemini 3 Flash &  & 0.03 & 0.02 & 0.09 & 0.64 & \textbf{0.00} \\
Gemini 3 Flash & \cmark & \textbf{0.07} & 0.08 & \textbf{0.17} & \textbf{0.88} & \textbf{0.00} & Gemini 3 Flash & \cmark & 0.02 & 0.01 & 0.05 & \textbf{0.65} & \textbf{0.00} \\
Gemini 3.1 Pro &  & 0.09 & 0.10 & 0.40 & 0.84 & \textbf{0.00} & Gemini 3.1 Pro &  & 0.03 & 0.02 & 0.09 & 0.00 & \textbf{0.00} \\
Gemini 3.1 Pro & \cmark & 0.08 & \textbf{0.08} & 0.40 & 0.76 & \textbf{0.00} & Gemini 3.1 Pro & \cmark & 0.01 & 0.01 & 0.04 & 0.00 & \textbf{0.00} \\
\addlinespace[0.35em]
\midrule
\addlinespace[0.35em]

\multicolumn{7}{>{\columncolor{green!12}}l}{\textbf{Vernacular Syriac-Kurdish-Turkish-English}} & \multicolumn{7}{>{\columncolor{green!12}}l}{\textbf{Yiddish-English}} \\
\multicolumn{7}{>{\columncolor{gray!12}}l}{\emph{OCR systems}} & \multicolumn{7}{>{\columncolor{gray!12}}l}{\emph{OCR systems}} \\
GLM-OCR &  & 0.63 & 1.67 & 1.59 & 0.00 & 0.45 & GLM-OCR &  & 0.39 & 0.51 & 0.47 & 0.10 & 0.45 \\
GLM-OCR & \cmark & 0.78 & 2.22 & 1.68 & 0.00 & 0.62 & GLM-OCR & \cmark & 0.77 & 0.78 & 0.77 & 0.00 & 0.68 \\
Mathpix &  & 0.59 & 0.54 & 0.63 & 0.00 & 0.45 & Mathpix &  & 0.42 & 0.40 & 0.47 & 0.00 & 0.56 \\
MinerU2.5-Pro &  & 0.58 & 3.03 & 0.73 & 0.00 & 0.41 & MinerU2.5-Pro &  & 0.59 & 0.76 & 0.64 & 0.00 & 0.51 \\
PaddleOCR-VL-1.5 &  & 0.59 & 0.56 & 0.64 & 0.00 & 0.47 & PaddleOCR-VL-1.5 &  & 0.87 & 1.38 & 1.60 & 0.00 & 0.91 \\
\multicolumn{7}{>{\columncolor{gray!12}}l}{\emph{Vision Language Models}} & \multicolumn{7}{>{\columncolor{gray!12}}l}{\emph{Vision Language Models}} \\
Qwen3-VL-235B &  & 0.29 & 0.30 & 0.36 & 0.00 & 0.18 & Qwen3-VL-235B &  & 0.34 & 0.40 & 0.40 & 0.16 & 0.31 \\
Qwen3-VL-235B & \cmark & 0.33 & 0.33 & 0.44 & 0.00 & 0.24 & Qwen3-VL-235B & \cmark & 0.30 & 0.34 & 0.35 & 0.17 & 0.21 \\
\multicolumn{7}{>{\columncolor{gray!12}}l}{\emph{General-purpose LLMs}} & \multicolumn{7}{>{\columncolor{gray!12}}l}{\emph{General-purpose LLMs}} \\
Claude Opus 4.7 &  & 0.17 & 0.16 & 0.27 & 0.59 & 0.32 & Claude Opus 4.7 &  & 0.30 & 0.33 & 0.40 & 0.30 & 0.20 \\
Claude Opus 4.7 & \cmark & 0.16 & 0.15 & 0.28 & 0.58 & 0.19 & Claude Opus 4.7 & \cmark & 0.42 & 0.46 & 0.53 & 0.27 & 0.45 \\
GPT-5.5 &  & 0.13 & 0.13 & 0.27 & 0.59 & 0.02 & GPT-5.5 &  & 0.10 & 0.12 & 0.22 & 0.41 & 0.11 \\
GPT-5.5 & \cmark & 0.14 & 0.12 & 0.25 & 0.59 & 0.02 & GPT-5.5 & \cmark & \textbf{0.10} & \textbf{0.12} & 0.23 & 0.56 & 0.05 \\
Gemini 3 Flash &  & 0.13 & 0.12 & \textbf{0.24} & 0.66 & 0.15 & Gemini 3 Flash &  & 0.31 & 0.34 & 0.34 & 0.45 & 0.24 \\
Gemini 3 Flash & \cmark & \textbf{0.13} & \textbf{0.11} & 0.25 & 0.66 & \textbf{0.02} & Gemini 3 Flash & \cmark & 0.15 & 0.16 & 0.25 & 0.58 & 0.33 \\
Gemini 3.1 Pro &  & 0.14 & 0.12 & 0.24 & 0.69 & 0.23 & Gemini 3.1 Pro &  & 0.22 & 0.25 & 0.35 & 0.35 & 0.09 \\
Gemini 3.1 Pro & \cmark & 0.14 & 0.11 & 0.24 & \textbf{0.72} & 0.03 & Gemini 3.1 Pro & \cmark & 0.14 & 0.17 & \textbf{0.20} & \textbf{0.75} & \textbf{0.01} \\
\bottomrule
\end{tabular}
\end{adjustbox}
\caption{Stage 1 dictionary-specific evaluation results grouped by dictionary (continued).}
\end{table*}

\clearpage

\onecolumn
\section{Stage-1 OCR Assisted Prompting Results}
\label{sec:app-ocr-asst}
\begin{table*}[!h]
\centering
\caption{Per-dictionary breakdown of the Stage 1 OCR-hint ablation summarised in Table~\ref{tab:stage1-aggregate-ocr-hint}. Each metric is shown as a paired (without hint, with hint) value. \textit{Best score per pair is bolded; lower is better except for Markup F1.}}
\scriptsize
\setlength{\tabcolsep}{3pt}
\renewcommand{\arraystretch}{1.08}
\begin{adjustbox}{max width=\textwidth}
\begin{tabular}{llc cc cc cc cc cc}
\toprule
 & & & \multicolumn{2}{c}{\textbf{Edit}} & \multicolumn{2}{c}{\textbf{GCER}} & \multicolumn{2}{c}{\textbf{WER}} & \multicolumn{2}{c}{\textbf{Mrk. F1}} & \multicolumn{2}{c}{\textbf{Order}} \\
\cmidrule(lr){4-5}\cmidrule(lr){6-7}\cmidrule(lr){8-9}\cmidrule(lr){10-11}\cmidrule(lr){12-13}
\textbf{Dictionary} & \textbf{Model} & \textbf{A} & -- & +hint & -- & +hint & -- & +hint & -- & +hint & -- & +hint \\
\midrule
Assyrian-English & Gemini-Pro & \cmark & \textbf{0.218} & 0.230 & \textbf{0.143} & 0.155 & \textbf{0.317} & 0.495 & 0.560 & \textbf{0.618} & 0.059 & \textbf{0.049} \\
Bengalese-English & Gemini-Pro & \cmark & \textbf{0.014} & 0.016 & \textbf{0.007} & 0.011 & \textbf{0.020} & 0.036 & \textbf{0.803} & 0.802 & \textbf{0.000} & \textbf{0.000} \\
Canala-English & Gemini-Pro & \cmark & 0.019 & \textbf{0.007} & 0.015 & \textbf{0.007} & 0.066 & \textbf{0.029} & \textbf{0.000} & \textbf{0.000} & \textbf{0.000} & \textbf{0.000} \\
Chepang-English & Gemini-Pro & \cmark & 0.021 & \textbf{0.019} & 0.022 & \textbf{0.021} & 0.083 & \textbf{0.073} & 0.961 & \textbf{0.966} & \textbf{0.000} & \textbf{0.000} \\
Chukchi-Russian & Gemini-Pro & \cmark & 0.051 & \textbf{0.022} & 0.052 & \textbf{0.024} & 0.299 & \textbf{0.124} & \textbf{0.981} & \textbf{0.981} & \textbf{0.000} & \textbf{0.000} \\
Circassian-English-Turkish & Gemini-Pro & \cmark & \textbf{0.284} & 0.465 & \textbf{0.316} & 0.483 & \textbf{0.665} & 0.702 & \textbf{0.484} & 0.361 & \textbf{0.000} & 0.360 \\
Efik-English & Gemini-Pro & \cmark & \textbf{0.012} & 0.014 & \textbf{0.016} & 0.019 & \textbf{0.052} & 0.069 & \textbf{0.968} & 0.959 & \textbf{0.000} & \textbf{0.000} \\
Evenki-Russian & Gemini-Pro & \cmark & 0.022 & \textbf{0.011} & 0.021 & \textbf{0.009} & 0.120 & \textbf{0.039} & 0.964 & \textbf{0.967} & \textbf{0.000} & \textbf{0.000} \\
Georgian-Russian & Gemini-Pro & \cmark & \textbf{0.040} & 0.043 & \textbf{0.033} & 0.039 & \textbf{0.247} & 0.263 & 0.516 & \textbf{0.818} & 0.058 & \textbf{0.050} \\
Gojri-English-Hindi & Gemini-Pro & \cmark & 0.003 & \textbf{0.000} & 0.003 & \textbf{0.000} & 0.029 & \textbf{0.008} & \textbf{0.000} & \textbf{0.000} & \textbf{0.000} & \textbf{0.000} \\
Greek-English & Gemini-Pro & \cmark & 0.013 & \textbf{0.013} & 0.020 & \textbf{0.018} & \textbf{0.063} & 0.067 & \textbf{0.607} & 0.329 & \textbf{0.000} & \textbf{0.000} \\
Gujarati-English & Gemini-Pro & \cmark & 0.025 & \textbf{0.017} & 0.037 & \textbf{0.029} & 0.149 & \textbf{0.100} & \textbf{0.871} & 0.838 & \textbf{0.000} & \textbf{0.000} \\
Iñupiatun Eskimo-English & Gemini-Pro & \cmark & 0.003 & \textbf{0.002} & \textbf{0.002} & 0.002 & 0.013 & \textbf{0.012} & \textbf{0.995} & 0.979 & \textbf{0.000} & \textbf{0.000} \\
Japanese-English & Gemini-Pro & \cmark & 0.017 & \textbf{0.013} & 0.014 & \textbf{0.010} & 0.043 & \textbf{0.035} & 0.801 & \textbf{0.816} & \textbf{0.005} & 0.010 \\
Kashmiri-English & Gemini-Pro & \cmark & 0.135 & \textbf{0.051} & 0.258 & \textbf{0.068} & 0.453 & \textbf{0.214} & 0.198 & \textbf{0.349} & 0.876 & \textbf{0.000} \\
Khmer-English & Gemini-Pro & \cmark & 0.058 & \textbf{0.029} & 0.099 & \textbf{0.030} & 0.158 & \textbf{0.066} & \textbf{0.000} & \textbf{0.000} & \textbf{0.024} & 0.329 \\
Malay-English & Gemini-Pro & \cmark & 0.034 & \textbf{0.027} & 0.043 & \textbf{0.029} & 0.169 & \textbf{0.103} & \textbf{0.000} & \textbf{0.000} & \textbf{0.000} & \textbf{0.000} \\
Na-English-Chinese-French & Gemini-Pro & \cmark & \textbf{0.008} & 0.017 & \textbf{0.008} & 0.016 & \textbf{0.048} & 0.064 & 0.074 & \textbf{0.667} & \textbf{0.000} & \textbf{0.000} \\
Nahuatl-French & Gemini-Pro & \cmark & 0.003 & \textbf{0.003} & 0.003 & \textbf{0.003} & 0.020 & \textbf{0.018} & 0.961 & \textbf{0.975} & \textbf{0.000} & \textbf{0.000} \\
Punjabi-English & Gemini-Pro & \cmark & 0.038 & \textbf{0.035} & 0.036 & \textbf{0.033} & 0.131 & \textbf{0.108} & 0.880 & \textbf{0.949} & \textbf{0.081} & 0.167 \\
Reel-English & Gemini-Pro & \cmark & \textbf{0.000} & 0.001 & \textbf{0.000} & 0.000 & \textbf{0.000} & 0.002 & 0.977 & \textbf{0.983} & \textbf{0.000} & \textbf{0.000} \\
Ritharngu-English & Gemini-Pro & \cmark & 0.023 & \textbf{0.016} & 0.024 & \textbf{0.017} & 0.109 & \textbf{0.086} & \textbf{0.748} & 0.610 & \textbf{0.000} & \textbf{0.000} \\
Sanskrit-English & Gemini-Pro & \cmark & \textbf{0.026} & 0.027 & \textbf{0.028} & 0.036 & 0.105 & \textbf{0.103} & \textbf{0.787} & 0.651 & \textbf{0.000} & \textbf{0.000} \\
Shilluk-English & Gemini-Pro & \cmark & \textbf{0.009} & 0.009 & 0.004 & \textbf{0.004} & 0.013 & \textbf{0.013} & 0.953 & \textbf{0.957} & \textbf{0.000} & \textbf{0.000} \\
Syriac-English & Gemini-Pro & \cmark & 0.132 & \textbf{0.126} & 0.128 & \textbf{0.121} & 0.230 & \textbf{0.211} & \textbf{0.805} & 0.736 & 0.023 & \textbf{0.011} \\
Telugu-English & Gemini-Pro & \cmark & \textbf{0.030} & 0.031 & \textbf{0.026} & 0.031 & 0.088 & \textbf{0.080} & 0.957 & \textbf{0.970} & \textbf{0.023} & 0.023 \\
Thai-Russian & Gemini-Pro & \cmark & \textbf{0.075} & 0.077 & \textbf{0.082} & 0.084 & \textbf{0.397} & 0.401 & 0.761 & \textbf{0.815} & \textbf{0.000} & \textbf{0.000} \\
Tiri-English & Gemini-Pro & \cmark & 0.014 & \textbf{0.011} & 0.008 & \textbf{0.007} & \textbf{0.036} & \textbf{0.036} & \textbf{0.000} & \textbf{0.000} & \textbf{0.000} & \textbf{0.000} \\
Vernacular Syriac-Kurdish-Turkish-English & Gemini-Pro & \cmark & 0.135 & \textbf{0.124} & 0.113 & \textbf{0.105} & 0.244 & \textbf{0.225} & \textbf{0.720} & 0.621 & 0.025 & \textbf{0.011} \\
Yiddish-English & Gemini-Pro & \cmark & \textbf{0.144} & 0.208 & \textbf{0.166} & 0.231 & \textbf{0.196} & 0.267 & \textbf{0.746} & 0.304 & \textbf{0.010} & 0.055 \\
\midrule
\textit{Aggregate} &  &  & 0.043 & \textbf{0.041} & 0.045 & \textbf{0.043} & 0.122 & \textbf{0.112} & \textbf{0.800} & 0.784 & 0.039 & \textbf{0.036} \\
\bottomrule
\end{tabular}
\end{adjustbox}
\label{tab:stage1-ocr-hint-per-language}
\end{table*}

\begin{table}[!h]
\centering
\small
\setlength{\tabcolsep}{3pt}
\caption{Stage 1 OCR-hint ablation study, aggregated over 30 dictionaries. For each dictionary, we hold Gemini~3.1 Pro with alphabet input fixed and compare transcription with vs.\ without a preliminary OCR transcript supplied to the model. \textit{Best score per metric is bolded.}}
\label{tab:stage1-aggregate-ocr-hint}
\begin{tabular}{lcrrrrr}
\toprule
\textbf{Configuration} & \textbf{OCR hint} & \textbf{Edit} & \textbf{GCER} & \textbf{WER} & \textbf{Mrk. F1} & \textbf{Order} \\
\midrule
Gemini 3.1 Pro + alphabet &  & 0.043 & 0.045 & 0.122 & \textbf{0.800} & 0.039 \\
Gemini 3.1 Pro + alphabet & \cmark & \textbf{0.041} & \textbf{0.043} & \textbf{0.112} & 0.784 & \textbf{0.036} \\
\midrule
$\Delta$ (with hint $-$ without) &  & $-0.002$ & $-0.002$ & $-0.010$ & $-0.016$ & $-0.004$ \\
\bottomrule
\end{tabular}
\end{table}

\clearpage
\onecolumn
\section{Stage 2 Results for Each Dictionary}
\label{sec:app-stage2-dict}

\providecommand{\cmark}{\checkmark}
\begin{table*}[!h]
  \centering
  \scriptsize
  \setlength{\tabcolsep}{2.4pt}
  \renewcommand{\arraystretch}{1.06}
  \begin{adjustbox}{max width=\textwidth}
  \begin{tabular}{lccccc@{\hspace{0.9em}}lccccc}
  \toprule
  \textbf{Model} & \textbf{Intro} & \textbf{MDF} & \textbf{Ent. F1} & \textbf{Fields F1} & \textbf{Order} & \textbf{Model} & \textbf{Intro} & \textbf{MDF} & \textbf{Ent. F1} & \textbf{Fields F1} & \textbf{Order} \\
  \midrule
  \rowcolor{green!12}\multicolumn{6}{l}{\textit{Chukchi-Russian}} & \multicolumn{6}{l}{\textit{Circassian-English-Turkish}} \\
  Gemini 3.1 Pro & \cmark & \cmark & \textbf{1.000} & \textbf{1.000} & \textbf{0.000} & Gemini 3.1 Pro & \cmark & \cmark & \textbf{1.000} & \textbf{1.000} & \textbf{0.000} \\
  Gemini 3.1 Pro & \cmark &  & \textbf{1.000} & \textbf{1.000} & \textbf{0.000} & Gemini 3.1 Pro & \cmark &  & \textbf{1.000} & 0.917 & \textbf{0.000} \\
  Gemini 3.1 Pro &  & \cmark & \textbf{1.000} & \textbf{1.000} & \textbf{0.000} & Gemini 3.1 Pro &  & \cmark & 0.900 & 0.495 & 0.100 \\
  Gemini 3.1 Pro &  &  & \textbf{1.000} & \textbf{1.000} & \textbf{0.000} & Gemini 3.1 Pro &  &  & \textbf{1.000} & 0.909 & \textbf{0.000} \\
  GPT-5.5 & \cmark & \cmark & \textbf{1.000} & \textbf{1.000} & \textbf{0.000} & GPT-5.5 & \cmark & \cmark & 0.976 & 0.858 & 0.048 \\
  GPT-5.5 & \cmark &  & \textbf{1.000} & \textbf{1.000} & \textbf{0.000} & GPT-5.5 & \cmark &  & \textbf{1.000} & \textbf{0.913} & \textbf{0.000} \\
  GPT-5.5 &  & \cmark & \textbf{1.000} & \textbf{1.000} & \textbf{0.000} & GPT-5.5 &  & \cmark & \textbf{1.000} & \textbf{0.913} & \textbf{0.000} \\
  GPT-5.5 &  &  & \textbf{1.000} & 0.963 & \textbf{0.000} & GPT-5.5 &  &  & \textbf{1.000} & 0.756 & \textbf{0.000} \\
  Claude Opus 4.7 & \cmark & \cmark & \textbf{1.000} & 0.968 & \textbf{0.000} & Claude Opus 4.7 & \cmark & \cmark & \textbf{1.000} & 0.880 & \textbf{0.000} \\
  Claude Opus 4.7 & \cmark &  & \textbf{1.000} & \textbf{1.000} & \textbf{0.000} & Claude Opus 4.7 & \cmark &  & \textbf{1.000} & \textbf{0.902} & \textbf{0.000} \\
  Claude Opus 4.7 &  & \cmark & \textbf{1.000} & \textbf{1.000} & \textbf{0.000} & Claude Opus 4.7 &  & \cmark & \textbf{1.000} & 0.901 & \textbf{0.000} \\
  Claude Opus 4.7 &  &  & \textbf{1.000} & \textbf{1.000} & \textbf{0.000} & Claude Opus 4.7 &  &  & 0.850 & 0.442 & 0.150 \\
  Qwen3-VL-235B & \cmark & \cmark & \textbf{1.000} & 0.868 & \textbf{0.000} & Qwen3-VL-235B & \cmark & \cmark & 0.550 & \textbf{0.324} & 0.500 \\
  Qwen3-VL-235B & \cmark &  & 0.906 & 0.763 & 0.143 & Qwen3-VL-235B & \cmark &  & \textbf{0.600} & 0.322 & \textbf{0.450} \\
  Qwen3-VL-235B &  & \cmark & \textbf{1.000} & 0.868 & \textbf{0.000} & Qwen3-VL-235B &  & \cmark & 0.348 & 0.271 & 0.692 \\
  Qwen3-VL-235B &  &  & \textbf{1.000} & \textbf{1.000} & \textbf{0.000} & Qwen3-VL-235B &  &  & 0.550 & 0.290 & 0.500 \\
  \midrule
  \rowcolor{green!12}\multicolumn{6}{l}{\textit{Efik-English}} & \multicolumn{6}{l}{\textit{Evenki-Russian}} \\
  Gemini 3.1 Pro & \cmark & \cmark & \textbf{1.000} & \textbf{0.979} & \textbf{0.000} & Gemini 3.1 Pro & \cmark & \cmark & \textbf{0.964} & 0.570 & \textbf{0.036} \\
  Gemini 3.1 Pro & \cmark &  & \textbf{1.000} & 0.878 & \textbf{0.000} & Gemini 3.1 Pro & \cmark &  & 0.929 & 0.546 & 0.071 \\
  Gemini 3.1 Pro &  & \cmark & \textbf{1.000} & 0.932 & \textbf{0.000} & Gemini 3.1 Pro &  & \cmark & \textbf{0.964} & \textbf{0.584} & \textbf{0.036} \\
  Gemini 3.1 Pro &  &  & \textbf{1.000} & 0.905 & \textbf{0.000} & Gemini 3.1 Pro &  &  & \textbf{0.964} & 0.530 & \textbf{0.036} \\
  GPT-5.5 & \cmark & \cmark & \textbf{1.000} & 0.878 & \textbf{0.000} & GPT-5.5 & \cmark & \cmark & \textbf{1.000} & 0.638 & \textbf{0.000} \\
  GPT-5.5 & \cmark &  & \textbf{1.000} & \textbf{0.888} & \textbf{0.000} & GPT-5.5 & \cmark &  & \textbf{1.000} & 0.611 & \textbf{0.000} \\
  GPT-5.5 &  & \cmark & \textbf{1.000} & 0.878 & \textbf{0.000} & GPT-5.5 &  & \cmark & \textbf{1.000} & \textbf{0.667} & \textbf{0.000} \\
  GPT-5.5 &  &  & \textbf{1.000} & 0.875 & \textbf{0.000} & GPT-5.5 &  &  & \textbf{1.000} & 0.629 & \textbf{0.000} \\
  Claude Opus 4.7 & \cmark & \cmark & \textbf{1.000} & 0.842 & \textbf{0.000} & Claude Opus 4.7 & \cmark & \cmark & 0.929 & 0.570 & 0.071 \\
  Claude Opus 4.7 & \cmark &  & \textbf{1.000} & \textbf{0.900} & \textbf{0.000} & Claude Opus 4.7 & \cmark &  & \textbf{1.000} & 0.662 & \textbf{0.000} \\
  Claude Opus 4.7 &  & \cmark & \textbf{1.000} & 0.862 & \textbf{0.000} & Claude Opus 4.7 &  & \cmark & \textbf{1.000} & \textbf{0.795} & \textbf{0.000} \\
  Claude Opus 4.7 &  &  & \textbf{1.000} & 0.863 & \textbf{0.000} & Claude Opus 4.7 &  &  & \textbf{1.000} & 0.788 & \textbf{0.000} \\
  Qwen3-VL-235B & \cmark & \cmark & \textbf{0.978} & \textbf{0.746} & \textbf{0.043} & Qwen3-VL-235B & \cmark & \cmark & 0.750 & 0.288 & 0.250 \\
  Qwen3-VL-235B & \cmark &  & 0.930 & 0.733 & 0.091 & Qwen3-VL-235B & \cmark &  & \textbf{0.857} & 0.427 & \textbf{0.143} \\
  Qwen3-VL-235B &  & \cmark & 0.000 & 0.000 & 1.000 & Qwen3-VL-235B &  & \cmark & \textbf{0.857} & 0.404 & \textbf{0.143} \\
  Qwen3-VL-235B &  &  & 0.000 & 0.000 & 1.000 & Qwen3-VL-235B &  &  & \textbf{0.857} & \textbf{0.430} & \textbf{0.143} \\
  \midrule
  \rowcolor{green!12}\multicolumn{6}{l}{\textit{Greek-English}} & \multicolumn{6}{l}{\textit{Iñupiatun Eskimo-English}} \\
  Gemini 3.1 Pro & \cmark & \cmark & \textbf{1.000} & \textbf{1.000} & \textbf{0.000} & Gemini 3.1 Pro & \cmark & \cmark & \textbf{1.000} & 0.996 & \textbf{0.000} \\
  Gemini 3.1 Pro & \cmark &  & \textbf{1.000} & 0.921 & \textbf{0.000} & Gemini 3.1 Pro & \cmark &  & \textbf{1.000} & \textbf{1.000} & \textbf{0.000} \\
  Gemini 3.1 Pro &  & \cmark & \textbf{1.000} & 0.878 & \textbf{0.000} & Gemini 3.1 Pro &  & \cmark & \textbf{1.000} & 0.996 & \textbf{0.000} \\
  Gemini 3.1 Pro &  &  & \textbf{1.000} & 0.723 & \textbf{0.000} & Gemini 3.1 Pro &  &  & \textbf{1.000} & 0.996 & \textbf{0.000} \\
  GPT-5.5 & \cmark & \cmark & \textbf{1.000} & 0.819 & \textbf{0.000} & GPT-5.5 & \cmark & \cmark & \textbf{1.000} & 0.946 & \textbf{0.000} \\
  GPT-5.5 & \cmark &  & \textbf{1.000} & \textbf{0.860} & \textbf{0.000} & GPT-5.5 & \cmark &  & \textbf{1.000} & 0.946 & \textbf{0.000} \\
  GPT-5.5 &  & \cmark & \textbf{1.000} & 0.838 & \textbf{0.000} & GPT-5.5 &  & \cmark & 0.992 & \textbf{0.971} & 0.016 \\
  GPT-5.5 &  &  & \textbf{1.000} & 0.838 & \textbf{0.000} & GPT-5.5 &  &  & \textbf{1.000} & 0.946 & \textbf{0.000} \\
  Claude Opus 4.7 & \cmark & \cmark & \textbf{1.000} & \textbf{0.842} & \textbf{0.000} & Claude Opus 4.7 & \cmark & \cmark & \textbf{1.000} & 0.993 & \textbf{0.000} \\
  Claude Opus 4.7 & \cmark &  & \textbf{1.000} & 0.622 & \textbf{0.000} & Claude Opus 4.7 & \cmark &  & \textbf{1.000} & 0.946 & \textbf{0.000} \\
  Claude Opus 4.7 &  & \cmark & \textbf{1.000} & 0.840 & \textbf{0.000} & Claude Opus 4.7 &  & \cmark & \textbf{1.000} & 0.993 & \textbf{0.000} \\
  Claude Opus 4.7 &  &  & \textbf{1.000} & 0.743 & \textbf{0.000} & Claude Opus 4.7 &  &  & \textbf{1.000} & \textbf{0.996} & \textbf{0.000} \\
  Qwen3-VL-235B & \cmark & \cmark & 0.983 & 0.771 & 0.017 & Qwen3-VL-235B & \cmark & \cmark & \textbf{1.000} & \textbf{0.996} & \textbf{0.000} \\
  Qwen3-VL-235B & \cmark &  & 0.983 & 0.583 & 0.033 & Qwen3-VL-235B & \cmark &  & \textbf{1.000} & \textbf{0.996} & \textbf{0.000} \\
  Qwen3-VL-235B &  & \cmark & \textbf{1.000} & \textbf{0.825} & \textbf{0.000} & Qwen3-VL-235B &  & \cmark & \textbf{1.000} & \textbf{0.996} & \textbf{0.000} \\
  Qwen3-VL-235B &  &  & \textbf{1.000} & 0.549 & \textbf{0.000} & Qwen3-VL-235B &  &  & \textbf{1.000} & \textbf{0.996} & \textbf{0.000} \\
  \midrule
  \rowcolor{green!12}\multicolumn{6}{l}{\textit{Kashmiri-English}} & \multicolumn{6}{l}{\textit{Na-English-Chinese-French}} \\
  Gemini 3.1 Pro & \cmark & \cmark & \textbf{1.000} & 0.849 & \textbf{0.000} & Gemini 3.1 Pro & \cmark & \cmark & \textbf{1.000} & 0.871 & \textbf{0.000} \\
  Gemini 3.1 Pro & \cmark &  & \textbf{1.000} & \textbf{0.865} & \textbf{0.000} & Gemini 3.1 Pro & \cmark &  & \textbf{1.000} & 0.641 & \textbf{0.000} \\
  Gemini 3.1 Pro &  & \cmark & \textbf{1.000} & \textbf{0.865} & \textbf{0.000} & Gemini 3.1 Pro &  & \cmark & \textbf{1.000} & \textbf{0.966} & \textbf{0.000} \\
  Gemini 3.1 Pro &  &  & \textbf{1.000} & 0.754 & \textbf{0.000} & Gemini 3.1 Pro &  &  & \textbf{1.000} & 0.771 & \textbf{0.000} \\
  GPT-5.5 & \cmark & \cmark & \textbf{1.000} & 0.501 & \textbf{0.000} & GPT-5.5 & \cmark & \cmark & \textbf{1.000} & \textbf{0.831} & \textbf{0.000} \\
  GPT-5.5 & \cmark &  & \textbf{1.000} & \textbf{0.618} & \textbf{0.000} & GPT-5.5 & \cmark &  & \textbf{1.000} & 0.771 & \textbf{0.000} \\
  GPT-5.5 &  & \cmark & \textbf{1.000} & \textbf{0.618} & \textbf{0.000} & GPT-5.5 &  & \cmark & \textbf{1.000} & 0.519 & \textbf{0.000} \\
  GPT-5.5 &  &  & \textbf{1.000} & \textbf{0.618} & \textbf{0.000} & GPT-5.5 &  &  & \textbf{1.000} & 0.529 & \textbf{0.000} \\
  Claude Opus 4.7 & \cmark & \cmark & \textbf{1.000} & 0.501 & \textbf{0.000} & Claude Opus 4.7 & \cmark & \cmark & \textbf{1.000} & 0.795 & \textbf{0.000} \\
  Claude Opus 4.7 & \cmark &  & \textbf{1.000} & 0.741 & \textbf{0.000} & Claude Opus 4.7 & \cmark &  & \textbf{1.000} & 0.795 & \textbf{0.000} \\
  Claude Opus 4.7 &  & \cmark & \textbf{1.000} & \textbf{0.749} & \textbf{0.000} & Claude Opus 4.7 &  & \cmark & \textbf{1.000} & 0.854 & \textbf{0.000} \\
  Claude Opus 4.7 &  &  & 0.976 & 0.719 & 0.024 & Claude Opus 4.7 &  &  & \textbf{1.000} & \textbf{0.905} & \textbf{0.000} \\
  Qwen3-VL-235B & \cmark & \cmark & \textbf{0.932} & \textbf{0.664} & \textbf{0.128} & Qwen3-VL-235B & \cmark & \cmark & 0.857 & 0.550 & 0.143 \\
  Qwen3-VL-235B & \cmark &  & 0.909 & 0.477 & 0.149 & Qwen3-VL-235B & \cmark &  & \textbf{1.000} & 0.344 & \textbf{0.000} \\
  Qwen3-VL-235B &  & \cmark & \textbf{0.932} & 0.652 & \textbf{0.128} & Qwen3-VL-235B &  & \cmark & 0.857 & 0.468 & 0.143 \\
  Qwen3-VL-235B &  &  & 0.909 & 0.528 & 0.149 & Qwen3-VL-235B &  &  & \textbf{1.000} & \textbf{0.742} & \textbf{0.000} \\
    \bottomrule
  \end{tabular}
  \end{adjustbox}
  \caption{Stage 2 MDF evaluation results grouped by dictionary. Check marks indicate enabled inputs; blank cells indicate disabled inputs. Bold scores mark the best setting for each model within a dictionary (highest Entry F1 and MDF Fields F1; lowest ReadOrderEdit).}
    \label{tab:stage2-mdf-by-dictionary-modelwise-best}
\end{table*}

\clearpage
\onecolumn

\providecommand{\cmark}{\checkmark}
\begin{table*}[!h]
  \centering
  \scriptsize
  \setlength{\tabcolsep}{8.4pt}
  \renewcommand{\arraystretch}{1.06}
  \begin{adjustbox}{max width=\textwidth}
  \begin{tabular}{lccccc@{\hspace{0.9em}}lccccc}
  \toprule
  \rowcolor{green!12}\multicolumn{6}{l}{\textit{Nahuatl-French}} & \multicolumn{6}{l}{\textit{Tiri-English}} \\
  Gemini 3.1 Pro & \cmark & \cmark & \textbf{0.984} & 0.630 & \textbf{0.032} & Gemini 3.1 Pro & \cmark & \cmark & 0.973 & 0.912 & 0.053 \\
  Gemini 3.1 Pro & \cmark &  & \textbf{0.984} & \textbf{0.674} & \textbf{0.032} & Gemini 3.1 Pro & \cmark &  & 0.960 & 0.898 & 0.077 \\
  Gemini 3.1 Pro &  & \cmark & \textbf{0.984} & 0.630 & \textbf{0.032} & Gemini 3.1 Pro &  & \cmark & \textbf{1.000} & \textbf{0.962} & \textbf{0.000} \\
  Gemini 3.1 Pro &  &  & 0.951 & 0.536 & 0.065 & Gemini 3.1 Pro &  &  & 0.864 & 0.804 & 0.222 \\
  GPT-5.5 & \cmark & \cmark & \textbf{0.984} & 0.796 & \textbf{0.032} & GPT-5.5 & \cmark & \cmark & 0.986 & 0.904 & 0.027 \\
  GPT-5.5 & \cmark &  & \textbf{0.984} & \textbf{0.797} & \textbf{0.032} & GPT-5.5 & \cmark &  & 0.960 & 0.827 & 0.077 \\
  GPT-5.5 &  & \cmark & \textbf{0.984} & 0.795 & \textbf{0.032} & GPT-5.5 &  & \cmark & \textbf{1.000} & \textbf{0.908} & \textbf{0.000} \\
  GPT-5.5 &  &  & \textbf{0.984} & 0.686 & \textbf{0.032} & GPT-5.5 &  &  & \textbf{1.000} & 0.904 & \textbf{0.000} \\
  Claude Opus 4.7 & \cmark & \cmark & 0.967 & 0.608 & 0.065 & Claude Opus 4.7 & \cmark & \cmark & \textbf{1.000} & \textbf{0.940} & \textbf{0.000} \\
  Claude Opus 4.7 & \cmark &  & 0.984 & \textbf{0.633} & 0.032 & Claude Opus 4.7 & \cmark &  & \textbf{1.000} & 0.912 & \textbf{0.000} \\
  Claude Opus 4.7 &  & \cmark & 0.984 & 0.601 & 0.032 & Claude Opus 4.7 &  & \cmark & \textbf{1.000} & 0.899 & \textbf{0.000} \\
  Claude Opus 4.7 &  &  & \textbf{1.000} & 0.628 & \textbf{0.000} & Claude Opus 4.7 &  &  & \textbf{1.000} & 0.935 & \textbf{0.000} \\
  Qwen3-VL-235B & \cmark & \cmark & \textbf{0.951} & \textbf{0.431} & 0.129 & Qwen3-VL-235B & \cmark & \cmark & 0.928 & 0.759 & \textbf{0.139} \\
  Qwen3-VL-235B & \cmark &  & 0.918 & 0.411 & 0.161 & Qwen3-VL-235B & \cmark &  & \textbf{0.930} & \textbf{0.787} & \textbf{0.139} \\
  Qwen3-VL-235B &  & \cmark & 0.918 & 0.388 & 0.129 & Qwen3-VL-235B &  & \cmark & 0.897 & 0.694 & 0.167 \\
  Qwen3-VL-235B &  &  & \textbf{0.951} & 0.430 & \textbf{0.065} & Qwen3-VL-235B &  &  & 0.904 & 0.720 & 0.189 \\
  \bottomrule
  \end{tabular}
  \end{adjustbox}
  \caption{Stage 2 MDF evaluation results grouped by dictionary. Check marks indicate enabled inputs; blank cells indicate disabled inputs. Bold scores mark the best setting for each model within a dictionary (highest Entry F1 and MDF Fields F1; lowest ReadOrderEdit).}
    \label{tab:stage2-mdf-by-dictionary-modelwise-best}
\end{table*}

\clearpage
\onecolumn
\section{Stage 2 Analysis}
\label{sec:app-stage-2-analisys}
\providecommand{\cmark}{\checkmark}


\begin{table}[!h]
\centering
\small
\caption{Stage~2 gold parse-rules diagnostic for every dictionary with imperfect MDF Field F1 under its best per-dictionary configuration, except Efik. Each row replaces the inferred Pass~1 parse-rules with human-validated gold parse-rules before Pass~2 while retaining the same model and introduction/manual setting.}
\label{tab:stage2-gold-cheat-sheet}
\setlength{\tabcolsep}{4pt}
\begin{tabular}{l l cc rr}
\toprule
\textbf{Dictionary} & \textbf{Model} & \textbf{Intro} & \textbf{MDF} & \textbf{Inf. F1} & \textbf{Gold F1} \\
\midrule
Evenki & Claude Opus 4.7 &  & \cmark & 0.80 & \textbf{0.91} \\
Kashmiri & Gemini 3.1 Pro & \cmark &  & 0.87 & \textbf{0.95} \\
Na (Mosuo) & Gemini 3.1 Pro &  & \cmark & \textbf{0.97} & 0.94 \\
Tiri & Gemini 3.1 Pro &  & \cmark & \textbf{0.96} & 0.91 \\
Nahuatl & GPT-5.5 & \cmark & \cmark & 0.80 & \textbf{0.88} \\
\midrule
\textbf{Macro avg.} &  &  &  & 0.88 & \textbf{0.92} \\
\bottomrule
\end{tabular}
\end{table}

\begin{table}[!h]
\centering
\small
\setlength{\tabcolsep}{2pt}
\caption{Stage 2 evaluation results aggregated by model, end-to-end condition: Stage 2 fed Stage 1's own predicted transcription rather than the gold-standard transcriptions. }
\label{tab:stage2-mdf-aggregate-e2e}
\begin{tabular}{lccccc}
\toprule
\textbf{Model} & \textbf{Intro} & \textbf{MDF} & \textbf{Ent. F1} & \textbf{MDF F1} & \textbf{ROE} \\
\midrule
\multicolumn{6}{l}{\cellcolor{gray!10}\textit{Vision Language Models}} \\
Qwen3-VL-235B &  &  & 0.96 & 0.65 & 0.09 \\
Qwen3-VL-235B & \cmark &  & 0.88 & 0.56 & 0.19 \\
Qwen3-VL-235B &  & \cmark & 0.92 & 0.57 & 0.20 \\
Qwen3-VL-235B & \cmark & \cmark & 0.94 & 0.64 & 0.19 \\
\midrule
\multicolumn{6}{l}{\cellcolor{gray!10}\textit{General-purpose LLMs}} \\
Claude Opus 4.7 &  &  & 0.96 & 0.76 & 0.08 \\
Claude Opus 4.7 & \cmark &  & 0.96 & 0.72 & 0.07 \\
Claude Opus 4.7 &  & \cmark & 0.95 & 0.76 & 0.08 \\
Claude Opus 4.7 & \cmark & \cmark & 0.96 & 0.78 & 0.08 \\
GPT-5.5 &  &  & \textbf{0.99} & 0.76 & \textbf{0.01} \\
GPT-5.5 & \cmark &  & \textbf{0.99} & 0.80 & \textbf{0.01} \\
GPT-5.5 &  & \cmark & 0.98 & 0.77 & 0.02 \\
GPT-5.5 & \cmark & \cmark & \textbf{0.99} & 0.77 & 0.02 \\
Gemini 3.1 Pro &  &  & \textbf{0.99} & 0.79 & 0.02 \\
Gemini 3.1 Pro & \cmark &  & 0.94 & 0.75 & 0.10 \\
Gemini 3.1 Pro &  & \cmark & \textbf{0.99} & \textbf{0.82} & 0.02 \\
Gemini 3.1 Pro & \cmark & \cmark & \textbf{0.99} & 0.78 & \textbf{0.01} \\
\bottomrule
\end{tabular}
\end{table}

\begin{table}[!h]
\centering
\small
\setlength{\tabcolsep}{3pt}
\caption{Stage 2 evaluation results with vs.\ without bold and italic markups in the gold Stage 2 input. Baseline values are the aggregate scores from Table~\ref{tab:stage2-mdf-aggregate}. \textit{Best score per pair is bolded: higher is better for Ent. F1 and MDF F1, while lower is better for ROE.}}
\label{tab:stage2-no-typography}
\begin{tabular}{lcccccccc}
\toprule
 & & & \multicolumn{2}{c}{\textbf{Ent. F1}} & \multicolumn{2}{c}{\textbf{MDF F1}} & \multicolumn{2}{c}{\textbf{ROE}} \\
\cmidrule(lr){4-5}\cmidrule(lr){6-7}\cmidrule(lr){8-9}
\textbf{Model} & \textbf{Intro} & \textbf{MDF} & \textbf{Tag} & \textbf{No Tag} & \textbf{Tag} & \textbf{No Tag} & \textbf{Tag} & \textbf{No Tag} \\
\midrule
Claude Opus 4.7 & \cmark & \cmark & 0.99 & \textbf{0.99} & 0.78 & \textbf{0.82} & 0.01 & \textbf{0.01} \\
Gemini 3.1 Pro & \cmark & \cmark & \textbf{0.99} & 0.98 & \textbf{0.88} & 0.79 & \textbf{0.01} & 0.04 \\
GPT-5.5 & \cmark & \cmark & \textbf{1.00} & 0.99 & 0.80 & \textbf{0.83} & \textbf{0.01} & 0.01 \\
\bottomrule
\end{tabular}
\end{table}

\clearpage
\twocolumn

\section{Prompts}
\label{sec:appendix-prompts}

This appendix reproduces the prompts used in our two-stage pipeline. Dynamic
values are shown in \texttt{\{braces\}}, and optional blocks are delimited by
\texttt{\{\% if ... \%\}} and \texttt{\{\% endif \%\}} conditions. Stage~1 produces a
structured transcription with \texttt{header}, \texttt{lines}, and
\texttt{footer} lists, which the pipeline serializes as newline-delimited text.
Stage~2 produces MDF text in two passes: Pass~1 discovers the dictionary's field
map, and Pass~2 exports each dictionary page as MDF.

\subsection{Stage 1: Faithful page transcription}
\label{sec:appendix-stage1}

\subsubsection{System prompt}
\label{sec:appendix-stage1-system}

\begin{promptbox}{System prompt \normalfont\small\textit{(stage\_1\_system)}}
\raggedright

You are a precise OCR transcription system specialising in historical and minority-language dictionaries. \\
\medskip

Your task is faithful OCR only --- do NOT parse dictionary entries or assign fields. \\
\medskip

Output structure:

{\hangindent=1.2em\hangafter=1\noindent
\textbullet\ \texttt{header}: page-level lines at the very top (running title, page number, letter band). One string per visible line. Empty list if none. Never put dictionary entries here.\par}

{\hangindent=1.2em\hangafter=1\noindent
\textbullet\ \texttt{lines}: every visible BODY line in reading order. For multi-column pages, transcribe the full left column top-to-bottom, then the next column, and so on --- as a single ordered list (no \texttt{column\_id} labels).\par}

{\hangindent=1.2em\hangafter=1\noindent
\textbullet\ \texttt{footer}: page-level lines at the very bottom (page numbers, footnotes, rules). One string per visible line. Empty list if none.\par}

\medskip
You may receive \texttt{\textless{}ocr\_reference\textgreater{}...\textless{}/ocr\_reference\textgreater{}} from a standard OCR engine. Use it only for ambiguous character shapes; always prioritise the page image.

\medskip

Rules for every line in header, lines, and footer:

{\hangindent=1.2em\hangafter=1\noindent
\textbullet\ Preserve ALL diacritics, stress marks, and special phonetic symbols exactly.\par}

{\hangindent=1.2em\hangafter=1\noindent
\textbullet\ Do NOT interpret, summarise, merge lines, or fix typos.\par}

{\hangindent=1.2em\hangafter=1\noindent
\textbullet\ Do NOT skip lines, including continuations and cross-references.\par}

{\hangindent=1.2em\hangafter=1\noindent
\textbullet\ Hyphenated wraps: when a word breaks across two printed lines with a trailing hyphen, emit TWO separate strings (e.g.\ ``intelligi-'' then ``ble, adj.\ clear'').\par}

\medskip
Typography annotation:

{\hangindent=1.2em\hangafter=1\noindent
\textbullet\ Wrap bold text in \texttt{\textless{}b\textgreater{}...\textless{}/b\textgreater{}} and italic text in \texttt{\textless{}i\textgreater{}...\textless{}/i\textgreater{}} when confident.\par}

{\hangindent=1.2em\hangafter=1\noindent
\textbullet\ Only mark formatting you are confident about; when in doubt, leave text plain.\par}

{\hangindent=1.2em\hangafter=1\noindent
\textbullet\ These tags may appear anywhere within a line, e.g.\ \texttt{\textless{}b\textgreater{}headword\textless{}/b\textgreater{} translation}.\par}
\end{promptbox}

\subsubsection{User prompt}
\label{sec:appendix-stage1-user}
\begin{promptbox}{User message template \normalfont\small\textit{(stage\_1\_user)}}
\footnotesize
\texttt{\{\% if alphabet\_text \%\}}\\
\texttt{\textless{}alphabet\textgreater{}}\\
\texttt{\{\{ alphabet\_text \}\}}\\
\texttt{\textless{}/alphabet\textgreater{}}\\[0.25em]

The \texttt{\textless{}alphabet\textgreater{}} is a reference guide to the list of characters in the script of the source language, not a strict whitelist. It may be incomplete or not perfectly match this document's script variant.\\[0.25em]

Rules:\\
1. Prefer \texttt{\textless{}alphabet\textgreater{}} matches over visually similar characters from other scripts.\\
2. For combinatorial scripts (Indic conjuncts, Ethiopic syllables, Hangul blocks, Arabic ligatures, pointed Hebrew/Syriac), treat \texttt{\textless{}alphabet\textgreater{}} entries as base components and form composites using the script's standard rules (virama, vowel diacritics, contextual forms).\\
3. If a glyph is clearly a legitimate character of the target script but not in \texttt{\textless{}alphabet\textgreater{}} (archaic letters, extensions, related-language characters), transcribe it correctly anyway --- do not force-fit.\\
4. Preserve diacritics, tone marks, case, and period-specific orthography exactly as shown.\\
5. Mark truly unidentifiable glyphs as \texttt{[?]}.\\
\texttt{\{\% endif \%\}}\\[0.5em]

\texttt{\{\% if ocr\_hint \%\}}\\
\texttt{\textless{}ocr\_reference\textgreater{}}\\
\texttt{\{\{ ocr\_hint \}\}}\\
\texttt{\textless{}/ocr\_reference\textgreater{}}\\[0.25em]

The OCR reference above may contain errors but can help you identify ambiguous character shapes.\\
\texttt{\{\% endif \%\}}\\[0.5em]

Now transcribe every line of text from the dictionary page image exactly as it appears. Preserve all diacritics and special characters.\\[0.5em]

\texttt{\{\% if guides \%\}}\\
USER DEFINED GUIDELINES\\
\texttt{\{\{ guides \}\}}\\
\texttt{\{\% endif \%\}}\\[0.5em]

\textit{[Attached after the preceding text: dictionary page image.]}
\end{promptbox}

\subsection{Stage 2: MDF generation from dictionary pages}
\label{sec:appendix-stage2}

The following prompts were used for Stage~2. 

\subsubsection{MDF parsing guide system prompt (Pass 1)}
\label{sec:appendix-stage2-pass1}

\begin{promptbox}{System prompt \normalfont\small\textit{(stage\_2\_pass\_1\_system)}}
You analyse dictionary front matter and a sample page to list which SIL Toolbox MDF
markers this dictionary uses and how entries are structured.

\end{promptbox}

\begin{promptbox}{System prompt \normalfont\small\textit{(stage\_2\_pass\_1\_system)}: continued}

Output a single JSON object --- NOT a full typography map. For each marker that appears
in this dictionary, give a one-line description (role + language tier). Add structure
rules for entry boundaries (homographs, senses, subentries, gloss lines).

Use MDF marker codes from the reference below. Pick gloss/usage/definition tiers to
match the dictionary's target languages (e=English, n=national, r=regional, v=vernacular).
Do NOT assign English-tier markers (ge, de, ue) when the dictionary has no English target.

\medskip
\noindent\textbf{MDF marker vocabulary (SIL Toolbox --- assign ONLY markers whose content appears on
THIS dictionary's intro + sample page; pick gloss/usage/definition tier to match target
languages: e=English, n=national, r=regional, v=vernacular):}

\medskip
\noindent\textbf{Entry structure (section boundaries; \textbackslash se, \textbackslash ps, \textbackslash sn order affects MDF printing)}\\
\hspace*{1em}\texttt{lx} \quad Lexeme / headword. Record marker; one per lexical entry; primary sort key.\\
\hspace*{1em}\texttt{lc} \quad Use instead when a more readable citation form should print.\\
\hspace*{1em}\texttt{hm} \quad Homonym number. Marks unrelated homographs; digits 1, 2, 3 immediately after \texttt{\textbackslash lx}.\\
\hspace*{1em}\texttt{se} \quad Subentry. Like \texttt{\textbackslash lx} but inside the record.\\
\hspace*{1em}\texttt{sn} \quad Sense number. Use 1, 2, 3 with no punctuation.\\
\hspace*{1em}\texttt{mn} \quad Main entry form. Points a minimal/minor entry back to its full main entry.

\medskip
\noindent\textbf{Glosses (short translation; used for interlinearizing and as gloss when no definition)}\\
\hspace*{1em}\texttt{ge} \quad Gloss (English). Short English equivalent.\\
\hspace*{1em}\texttt{gn} \quad Gloss (national). Short national-language equivalent.\\
\hspace*{1em}\texttt{gr} \quad Gloss (regional). Clarifies a national gloss in a regional language.\\
\hspace*{1em}\texttt{gv} \quad Gloss (vernacular). Mainly for monolingual lexicons.

\medskip
\noindent\textbf{Definitions (longer explication of semantic domain; suppresses gloss in print)}\\
\hspace*{1em}\texttt{de} \quad Definition (English).\\
\hspace*{1em}\texttt{dn} \quad Definition (national).\\
\hspace*{1em}\texttt{dr} \quad Definition (regional).\\
\hspace*{1em}\texttt{dv} \quad Definition (vernacular).\\
\hspace*{1em}\texttt{lt} \quad Literal meaning. Explains parts of an idiom or complex phrase.

\medskip
\noindent\textbf{Reversal / finderlist (normally does NOT print in the dictionary body)}\\
\hspace*{1em}\texttt{re} \quad Reversal form (English).\\
\hspace*{1em}\texttt{rn} \quad Reversal form (national).\\
\hspace*{1em}\texttt{rr} \quad Reversal form (regional).

\medskip
\noindent\textbf{Word-level glosses (morpheme-level interlinear, not full-sentence translation)}\\
\hspace*{1em}\texttt{we} \quad Word gloss (English).\\
\hspace*{1em}\texttt{wn} \quad Word gloss (national).\\
\hspace*{1em}\texttt{wr} \quad Word gloss (regional).

\medskip
\noindent\textbf{Grammar \& inflectional form}\\
\hspace*{1em}\texttt{ps} \quad Part of speech. Classifies the vernacular lexeme.\\
\hspace*{1em}\texttt{pn} \quad Part of speech (national). Follows \texttt{\textbackslash ps}.\\
\hspace*{1em}\texttt{ph} \quad Phonetic / phonemic form.\\
\hspace*{1em}\texttt{pl} \quad Plural form.\\
\hspace*{1em}\texttt{sg} \quad Singular form.\\
\hspace*{1em}\texttt{pd} \quad Paradigm set.\\

\end{promptbox}

\begin{promptbox}{System prompt \normalfont\small\textit{(stage\_2\_pass\_1\_system)}: continued}

\hspace*{1em}\texttt{rd} \quad Reduplication form(s).\\
\hspace*{1em}\texttt{mr} \quad Morphemic representation.

\medskip
\noindent\textbf{Examples (bundled: \textbackslash rf + \textbackslash xv + \textbackslash xe/\textbackslash xn/\textbackslash xr per example)}\\
\hspace*{1em}\texttt{xv} \quad Example sentence (vernacular).\\
\hspace*{1em}\texttt{xe} \quad Example translation (English).\\
\hspace*{1em}\texttt{xn} \quad Example translation (national).\\
\hspace*{1em}\texttt{xr} \quad Example translation (regional).\\
\hspace*{1em}\texttt{xg} \quad Interlinear gloss line.\\
\hspace*{1em}\texttt{rf} \quad Reference to source.

\medskip
\noindent\textbf{Semantics \& lexical relations}\\
\hspace*{1em}\texttt{sd} \quad Semantic domain.\\
\hspace*{1em}\texttt{is} \quad Index of semantics.\\
\hspace*{1em}\texttt{th} \quad Thesaurus.\\
\hspace*{1em}\texttt{sy} \quad Synonym.\\
\hspace*{1em}\texttt{an} \quad Antonym.\\
\hspace*{1em}\texttt{lf} \quad Lexical function label.\\
\hspace*{1em}\texttt{lv} \quad Vernacular lexeme referenced by \texttt{\textbackslash lf}.\\
\hspace*{1em}\texttt{le}, \texttt{ln}, \texttt{lr} \quad Gloss of \texttt{\textbackslash lv} (English, national, regional).\\
\hspace*{1em}\texttt{cf} \quad Cross-reference.\\
\hspace*{1em}\texttt{ce}, \texttt{cn}, \texttt{cr} \quad Cross-reference gloss (English, national, regional).

\medskip
\noindent\textbf{Etymology \& encyclopedic}\\
\hspace*{1em}\texttt{et} \quad Etymology. Proto-form or source form.\\
\hspace*{1em}\texttt{eg} \quad Etymology gloss.\\
\hspace*{1em}\texttt{es} \quad Etymology source.\\
\hspace*{1em}\texttt{ec} \quad Etymology comment.\\
\hspace*{1em}\texttt{ee}, \texttt{en}, \texttt{er}, \texttt{ev} \quad Encyclopedic information (English, national, regional, vernacular).\\
\hspace*{1em}\texttt{bw} \quad Borrowed word.

\medskip
\noindent\textbf{Variants, usage, restrictions}\\
\hspace*{1em}\texttt{va} \quad Variant forms.\\
\hspace*{1em}\texttt{ve}, \texttt{vn}, \texttt{vr} \quad Variant comment (English, national, regional).\\
\hspace*{1em}\texttt{ue} \quad Usage (English).\\
\hspace*{1em}\texttt{un} \quad Usage (national).\\
\hspace*{1em}\texttt{ur} \quad Usage (regional).\\
\hspace*{1em}\texttt{uv} \quad Usage (vernacular).\\
\hspace*{1em}\texttt{oe} \quad Restriction / ``only'' (English).\\
\hspace*{1em}\texttt{on}, \texttt{or}, \texttt{ov} \quad Restriction (national, regional, vernacular).

\medskip
\noindent\textbf{Other (uncommon on printed body pages)}\\
\hspace*{1em}\texttt{sc} \quad Scientific name.\\
\hspace*{1em}\texttt{so} \quad Source of data.\\
\hspace*{1em}\texttt{st} \quad Status.\\
\hspace*{1em}\texttt{dt} \quad Datestamp.\\
\hspace*{1em}\texttt{bb} \quad Bibliographical reference.\\
\hspace*{1em}\texttt{pc} \quad Picture / graphic link.\\
\hspace*{1em}\texttt{tb} \quad Table.\\
\hspace*{1em}\texttt{na}, \texttt{nd}, \texttt{ng}, \texttt{np}, \texttt{nq},
\texttt{ns}, \texttt{nt} \quad Compiler notes by domain (anthropology, discourse,
grammar, phonology, questions, sociolinguistics, general).
\end{promptbox}

\subsubsection{MDF parsing guide user prompt
(Pass 1)}
\label{sec:appendix-stage2-pass1-user-single}

\begin{promptbox}{User prompt \normalfont\small\textit{(stage\_2\_pass\_1\_user)}}
Study this dictionary and return a MDF parsing guide JSON object.

\par\vspace{0.7\baselineskip}
Inputs:

1. Introduction pages (attached below as images or PDF documents).

\end{promptbox}
\begin{promptbox}{User prompt \normalfont\small\textit{(stage\_2\_pass\_1\_user)}: continued}

2. One sample dictionary page (image/PDF + flat transcription).

\par\vspace{0.7\baselineskip}
\texttt{\{\% if config\_hint \%\}}\\
\{config\_hint\}\\
\texttt{\{\% endif \%\}}

\par\vspace{0.7\baselineskip}
Return JSON with this shape:

\texttt{\{}\\
\hspace*{1em}\texttt{"markers": [}\\
\hspace*{2em}\texttt{\{"marker": "lx", "description": "headword / lexeme (vernacular lemma)"\},}\\
\hspace*{2em}\texttt{\{"marker": "gn", "description": "Russian gloss / translation"\},}\\
\hspace*{2em}\texttt{\{"marker": "un", "description": "usage note (parenthetical Russian expansion)"\}}\\
\hspace*{1em}\texttt{],}\\
\hspace*{1em}\texttt{"rules": [}\\
\hspace*{2em}\texttt{"Separate homograph mains: two \textbackslash lx blocks with same lemma, each with \textbackslash hm N.",}\\
\hspace*{2em}\texttt{"Numbered senses: one \textbackslash lx, then \textbackslash sn 1, \textbackslash gn \ldots, \textbackslash sn 2, \textbackslash gn \ldots",}\\
\hspace*{2em}\texttt{"Run-on sub-entries: \textbackslash se sub-lemma then \textbackslash gn \ldots\ inside the same record block."}\\
\hspace*{1em}\texttt{],}\\
\hspace*{1em}\texttt{"abbreviations": \{"мн.": "plural"\}}\\
\texttt{\}}

\par\vspace{0.7\baselineskip}
Include ONLY markers that appear on the sample page or are defined in the intro.
Omit markers not used. Keep descriptions short (one line each). Rules should cover
entry structure and gloss-line conventions (semicolons, hyphen rejoining).
Return JSON only --- no markdown fences.

\par\vspace{0.7\baselineskip}
\texttt{<sample\_transcription>}\\
\{transcription\}\\
\texttt{</sample\_transcription>}
\end{promptbox}

\subsubsection{MDF extraction system prompt (Pass 2)}
\label{sec:appendix-stage2-pass2-system}

\begin{promptbox}{System prompt \normalfont\small\textit{(stage\_2\_pass\_2\_system)}}
You digitize dictionary pages into SIL Toolbox MDF (Multi-Dictionary Formatter) text.

\par\vspace{0.7\baselineskip}
The flat transcription is the provided Stage-1 OCR text for this page. Copy every
vernacular and gloss character from it exactly --- do NOT re-read, correct, normalise,
or substitute letters from the page image. Use the page image and field map (from
Pass~1) only to decide entry boundaries, field roles, and MDF marker assignment.

\par\vspace{0.7\baselineskip}
Output ONLY MDF text --- no JSON, no markdown fences, no commentary.
Use one blank line between lexicon records (between main entry blocks).

\par\vspace{0.7\baselineskip}
Allowed changes (structure only --- never alter alphabet/characters):

\texttt{-} Strip typography markup (\texttt{<b>}, \texttt{<i>}) when emitting MDF
lines; do not change the underlying characters inside those spans.\\
\texttt{-} Rejoin hyphenated line breaks from the transcription when forming
headwords/glosses.\\
\texttt{-} Normalise sense/homograph numbers: \texttt{\textbackslash sn} and
\texttt{\textbackslash hm} as integers 1, 2, 3, \ldots\ (convert Roman numerals; strip
trailing \texttt{)} or \texttt{.}).

\end{promptbox}
\begin{promptbox}{System prompt \normalfont\small\textit{(stage\_2\_pass\_2\_system)}: continued}

\texttt{-} Strip end-of-sentence punctuation (\texttt{.}, \texttt{!}, \texttt{?}) from
gloss, definition, and example field values --- MDF does not require them.\\
\texttt{-} Split or merge lines across MDF marker fields per the field map
(e.g.\ \texttt{\textbackslash un} vs.\ \texttt{\textbackslash gn}).\\
\texttt{-} For note fields (\texttt{\textbackslash np}, \texttt{\textbackslash nt},
\texttt{\textbackslash ng}, etc.): when the field map or transcription includes a
printed English label (e.g.\ \texttt{"Tone:"}, \texttt{"Commonly used classifier:"}),
keep that label in the field value unless the field map explicitly says otherwise.

\par\vspace{0.7\baselineskip}
Do NOT invent entries or text spans absent from the transcription.
\end{promptbox}

\subsubsection{MDF extraction user prompt (Pass 2)}
\label{sec:appendix-stage2-pass2-user}

\begin{promptbox}{System prompt \normalfont\small\textit{(stage\_2\_pass\_2\_user)}}
Dictionary page extraction.

\par\vspace{0.7\baselineskip}
Inputs:

1. Stage-1 OCR transcription (bold = \texttt{<b>\ldots</b>}, italic =
\texttt{<i>\ldots</i>}) or column TSV (\texttt{column\_id}, \texttt{line\_number},
\texttt{text}). Treat this as authoritative for all characters.

2. Photo of the dictionary page (image below) --- layout and field-boundary reference only.

\par\vspace{0.7\baselineskip}
\texttt{\{\% if toolbox\_reference\_mode == 'pdf' \%\}}\\
4. Attached SIL Toolbox MDF Reference Manual (PDF) --- marker reference.\\
\texttt{\{\% elif toolbox\_reference\_mode == 'text\_fallback' \%\}}\\
4. SIL Toolbox MDF Reference Manual (text excerpt) --- marker reference:\\
\{mdf\_marker\_reference\}\\
\texttt{\{\% endif \%\}}

\par\vspace{0.7\baselineskip}
\{field\_block\}

\par\vspace{0.7\baselineskip}
Parse every dictionary entry on the page into MDF text using the field map above.
Copy headword and gloss characters verbatim from the transcription except for the
structural normalisations listed in the system prompt.

\par\vspace{0.7\baselineskip}
\texttt{<transcription>}\\
\{transcription\}\\
\texttt{</transcription>}

\par\vspace{0.7\baselineskip}
\texttt{\{\% if guides \%\}}\\
USER DEFINED GUIDELINES\\
\{guides\}\\
\texttt{\{\% endif \%\}}
\end{promptbox}

\subsection{Model Cost}
To guide practitioners in balancing performance with budget constraints, we present a comprehensive cost analysis of our proposed digitization pipelines. For a comparable baseline, these end-to-end estimates use a bare prompt (no alphabet or OCR hints) for Stage 1 extraction, combined with a comprehensive context prompt (introductory pages and MDF guidelines) for Stage 2 parsing.

\begin{table}[!h]
\centering
\small
\setlength{\tabcolsep}{2pt}
\begin{tabular}{l ccr}
\toprule
\textbf{Model} & \textbf{S1 \$/Page} & \textbf{S2 \$/Page} & \textbf{Total \$/Page} \\
\midrule
\multicolumn{4}{l}{\textbf{Free / Open-Weight Tier}} \\
\midrule
MinerU 2.5 Pro               & \$0     & ---     & \textbf{\$0} \\
PaddleOCR-VL 1.5             & \$0     & ---     & \textbf{\$0} \\
GLM-OCR                      & \$0     & ---     & \textbf{\$0} \\
\midrule
\multicolumn{4}{l}{\textbf{Budget Tier}} \\
\midrule
Qwen3-VL 235B                & \$0.002 & \$0.005 & \textbf{\textasciitilde\$0.007} \\
Gemini 3 Flash               & \$0.005 & ---     & \textbf{\$0.005} \\
Mathpix-OCR                  & \$0.005 & ---     & \textbf{\$0.005} \\
\midrule
\multicolumn{4}{l}{\textbf{Premium Tier}} \\
\midrule
Gemini 3.1 Pro               & \$0.047 & \$0.085 & \textbf{\textasciitilde\$0.132} \\
Claude Opus 4.7              & \$0.070 & \$0.259 & \textbf{\textasciitilde\$0.329} \\
GPT-5.5                      & \$0.069 & \$1.566 & \textbf{\textasciitilde\$1.635} \\
\bottomrule
\end{tabular}
\caption{End-to-end pipeline costs per page. Total cost is the sum of Stage 1 (bare configuration) and Stage 2 (intro + MDF configuration).}
\label{tab:aggregated_costs_per_page}
\end{table}


\end{document}